\documentclass[10pt,twocolumn,letterpaper]{article}
\usepackage[accsupp]{axessibility}
\usepackage[pagenumber]{cvpr}  
\definecolor{cvprblue}{rgb}{0.21,0.49,0.74}
\usepackage[pagebackref,breaklinks,colorlinks,allcolors=cvprblue]{hyperref}
\usepackage[T1]{fontenc} 
\usepackage{url}         
\usepackage{amsfonts}    
\usepackage{nicefrac}    
\usepackage{microtype}
\usepackage{xcolor}
\usepackage{amsmath}
\usepackage{multirow}
\usepackage{enumitem}
\usepackage{makecell}
\usepackage{booktabs}
\usepackage{graphicx}
\usepackage[export]{adjustbox}
\usepackage{subcaption}
\usepackage{wrapfig}
\usepackage[linewidth=1pt]{mdframed}
\usepackage{graphicx}
\usepackage{tabularx}
\usepackage{algorithm}
\usepackage{algpseudocode}
\usepackage{listings}
\usepackage{multicol}
\def\eg{\textit{e.g.}}
\newcommand{\BEST}[1]{\textbf{\textcolor[rgb]{1.00,0.00,0.00}{#1}}}
\newcommand{\SBEST}[1]{\textbf{\textcolor[rgb]{0.00,0.00,1.00}{#1}}}

\title{Generative Edge Detection with Stable Diffusion}

\author{Caixia Zhou$^{1}$, Yaping Huang$^{1}$, Mochu Xiang$^{2}$, Jiahui Ren$^{2}$, Haibin Ling$^{3}$, Jing Zhang$^{4}$\\ 
$^1$Beijing Key Laboratory of Traffic Data Analysis and Mining, Beijing Jiaotong University \\ 
$^2$ Northwest Polytechnical University  \quad
$^3$Stony Brook University  \quad
$^4$Australian National University\\
 {\tt\small
 \{cxzhou,yphuang\}@bjtu.edu.cn, \{xiangmochu,renjiahui\}@mail.nwpu.edu.cn,}\\
  {\tt\small hling@cs.stonybrook.edu, jing.zhang@anu.edu.au}
}

\begin{document}
\maketitle

\begin{abstract}
Edge detection is typically viewed as a pixel-level classification problem mainly addressed by discriminative methods. Recently, generative edge detection methods, especially diffusion model based solutions, are initialized in the edge detection task. Despite great potential, the retraining of task-specific designed modules and multi-step denoising inference limits their broader applications. Upon closer investigation, we speculate that part of the reason is the under-exploration of the rich discriminative information encoded in extensively pre-trained large models (\eg, stable diffusion models). Thus motivated, we propose a novel approach, named Generative Edge Detector (GED), by fully utilizing the potential of the pre-trained stable diffusion model. Our model can be trained and inferred efficiently without specific network design due to the rich high-level and low-level prior knowledge empowered by the pre-trained stable diffusion.  Specifically, we propose to finetune the denoising U-Net and predict latent edge maps directly, by taking the latent image feature maps as input. Additionally, due to the subjectivity and ambiguity of the edges, we also incorporate the granularity of the edges into the denoising U-Net model as one of the conditions to achieve controllable and diverse predictions. Furthermore, we devise a granularity regularization to ensure the relative granularity relationship of the multiple predictions. We conduct extensive experiments on multiple datasets and achieve competitive performance (\eg, 0.870 and 0.880 in terms of ODS and OIS on the BSDS test dataset).
\end{abstract}

\begin{figure}[!t]
\small
\centering
\begin{tabular}{cccc}
\hspace{-.3cm}
\includegraphics[width=.1\textwidth]{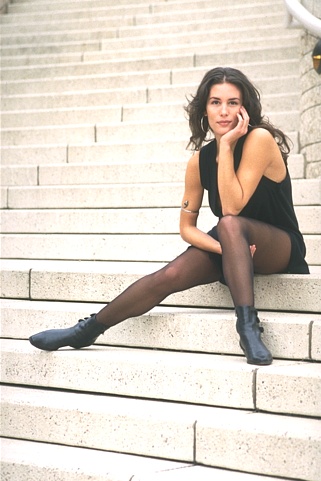}
&\hspace{-.3cm}\includegraphics[width=.1\textwidth,frame]{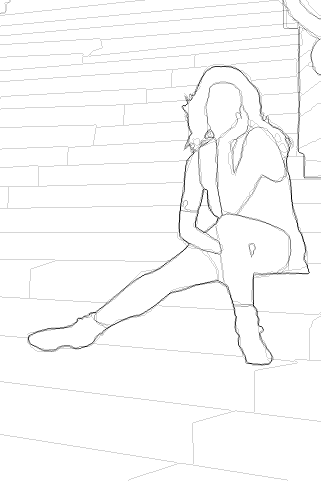}
&\hspace{-.3cm}\includegraphics[width=.1\textwidth,frame]{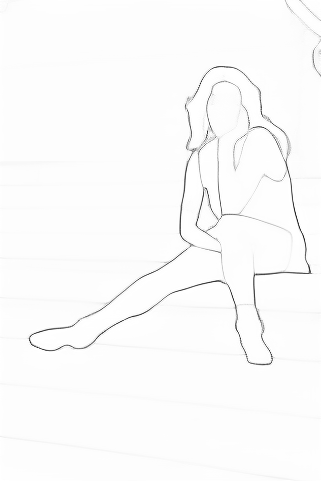}
&\hspace{-.3cm}\includegraphics[width=.1\textwidth,frame]{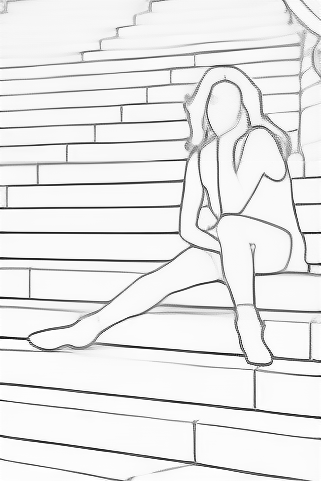}
\vspace{-.06cm}\\

\hspace{-.3cm} Input & \hspace{-.3cm} GT &\hspace{-.3cm}  Ours (0) &\hspace{-.3cm}  Ours (1) \\
\end{tabular}
\caption{Qualitative results of our proposed GED with different edge granularity on the BSDS~\cite{arbelaez2010contour} test dataset.}
\label{Fig1}
\vspace{-15pt}
\end{figure}

\begin{figure*}[!t]
\small
\setlength{\abovecaptionskip}{3pt}
\setlength{\belowcaptionskip}{0pt}
\centering
\begin{tabular}{ccc}
\hspace{-.23cm}
\includegraphics[width=.33\textwidth]{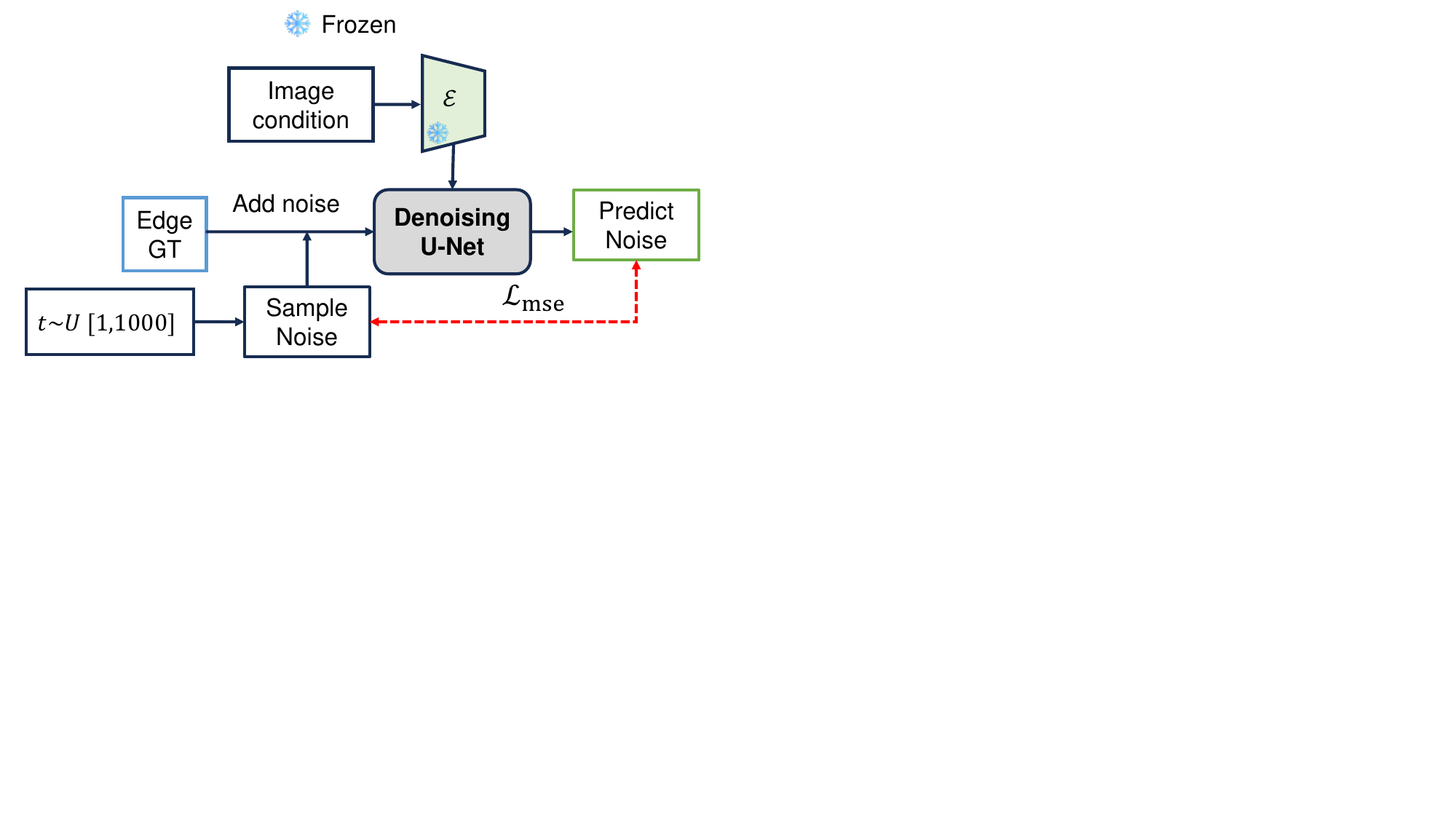}
&\hspace{-.23cm}\includegraphics[width=.33\textwidth]{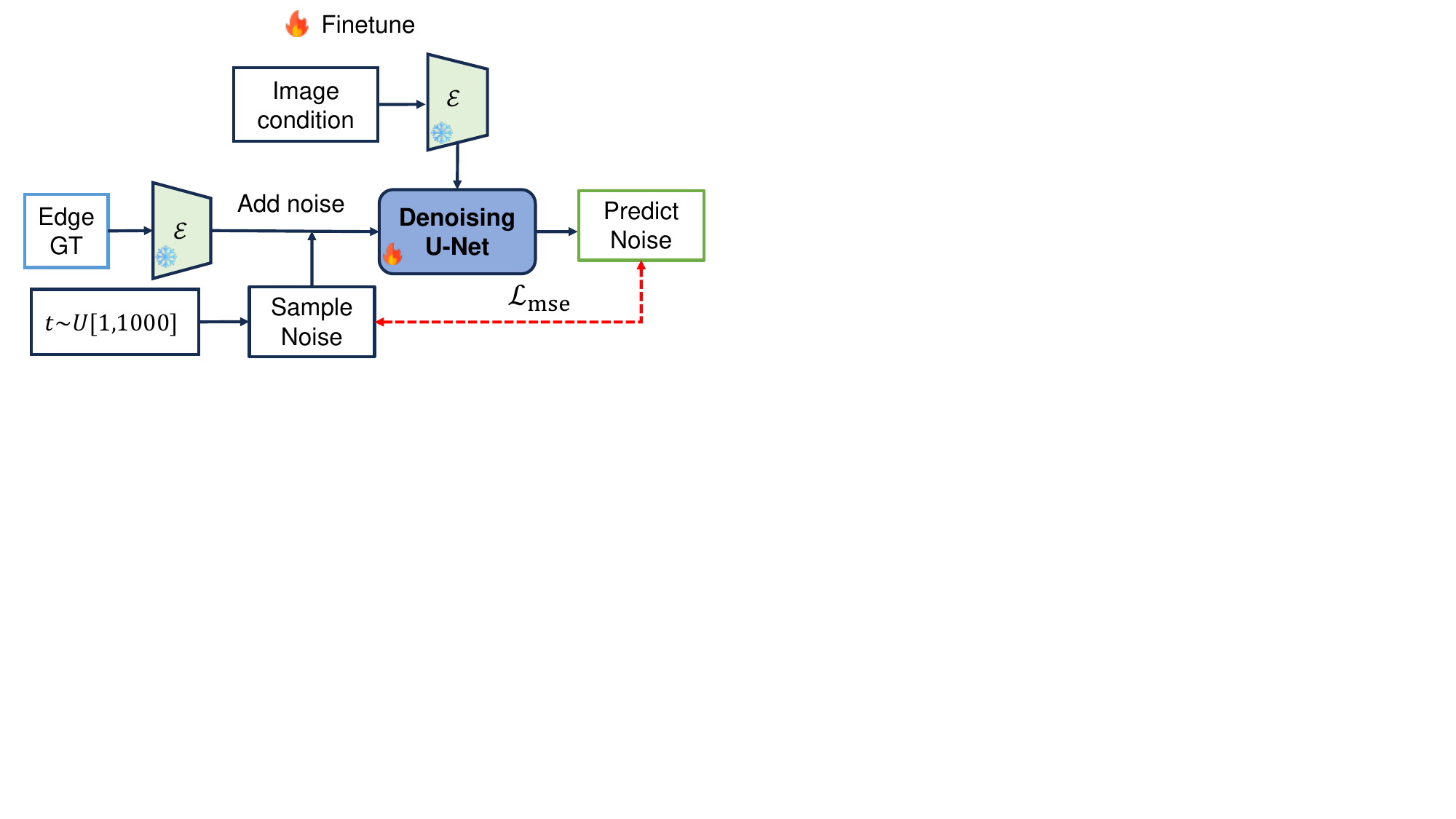}
&\hspace{-.23cm}\includegraphics[width=.33\textwidth]{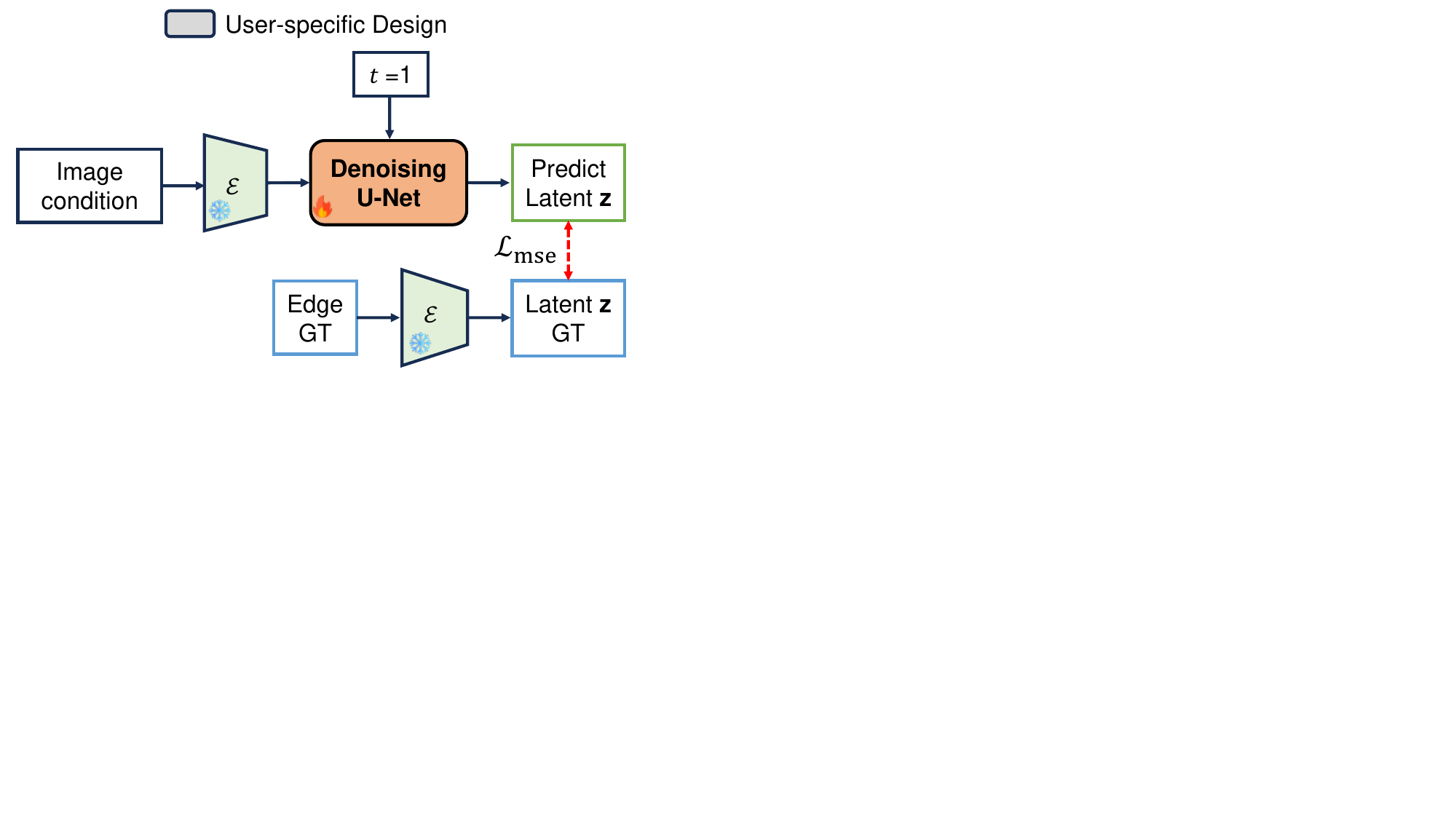}
\vspace{-.06cm}\\

 \hspace{-.23cm} (a) Training from Scratch~\cite{ye2024diffusionedge} &\hspace{-.23cm}  (b) Finetuning Stable Diffusion~\cite{ke2023repurposing} &\hspace{-.23cm}  (c) Predicting Latent Maps (Ours) \\
\end{tabular}
\caption{Different ways of using the diffusion model. (a) is a conditional DDPM model with images as the condition training from scratch; (b) finetunes a conditional latent DDPM model with pre-trained parameters from the stable diffusion model; and (c) aligns edge predictions with ground truth in the latent space by pre-trained DDPM. On the BSDS~\cite{arbelaez2010contour} test dataset, (a) achieves 0.834 ODS and 0.848 OIS with step 5, and 0.833 ODS and 0.846 OIS with step 50. (b) achieves 0.841 ODS and 0.856 OIS with step 5, and 0.832 ODS and 0.848 OIS with step 50. (c) achieves 0.870 ODS and 0.880 OIS with step 1, which has a large margin improvement with less inference time.}
\label{Fig2}
\vspace{-15pt}
\end{figure*}

\section{Introduction}
The goal of edge detection is to localize curves with intensity changes in an image caused by surface discontinuities, reflectance or illumination boundaries~\cite{marr1980theory}, which can not only benefit the performance of the downstream tasks~\cite{chen2023self}, such as semantic segmentation~\cite{wang2022active, zhang2023learning}, depth estimation~\cite{talker2022mind}, and camouflaged object detection~\cite{he2024strategic}, but also help reduce texture bias~\cite{he2023shift} and improve model robustness~\cite{sun2021can, tripathi2023edges, ding2024edge}. 

Deep learning based methods regard edge detection as a discriminative task, which have surpassed human by trying the latest network backbones, including Convolutional Neural Network (CNN)~\cite{xie2015holistically, liu2017richer, he2019bi} and Transformer~\cite{gao2021end, pu2022edter}. However, the generative method has been found strongly aligned with human perception (\ie, shape-bias recognition ability)~\cite{jaini2024intriguing}, but it is not fully explored in edge detection. The recent generative method, DiffusionEdge~\cite{ye2024diffusionedge}, borrows the idea of Denoising Diffusion Probabilistic Models (DDPM)~\cite{ho2020denoising}, which treats the edge prediction as a conditional denoising process from the random noise (Fig.~\ref{Fig2} (a)). However, DiffusionEdge~\cite{ye2024diffusionedge} does not show performance gain as the denoising step increases, and the multi-step denoising increases inference time, which is undesired as edge detection is usually performed in time-sensitive scenarios.

Recent large text-to-image diffusion models (\eg, DALL·E 2~\cite{ramesh2022hierarchical}, stable diffusion~\cite{rombach2022high} and Imagen~\cite{saharia2022photorealistic}) are trained on vast amounts of data across various scenes, thus capable of learning both high-level and low-level visual concepts~\cite{baranchuk2022labelefficient, zhao2023unleashing, wang2024implicit}. Due to their remarkable capability, the rich scene understanding priors can be leveraged to boost the performance of perception tasks. A straight way to utilize the pre-trained text-to-image diffusion models for downstream perception tasks is to finetune the denoising U-Net~\cite{ho2020denoising} to predict ground truth from random noise with images as the condition~\cite{ke2023repurposing,lee2024dmp} (Fig.~\ref{Fig2} (b)). These methods have achieved success in several dense prediction tasks such as depth estimation and semantic segmentation, but the multi-step denoising process can not bring performance benefits and increases the inference cost for the edge detection task. The performances of previous diffusion-based edge detectors decrease as the denoising steps increase from 5 to 50. So a question arises: \textit{How can the edge detection task benefit from a pre-trained text-to-image diffusion model?}

To address the above mentioned issues, we propose a novel generative edge detector, called GED, to explore the rich knowledge derived from the pre-trained diffusion models for edge detection, while does not require the expensive multi-step denoising process or special network design. Specifically, as shown in Fig.~\ref{Fig2} (c), we first encode the images and the corresponding edge maps into the latent space using the variational autoencoder (VAE)~\cite{kingma2013auto} of the stable diffusion model. Then we feed the clean image latent features and image descriptions into the denoising U-Net to predict the latent edge maps. Note that VAE is frozen in the whole process and only the denoising U-Net is finetuned. 

The proposed GED fully adopts the network structure of stable diffusion, but does not require multi-step denoising inference, making it different from Fig.~\ref{Fig2} (a) and Fig.~\ref{Fig2} (b). It takes fewer training steps and achieves a new state-of-the-art (SOTA). Moreover, we also integrate the granularity into the denoising U-Net as one of the conditions due to the necessity of multiple granularity edge detection~\cite{zhou2024muge}, and devise an ordinal regularization to guarantee the granularity of the predictions consistent, which can generate multiple diverse edge maps and further improve the performance (Fig~\ref{Fig1}).

In summary, our contributions are as follows: 
\textbf{(1)} We propose a novel generative edge detector (GED) by exploring the rich priors derived from the pre-trained diffusion model, hence largely reducing training steps.
\textbf{(2)} GED predicts latent edge maps rather than noise, thus avoiding the multi-step denoising process and specific network design. \textbf{(3)} We integrate granularity into the denoising U-Net to obtain diverse and controllable edge predictions, where explicit ordinal regularization is designed to constrain the granularity of the predictions reasonably. \textbf{(4)} Experiments on several edge detection datasets demonstrate the effectiveness of our proposed method.  

\section{Related Work}
\textbf{Edge Detection.} As a fundamental block in computer vision, edge detection is a long-standing task and closely related to many downstream tasks. Early pioneering methods~\cite{canny1986computational, marr1980theory, kittler1983accuracy} learn edges in an unsupervised manner by clustering multi-scale local intensity changes. Traditional supervised methods~\cite{arbelaez2010contour, dollar2014fast, lim2013sketch, martin2004learning} introduce machine learning algorithms and hand-crafted features, such as local brightness, color, and texture cues, to extract edges. DeepEdge~\cite{bertasius2015deepedge} and DeepContour~\cite{shen2015deepcontour} utilize CNN to extract high-level features of candidate patches for the first time. Later CNN-based methods design more powerful network structures and loss functions to extract pixel-wise features for binary classification, such as VGG16~\cite{xie2015holistically, liu2017richer, he2019bi, xuan2022fcl, liu2016learning, maninis2016cob, deng2020deep}, ResNet50~\cite{xu2017AMHNet}, MobileNet~\cite{deng2021learning}, pixel difference convolution (PDC)~\cite{su2021pixel}, dice loss~\cite{deng2018learning}, global structural loss~\cite{deng2020deep}, tracing loss~\cite{huan2021unmixing}, and ranking-based loss~\cite{cetinkaya2024ranked}. EDTER~\cite{pu2022edter} and DiffusionEdge~\cite{ye2024diffusionedge} successfully adopt Transformer~\cite{zheng2021rethinking} and DDPM~\cite{ho2020denoising} for the edge detection task. UAED~\cite{zhou2023treasure} solves the problem from the uncertainty estimation perspective, and MuGE~\cite{zhou2024muge} can generate diverse predictions for different downstream applications. Visual foundation models~\cite{zou2024segment}, such as Segmentation Anything Model (SAM)~\cite{kirillov2023segment} pre-trained on a massive segmentation corpus, also have a great impact on edge detection, \eg,  SCESAME~\cite{yamagiwa2024zero} and EdgeSAM~\cite{yang2024boosting} achieve excellent performance for unsupervised edge detection.

\textbf{Diffusion Models for Dense Prediction Tasks.} A series of outstanding dense prediction works have been successfully proposed driven by the diffusion models. One type of approach treats the diffusion model as a network framework and benefits from the hundreds of denoising steps. Representative works include DDP~\cite{ji2023ddp} for segmentation and depth estimation task, Ground-Diffusion~\cite{li2023open} for open-vocabulary object segmentation task, CCDM~\cite{zbinden2023stochastic} for medical image segmentation task, as well as CamoDiffusion~\cite{chen2024camodiffusion} for camouflaged object detection task. Another line of work exploits the potential of pre-trained diffusion models. DDPM-based segmentation~\cite{baranchuk2022labelefficient} extracts features from different steps and levels to predict pixel-wise semantic labels. LGP~\cite{voynov2023sketch} trains a latent edge predictor to help sketch-guided text-to-image synthesis. VPD~\cite{zhao2023unleashing} designs multi-scale cross-attention maps and task-specific decoders to achieve SOTA on semantic segmentation and depth estimation. TADP~\cite{kondapaneni2023text} and ECoDepth~\cite{patni2024ecodepth} improve VPD by better text-image alignment. 
However, the extracted attention maps are not friendly for generating crisp edges because of low resolution~\cite{wang2023diffusion, khani2024slime}. Marigold~\cite{ke2023repurposing} finetunes the denoising U-Net~\cite{ho2020denoising} to obtain a depth estimator with great generalization and performance. DMP~\cite{lee2024dmp} utilizes LoRA~\cite{hu2022lora} to only fintune attention layers of the denoising U-Net to achieve a generalizable semantic predictor. ILoRA~\cite{du2023intrinsic} discovers scene intrinsics from a wide array of generative models using a LoRA adaptor. LDMSeg~\cite{van2024simple} trains a shallow autoencoder and uses the pre-trained denoising U-Net to obtain latent panoptic segmentation from the random noise. 

Unlike the above efforts, in this paper, we aim to utilize the rich visual-world knowledge contained in the pre-trained generative models to boost the edge detection task without a multi-step denoising process and specific network design. 

\begin{figure*}[!t]
 \centering
 \includegraphics[scale=0.5]{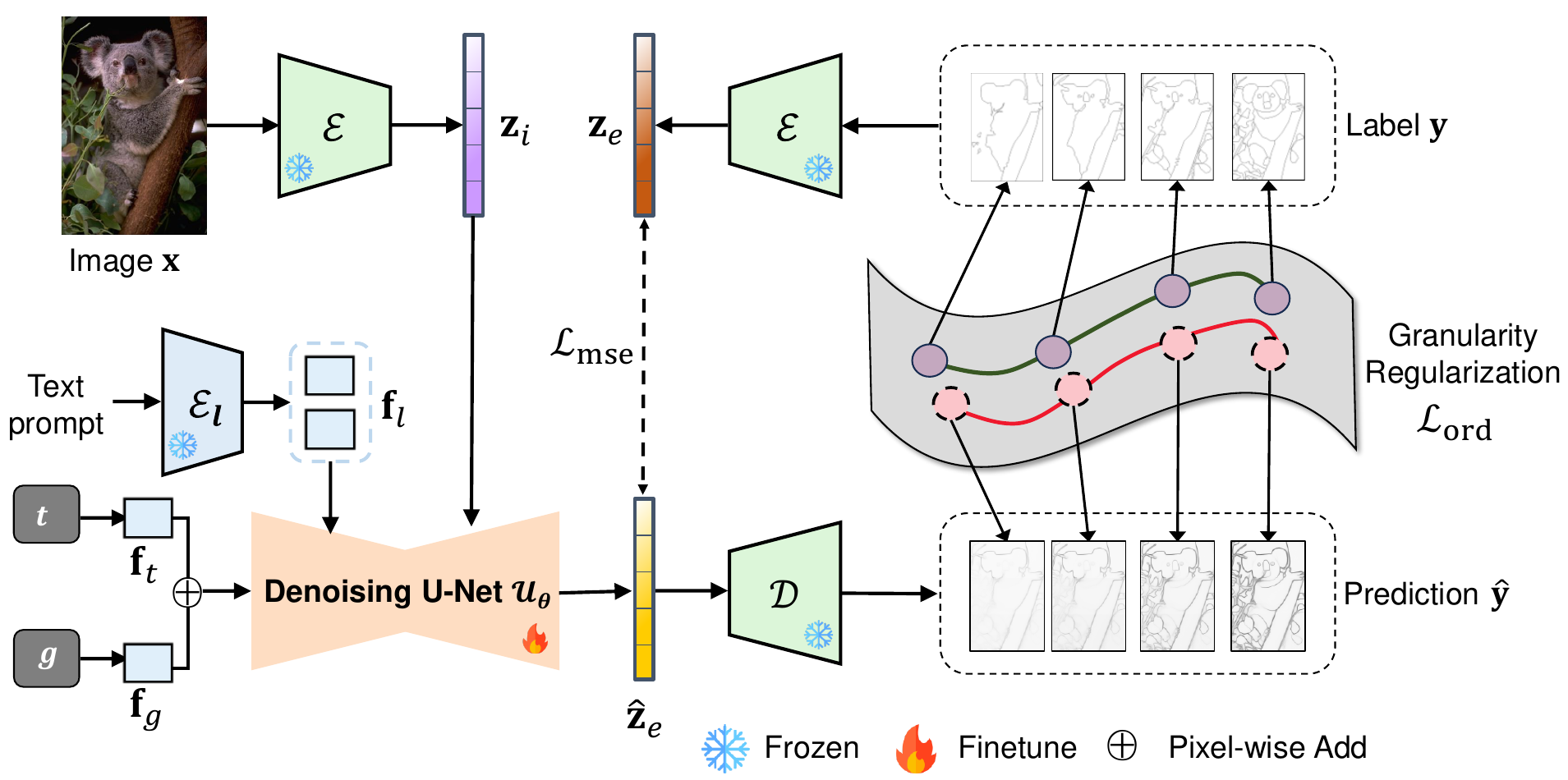}
 \caption{The overall framework of our proposed GED. Given an input image $\mathbf{x}$, its corresponding label sets $\mathbf{y}$, and text prompt $p$, we first obtain granularity $g$ for each label by normalization. Then we extract text features $\mathbf{f}_l$ by text encoder $\mathcal{E}_l$,  latent image feature maps $\mathbf{z}_i$ and latent edge maps $\mathbf{z}_e$ by VAE encoder $\mathcal{E}$. We feed granularity $g$, latent image maps $\mathbf{z}_i$, corresponding time $t=1$ and text features $\mathbf{f}_l$ into the denosing U-Net $\mathcal{U}_\theta$ to obtain predicted latent edge maps $\hat{\mathbf{z}}_e$, which are decoded by $\mathcal{D}$ to the final edge prediction $\hat{\mathbf{y}}$. The granularity $g$ is encoded to $\mathbf{f}_g$ as the same dimension of the time embeddings $\mathbf{f}_t$ by two fully connected layers and then pixel-wise added to the time embeddings $\mathbf{f}_t$. We also add explicit regularizations to ensure relative ordinal granularity relationships.}
 \vspace{-10pt}
 \label{Fig3}
\end{figure*}

\section{Methodology}
We describe how to fully utilize the pre-trained large stable diffusion for the edge detection task. First, we introduce stable diffusion in Section~\ref{pr}. Then, we depict our design for adapting stable diffusion to the edge detection task in Section~\ref{fr}. Finally, we formulate the loss function in Section~\ref{nt} and detail the finetuning process in Section~\ref{td}.

\subsection{Preliminaries}
\label{pr}
Stable diffusion~\cite{rombach2022high} is a latent text-to-image diffusion model~\cite{rombach2022high} capable of generating photo-realistic images given any text input, which is trained on LAION-5B~\cite{schuhmann2022laion} dataset including 5.85 billion CLIP-filtered image-text pairs. The latent diffusion model conducts DDPM in the latent space due to GPU and inference limitations. Stable diffusion contains a text encoder ($\mathcal{E}_l$), a VAE (including encoder $\mathcal E$ and decoder $\mathcal D$), and a denoising U-Net ($\mathcal{U}_\theta$), where $\theta$ indicates the trainable model parameters. Given a text prompt $p$, the text encoder extracts the text features and feeds them into $\mathcal{U}_\theta$ to generate lifelike images. VAE~\cite{kingma2013auto} encoder converts the image from the pixel space $\mathbf{x}$ to the latent space $\mathbf{z}$. Denoising U-Net is a DDPM, which contains forward and reverse processes. Forward process corrupts the data by incrementally adding Gaussian noise to each sample $\mathbf{z}_0$ in $T$ time step, producing a sequence of noisy samples $\mathbf{z}_1, \ldots, \mathbf{z}_T$ as latents of the same dimensionality as the original sample $\mathbf{z}_0$. The noise is related to the time step and is pre-defined, so it does not need to be learned in the forward process. The corresponding reverse process removes the noise step by step so that new samples can be generated from the data distribution starting from random noise. At each step, the learnable neural network $\mathcal{U}_{\theta}(\mathbf{z}_t, t, \mathcal{E}_l(p))$ is trained to predict the current time-aware noise. 
 
\subsection{Framework}
\label{fr}
Labeling the edge map of an image is subjective due to the complexity of scenes. Therefore, most of the datasets~\cite{arbelaez2010contour,mely2016systematic} are constructed by multiple annotators, which brings diverse edge maps. Due to application-dependent requirements, diverse edge predictions with controllable granularity are necessary~\cite{zhou2024muge}. So we follow MuGE~\cite{zhou2024muge} to provide a multi-granularity edge detector. Formally, given an input image $\mathbf{x} \in \mathbb{R}^{H \times W \times 3}$, the corresponding multiple edge maps are represented as $\{\mathbf{y}^{k}\}_{k=1}^{K}$, where $\mathbf{y}^{k}\in {\{0,1\}}^{H \times W }$ is the $k\textrm{-th}$ edge map and $K$ is the total number of edge maps. To fully utilize all annotated labels and introduce granularity, we arrange and combine $K$ labels without considering order to get $\sum\nolimits_{n=2}^{K} C_K^n$ labels.
For simplicity, we choose the normalization strategy in MuGE~\cite{zhou2024muge} to denote granularity $g$, which calculates the number of edge pixels in each map and then normalizes each edge map according to the maximum and minimum values, resulting in a scalar between 0 and 1 (\ie, small for rough objects and large for detailed edges).

\textbf{Overview.} Fig.~\ref{Fig3} depicts the overview of the proposed GED, which is built upon the pre-trained Stable Diffusion v2.1~\cite{rombach2022high}. GED consists of an image encoder ($\mathcal{E}$), an image decoder ($\mathcal{D}$), a text encoder ($\mathcal{E}_l$) and a denoising U-Net ($\mathcal{U}_\theta$). Only the denoising U-Net is finetuned with the image-edge pairs. The initial denoising U-Net of the stable diffusion can be represented as $\mathcal{U}_\theta(\mathbf{z}_t, t,\mathcal{E}_l(p) )$, where $\mathbf{z}_t$ means the noisy ground truth at the current time $t$, and $p$ is the given text prompt. By contrast, we feed the clean image in the latent space, image caption, the corresponding time step $t$ and the desired granularity $g$ to the denoising U-Net. For the input image, instead of predicting noise, the latent edge map is directly predicted, which avoids the time-consuming multi-step denoising process.

\textbf{Input Encoding.} For the input image $\mathbf{x}$, we first extract the latent image features $\mathbf{z}_i = \mathcal{E}(\mathbf{x})\in \mathbb{R}^{H/8 \times W/8 \times 4}$ by the VAE encoder $\mathcal{E}$. The utilized text prompt in this paper is generated by inputting the image into Minigpt-4~\cite{zhu2022minigpt4} with the prompt `Describe this picture without imagination', and the pre-trained text encoder $\mathcal{E}_l$ is used to extract text features $\mathbf{f}_l\in \mathbb{R}^{77\times 1024}$.

\textbf{Granularity Fusion.} To integrate the edge granularity into the stable diffusion, we use two fully connected layers (FC) to convert $g$ to a vector $\mathbf{f}_g$ with the same dimension of the time feature $\mathbf{f}_t$ and add them pixel-wise denoted as $\mathbf{f}_t \oplus\mathbf{f}_g$. The input of denoising U-Net is revised as follows:

\begin{equation}
\begin{gathered}
\hat{\mathbf{z}}_e = \mathcal{U}_\theta(\mathbf{z}_i, \mathbf{f}_t \oplus\mathbf{f}_g, \mathcal{E}_l(p)),
\\
\mathbf{f}_t
=\textrm{time\_embedding}(t), \ \mathbf{f}_g = \textrm{FC}(g).
\end{gathered}
\end{equation}

\textbf{Edge Map Prediction.} After getting the latent image features, text features, time step, and granularity as input, we only finetune the denoising U-Net to predict the corresponding latent edge map, which can be further fed into the decoder $\mathcal{D}$ to generate the final edge prediction $\hat{\mathbf{y}}$.

\subsection{Network Training}
\label{nt}
We devise two loss functions to finetune the pre-trained denoising U-Net: latent edge map alignment loss and granularity regularization loss.

\textbf{Latent Edge Map Alignment Loss.}
Given corresponding edge maps $\{\mathbf{y}^{k}\}_{k=1}^{K}\in {\{0,1\}}^{H \times W }$, we first utilize VAE encoder to extract latent edge maps of the ground truth $\mathbf{z}_e = \mathcal{E}(\mathbf{y})\in \mathbb{R}^{H/8 \times W/8 \times 4}$. To obtain the predicted edge maps, we make the output of denoising U-Net to predict latent edge maps, which can be denoted as $\hat{\mathbf{z}}_e$. To align predicted and ground truth latent edge maps, we minimize the mean square error (MSE) between them:
\begin{equation}
\mathcal{L}_\text{mse} = \|\hat{\mathbf{z}}_e - \mathbf{z}_e\|^2_2.
\end{equation}

\textbf{Granularity Regularization Loss.}
There exists an obvious ordinal relationship between granularity, so we design an explicit regularization to keep relative granularity relationships among the diverse predictions.
For every step iteration, we input an image $\mathbf{x}$ and multiple edge maps with different granularities, then we restrict the output granularity consistent with the ground truth granularity. Specifically, we ensure the relative distances between predicted latent edge maps and ground truth latent edge maps in the latent space. Furthermore, for the granularity information, we constrain the predicted granularity is aligned with ground truth granularity, where the predicted granularity $\hat{g}$ follows the same normalization of ground truth $g$. Thus, we minimize the two MSE losses as follows:
\begin{equation}
\mathcal{L}_\text{ord} = \|d(\hat{\mathbf{z}}^i_e, \hat{\mathbf{z}}^j_e) - d({\mathbf{z}}^i_e, {\mathbf{z}}^j_e)\|_2^2 + \|\hat{g}- g\|_2^2,
\end{equation}
where 
$d(\hat{\mathbf{z}}^i_e, \hat{\mathbf{z}}^j_e)$ means the Euclidean distance of predicted latent edge maps among different granularities, and $d({\mathbf{z}}^i_e, {\mathbf{z}}^j_e)$ means the Euclidean distance latent edge label maps. 

\textbf{Total Loss.} In summary, the total loss function is 
\begin{equation}  \mathcal{L} = \mathcal{L}_\text{mse} + \mathcal{L}_\text{ord}.
\end{equation}

\subsection{Finetuning Details}
\label{td}
We first finetune all parameters of the denoising U-Net ($\theta$) and find the last two stages of the denoising U-Net decoder and time embedding layers exhibit distinct changes in parameter values. So in the final experimental setting, we only finetune these layers while keeping the others frozen, which yields results similar to those of full-parameter finetuning, but significantly reduces the training cost. 

\section{Experiments}
\subsection{Experiment Settings}
\label{set}

\textbf{Datasets and Evaluation Metrics.} We employ four commonly-used edge detection datasets for evaluation, \ie,  BSDS~\cite{arbelaez2010contour}, Multicue~\cite{mely2016systematic}, NYUD~\cite{silberman2012indoor} and BIPED~\cite{poma2020dense}. \textbf{BSDS500} contains 500 natural scene images sized at $321\times 481$, of which 200 are used for training, 100 for verification, and 200 for testing. Each image involves 4 to 9 annotators for labeling. \textbf{Multicue} contains 100 left-view challenging natural scene images, which are collected for the study of boundary and edge detection. Each scene image is annotated with edges by 6 annotators and boundaries by 5 annotators. We randomly select 80 images for training and the remaining for testing. The scores of three independent trials are averaged as the final results. \textbf{NYUD} is proposed for indoor scene parsing, including 1449 image-depth-segmentation maps sized at $640\times 480$, where 795 images are used for training and 654 images for testing. Edge labels are extracted from the segmentation maps. \textbf{BIPED} contains 250 outdoor images with a $1280\times 780$ resolution, where 200 images are used for training and 50 images for testing. Each image is annotated by one expert and cross-checked. For data augmentation, we apply random crop, scaling and flipping on the BSDS dataset, and random crop and flipping for the other datasets following DiffusionEdge~\cite{ye2024diffusionedge}.

We report the commonly used metrics for comparison, including optimal dataset
scale (ODS), optimal image scale (OIS), and Average Precision (AP) based on precision and recall. We also calculate the best ODS and OIS to show the results of multiple granularity edge detection~\cite{zhou2024muge}.

\textbf{Implementation Details.} We utilize Stable Diffusion v2.1~\cite{rombach2022high} (SD2.1) as the pre-trained diffusion model. Images are randomly cropped to $320\times 320$ for training. All parameters are updated by AdamW optimizer~\cite{loshchilov2018decoupled} with an attenuated learning rate from 5e-5 to 5e-6. Experiments are conducted on a single RTX 3090 with Pytorch~\cite{paszke2019pytorch}. Training with a batchsize of 4 costs about 12G GPU memory, and a batchsize of 16 is achieved by gradient accumulation. The total iteration steps are set to 5000.

\subsection{Comparison with State-of-the-Art}
\begin{table*}[t]
\setlength{\abovecaptionskip}{5pt}
\caption{Results on the \textbf{BSDS500}~\cite{arbelaez2010contour} and \textbf{Multicue}~\cite{mely2016systematic} test dataset. All results are obtained under single-scale inference. $M$ means the number of predictions, and $M=11$ means we generate 11 edge maps with granularity $\{0, 0.1, 0.2, 0.3, 0.4, 0.5, 0.6, 0.7, 0.8, 0.9, 1\}$. The best two results are denoted as \BEST{red} and \SBEST{blue} respectively, and the same for other tables.}
\centering
\renewcommand\tabcolsep{5pt}
\begin{tabular}{l|c|ccc|ccc|ccc}
     \Xhline{1px}
   \multicolumn{1}{l}{\multirow{2}*{Method}}\vline&\multicolumn{1}{c}{\multirow{2}*{Backbone}}\vline&
    \multicolumn{3}{c}{\multirow{1}*{BSDS}}\vline&\multicolumn{3}{c}{\multirow{1}*{Multicue Edge}}\vline&
    \multicolumn{3}{c}{\multirow{1}*{Multicue Boundary}}
    \\
    \cline{3-11}
    &&ODS & OIS& AP 
    &ODS & OIS& AP 
    &ODS & OIS& AP \\
    \hline
Human&-& 0.803&0.803&-&0.750&-&-&0.760&-&-\\
Canny$_\text{ TPAMI'86}$~\cite{canny1986computational}&-& 0.611&0.676&0.520&-&-&-&-&-&-\\
gPb-UCM$_\text{ TPAMI'10}$~\cite{arbelaez2010contour}&-&0.729&0.755&0.745&-&-&-&-&-&-\\
DeepContour$_\text{ CVPR'15}$~\cite{shen2015deepcontour}&AlexNet&0.757     &0.776&0.790&-&-&-&-&-&-\\
HED$_\text{ ICCV'2015}$~\cite{xie2015holistically}&VGG16&0.788&0.808&0.840&0.851&0.864&0.890&0.814&0.822&0.869\\
Deep Boundary$_\text{ ICLR'15}$~\cite{kokkinos2015pushing}&VGG16&0.789&0.811&0.789&-&-&-&-&-&-\\
RCF$_\text{ CVPR'17}$~\cite{liu2017richer}&VGG16&0.798&0.815&-&0.857&0.862&-&0.817&0.825&-\\
BDCN$_\text{ CVPR'19}$~\cite{he2019bi}&VGG16&0.806&0.826&0.847&0.891&0.898&0.935&0.836&0.846&0.893\\
PiDiNet$_\text{ ICCV'21}$~\cite{su2021pixel}&PDC&0.789&0.803&-&0.874&0.878&-&0.818&0.830&-\\
LDC$_\text{ ACM'21}$~\cite{deng2021learning}&MobileNet&0.799&0.816&0.837&0.881 &0.893 &-&0.839 &0.853 &-\\
EDTR$_\text{ ICONIP'21}$~\cite{gao2021end}&Transformer&0.820&0.839&0.861&-&-&-&-&-&-\\
EDTER$_\text{ CVPR'22}$~\cite{pu2022edter}&Transformer&0.824&0.841&0.880&0.894 &0.900 &0.944&0.861&0.870&0.919\\
FCL-Net$_\text{ NN'22}$~\cite{xuan2022fcl}&VGG16&0.807&0.822&-&0.875&0.880&-&0.834&0.840&-\\
LRCED$_\text{ TIP'23}$~\cite{ye2023delving}&PDC&0.771&0.782&-&0.892&0.907&-&-&-&-\\
UAED$_\text{ CVPR'23}$~\cite{zhou2023treasure}&EfficientNet&0.829&0.847&0.892&0.895&0.902&0.949&0.864&0.872&0.927\\
PEdger$_\text{ ACM'23}$~\cite{fu2023practical}&Recurrent&0.823&0.841&-&-&-&-&-&-&-\\
DiffusionEdge$_\text{ AAAI'24}$~\cite{ye2024diffusionedge} &DDPM&0.834&0.848&0.815&0.904&0.909&-&-&-&-\\
EdgeSAM-U$_\text{ TII'24}$~\cite{yang2024boosting}&SAM&0.813&0.825&0.838&-&-&-&-&-&-\\
EdgeSAM-S$_\text{ TII'24}$~\cite{yang2024boosting}&SAM&0.838&0.852&0.893&-&-&-&-&-&-\\
MuGE$_\text{ CVPR'24}$~\cite{zhou2024muge} &EfficientNet&0.831&0.847&0.886&0.896&0.900&0.948&0.866&0.875&\SBEST{0.927}\\
MuGE (M=11)$_\text{ CVPR'24}$~\cite{zhou2024muge} &EfficientNet&0.850&0.856&0.896&0.898&0.900&0.950&0.875&0.879&\BEST{0.932}\\
\hline
GED (Ours)&SD2.1&\SBEST{0.870}&\SBEST{0.880} &\SBEST{0.907}&\SBEST{0.910}&\SBEST{0.917}&\SBEST{0.961}&\SBEST{0.878}&\SBEST{0.885}&0.919\\
GED (M=11)&SD2.1&\BEST{{0.886}}&\BEST{0.891} &\BEST{{0.919}}&\BEST{0.914}&\BEST{0.918}&\BEST{0.964}&\BEST{0.882}&\BEST{0.886}&0.919\\
     \Xhline{1px}
\end{tabular}
\label{table1}
\vspace{-10pt}
\end{table*}

We compare our proposed GED with some edge detectors, including Canny~\cite{canny1986computational}, gPb-UCM~\cite{arbelaez2010contour}, DeepContour~\cite{shen2015deepcontour}, HED~\cite{xie2015holistically}, Deep Boundary~\cite{kokkinos2015pushing}, AMH-Net~\cite{xu2017AMHNet}, RCF~\cite{liu2017richer}, 
CED~\cite{wang2017deep}, BDCN~\cite{he2019bi}, DexiNed~\cite{poma2020dense}, LDC~\cite{deng2021learning}, PiDiNet~\cite{su2021pixel}, LPCB~\cite{kim2022learning}, EDTER~\cite{pu2022edter}, FCL-Net~\cite{xuan2022fcl}, UAED~\cite{zhou2023treasure}, PEdger~\cite{fu2023practical}, LRCED~\cite{ye2023delving}, DiffusionEdge~\cite{ye2024diffusionedge}, EdgeSAM~\cite{yang2024boosting}, RankED~\cite{cetinkaya2024ranked}, and MuGE~\cite{zhou2024muge}.

\begin{figure*}
\small
\centering
\renewcommand\arraystretch{1}
\renewcommand\tabcolsep{3pt}
\begin{tabular}{cccccc}

\includegraphics[width=.15\textwidth,frame]{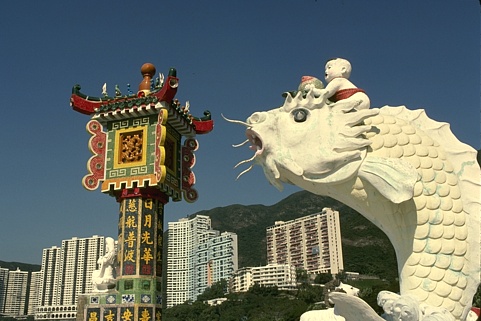} 
&\includegraphics[width=.15\textwidth,frame]{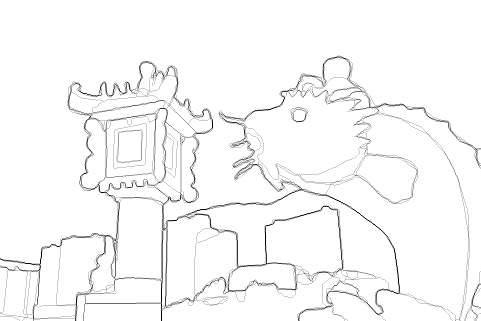}
&\includegraphics[width=.15\textwidth,frame]{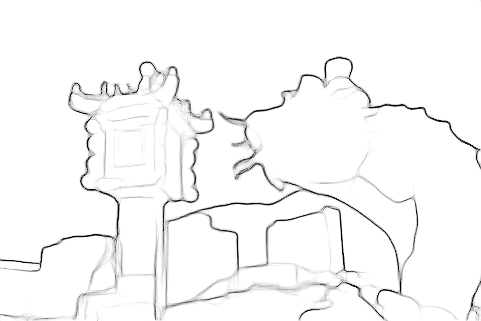}
&\includegraphics[width=.15\textwidth,frame]{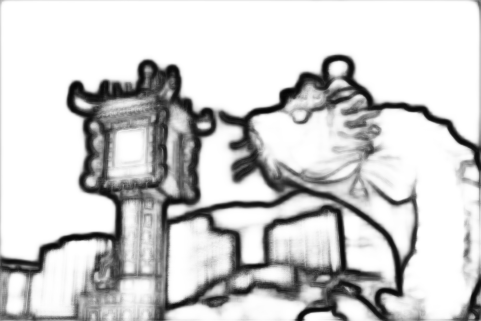}
&\includegraphics[width=.15\textwidth,frame]{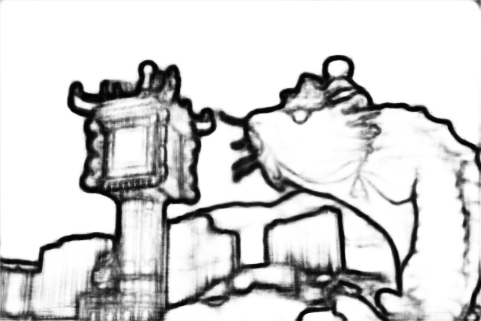}
&\includegraphics[width=.15\textwidth,frame]{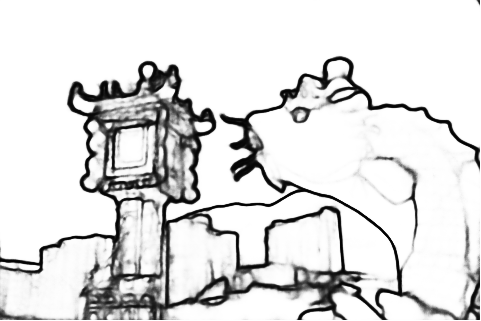}
\\
Input&  GT&  DiffEdge~\cite{ye2024diffusionedge}&RCF~\cite{liu2017richer}& BDCN~\cite{he2019bi}& EDTER~\cite{pu2022edter} \\
\includegraphics[width=.15\textwidth,frame]{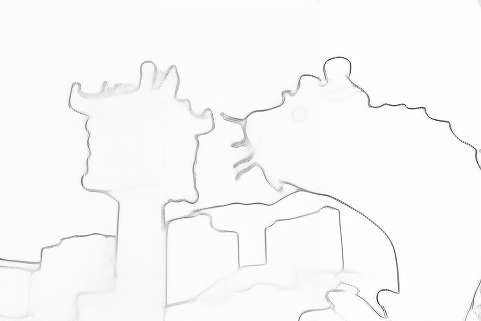} 
&\includegraphics[width=.15\textwidth,frame]{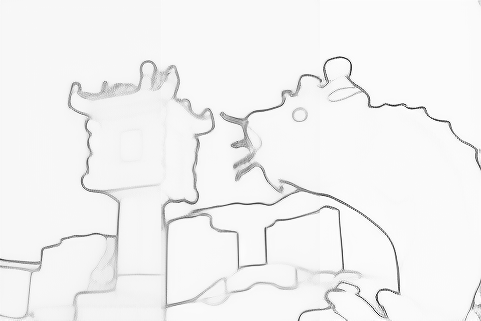}
&\includegraphics[width=.15\textwidth,frame]{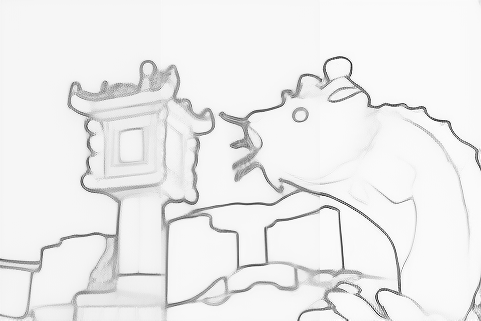}
&\includegraphics[width=.15\textwidth,frame]{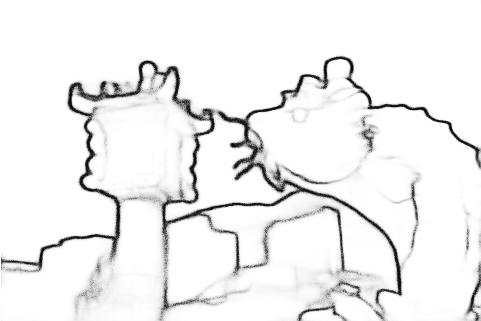} 
&\includegraphics[width=.15\textwidth,frame]{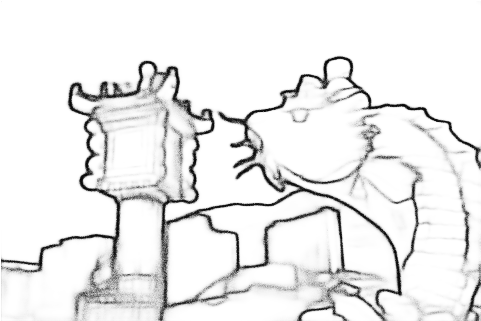}
&\includegraphics[width=.15\textwidth,frame]{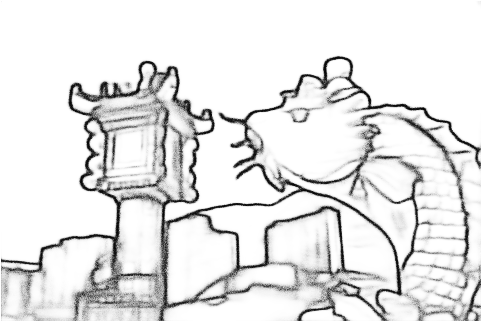}
\\
Ours (0) & Ours (0.5) & Ours (1)&MuGE (0)~\cite{zhou2024muge} &  MuGE (0.5)~\cite{zhou2024muge} &  MuGE (1)~\cite{zhou2024muge}  \\

\includegraphics[width=.15\textwidth,frame]{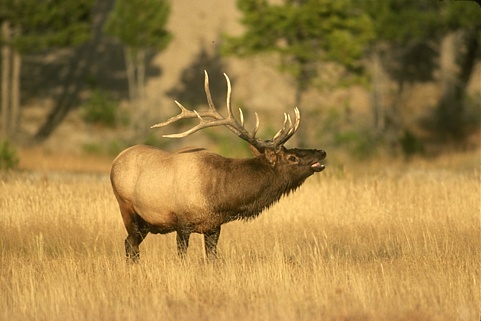} 
&\includegraphics[width=.15\textwidth,frame]{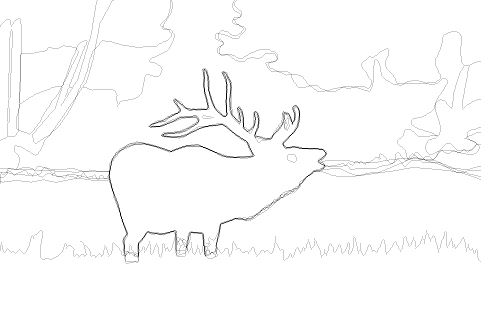}
&\includegraphics[width=.15\textwidth,frame]{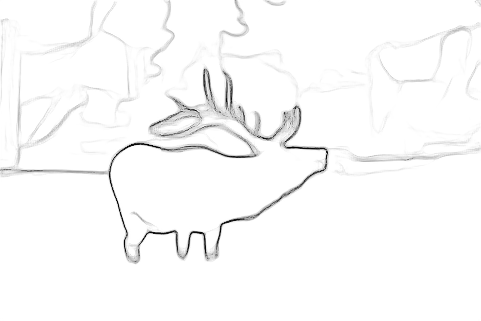}
&\includegraphics[width=.15\textwidth,frame]{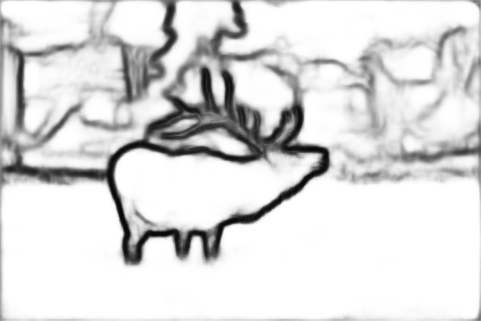}
&\includegraphics[width=.15\textwidth,frame]{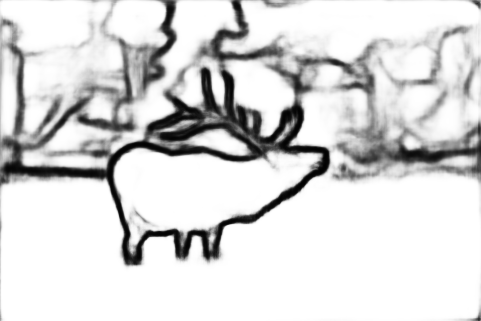}
&\includegraphics[width=.15\textwidth,frame]{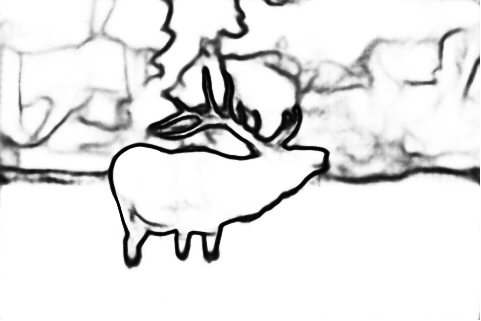}
\\
Input&  GT&  DiffEdge~\cite{ye2024diffusionedge}&RCF~\cite{liu2017richer}& BDCN~\cite{he2019bi}& EDTER~\cite{pu2022edter} \\
\includegraphics[width=.15\textwidth,frame]{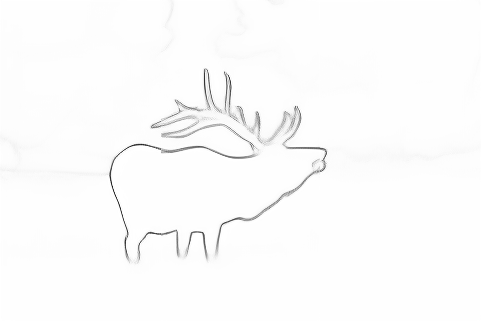} 
&\includegraphics[width=.15\textwidth,frame]{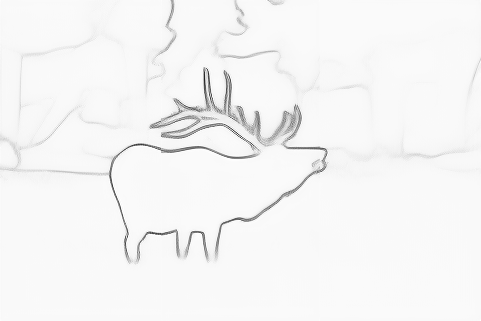}
&\includegraphics[width=.15\textwidth,frame]{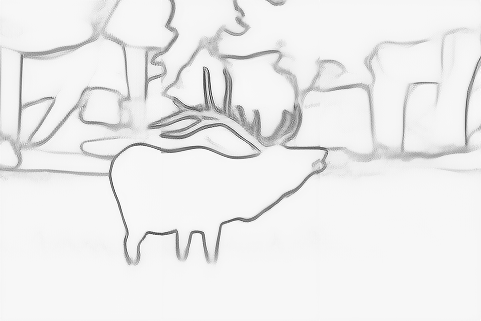}
&\includegraphics[width=.15\textwidth,frame]{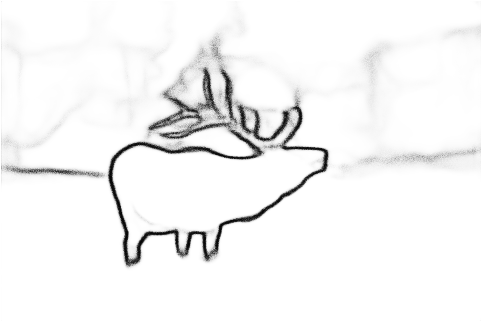} 
&\includegraphics[width=.15\textwidth,frame]{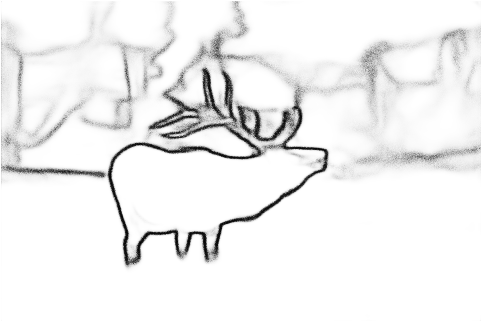}
&\includegraphics[width=.15\textwidth,frame]{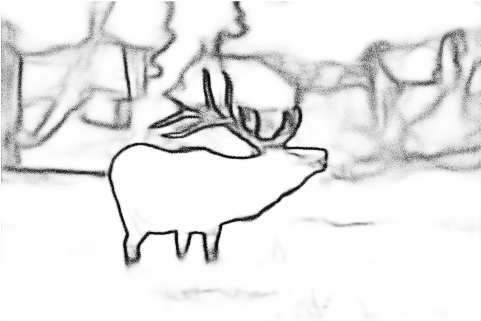}
\\
Ours (0) & Ours (0.5) & Ours (1)&MuGE (0)~\cite{zhou2024muge} &  MuGE (0.5)~\cite{zhou2024muge} &  MuGE (1)~\cite{zhou2024muge}  \\
\end{tabular}
\caption{Qualitative comparisons on challenging samples in the BSDS500 test set. Note that MuGE and our proposed GED produce diverse results with edge granularity of 0, 0.5, and 1, respectively.}
\label{Fig4}
\vspace{-10pt}
\end{figure*}

\textbf{Results on the BSDS Dataset.} Table~\ref{table1} concludes the results of the BSDS test dataset. The previous best two models are DiffusionEdge~\cite{ye2024diffusionedge} and EdgeSAM~\cite{yang2024boosting}, where EdgeSAM-U denotes the unsupervised setting and EdgeSAM-S denotes the supervised setting.  EdgeSAM-S achieves excellent performance by employing the pre-trained SAM trained on plenty of image-segmentation pairs. In contrast, our proposed GED achieves a new SOTA by the pre-trained stable diffusion model, especially improving ODS from 0.838 to 0.870, OIS from 0.852 to 0.880, and AP from 0.893 to 0.907. As shown in Fig.~\ref{Fig4},  both DiffusionEdge and our method can generate crisp edge predictions, because generative models avoid the usage of weighted binary cross-entropy loss, which gives the edge pixels larger weights due to the heavy imbalance problem~\cite{deng2018learning}. Compared with DiffusionEdge, our network structure decreases the trainable parameters from 225M to 116M, and the iteration step from 100,000 to 5,000. Meanwhile, we dramatically improve ODS, OIS and AP metrics by 3.6\%, 3.2\% and 9.2\% without multi-step inference. 

In addition, the integration of granularity information enables us to predict diverse edge maps. Therefore, following MuGE~\cite{zhou2024muge}, we generate 11 predictions with granularity $\{0, 0.1, 0.2, \cdots, 0.9, 1\}$. Accordingly, we can obtain the best ODS score of 0.886, and the best OIS score of 0.891, which improves MuGE~\cite{zhou2024muge} by 3.6\% and 3.5\% in ODS and OIS. The qualitative results in Fig.~\ref{Fig4} can also demonstrate the superiority of our proposed method GED. Compared to DiffusionEdge, GED has controllable and diverse predictions, and the results with granularity capture more desired details. Compared with MuGE, our predicted maps are more crisp and consistent with ground truths, which largely alleviates the false positive detection issue in CNN-based methods.

\begin{table*}[t] 
\setlength{\abovecaptionskip}{5pt}
\caption{Results on the \textbf{NYUD}~\cite{silberman2012indoor} and \textbf{BIPED}~\cite{poma2020dense} test  dataset.}
\centering
\renewcommand\arraystretch{1.1}
\renewcommand\tabcolsep{10pt}
\begin{tabular}{l|c|ccc|ccc}
\Xhline{1px}
\multicolumn{1}{l|}{\multirow{2}*{Method}}&\multicolumn{1}{c|}{\multirow{2}*{Backbone}}&
\multicolumn{3}{c|}{\multirow{1}*{NYUD}} &\multicolumn{3}{c}{\multirow{1}*{BIPED}}
\\
\cline{3-8}
&&ODS & OIS& AP&ODS & OIS& AP \\
\hline
HED$_\text{ ICCV'2015}$~\cite{xie2015holistically}&VGG16&0.720&0.734&0.734&0.829&0.847&0.869\\
RCF$_\text{ CVPR'17}$~\cite{liu2017richer}&VGG16&0.729&0.742&-&0.843&0.859&0.882\\
AMH-Net$_\text{ NIPS'17}$~\cite{xu2017AMHNet}&ResNet50& 0.744&0.758&0.765&-&-&-\\
CED$_\text{ CVPR'17}$~\cite{wang2017deep,poma2020dense}&VGG16 &-&-&-&0.795&0.815&0.830\\
BDCN$_\text{ CVPR'19}$~\cite{he2019bi}&VGG16&0.748&0.763&0.770&0.839&0.854&0.887\\
DexiNed$_\text{ WACV'20}$~\cite{poma2020dense}&Inception&-&-&-&0.859&0.867&\SBEST{0.905}\\
PiDiNet$_\text{ ICCV'21}$~\cite{su2021pixel}&PDC&0.733&0.747&-&0.868&0.876&-\\
EDTER$_\text{ CVPR'22}$~\cite{pu2022edter}&Transformer&0.774&0.789&0.797&0.893&0.898&-\\
DiffusionEdge$_\text{ AAAI'24}$~\cite{ye2024diffusionedge} &DDPM&0.761&0.766&0.508&\BEST{0.899}&\SBEST{0.901}&0.642\\
EdgeSAM-U$_\text{ TII'24}$~\cite{yang2024boosting}&SAM&0.743&0.760&0.768&-&-&-\\
EdgeSAM-S$_\text{ TII'24}$~\cite{yang2024boosting}&SAM&\SBEST{0.783}&\SBEST{0.797}&\SBEST{0.805}&-&-&-\\
RankED$_\text{ CVPR'24}$~\cite{cetinkaya2024ranked}&Transformer&0.780&0.793&\BEST{0.826}&-&-&-\\
GED (Ours)&SD2.1&\BEST{0.800}&\BEST{0.809}&0.798&\SBEST{0.896}&\BEST{0.906}&\BEST{0.934}\\
\Xhline{1px}
\end{tabular}
\label{table2}
\vspace{-10pt}
\end{table*}

\begin{table}[ht]
\setlength{\abovecaptionskip}{5pt}
\centering
\small
\renewcommand\arraystretch{1}
\renewcommand\tabcolsep{1pt}
\caption{The ablation study on the BSDS test set.}
\begin{tabular}{cccc|ccc}
\toprule
Generative&Text&Granularity& Regularzation&ODS&OIS& AP \\
\midrule
\checkmark&&&&0.859&0.875&0.910\\
\checkmark&\checkmark&&&0.861&0.877&0.907\\
\checkmark&\checkmark&\checkmark&&0.865&0.880&0.909\\
\checkmark&\checkmark&\checkmark&\checkmark&0.870&0.880&0.907\\
\bottomrule
\end{tabular}
\label{tab:ablation}
\vspace{-10pt}
\end{table}

\begin{table}[ht]
\setlength{\abovecaptionskip}{5pt}
\centering
\caption{Comparison of network design.}
\renewcommand\arraystretch{1.11}
\renewcommand\tabcolsep{5pt}
\begin{tabular}{c|c|ccc}
\toprule
Methods & Steps & ODS & OIS & AP \\
\hline
\multicolumn{1}{c|}{\multirow{2}*{\shortstack{DiffusionEdge  \\(Fig.~\ref{Fig2} (a))}}} &5&0.834&0.848&0.815\\
 & 50 & 0.833&0.846&-\\
\midrule
\multicolumn{1}{c|}{\multirow{2}*{\shortstack{Marigold  \\(Fig.~\ref{Fig2} (b))}}}& 5 &0.841&0.856&0.906\\
& 50 &0.832&0.848&0.899\\
\bottomrule
\end{tabular}
\label{tab:step}
\vspace{-20pt}
\end{table}

\textbf{Results on the Multicue Dataset.} We conduct experiments on the Multicue edge and boundary datasets, as shown in Table~\ref{table1}. DiffusionEdge obtains 0.904 ODS and 0.909 OIS on the Multicue edge dataset. Our proposed GED further improves the performance. For edge detection, we obtain 0.910, 0.917 and 0.961 in ODS, OIS and AP. For boundary detection, we obtain the best scores of 0.878 and 0.885 in ODS and OIS. By introducing multiple predictions, our performance is further improved.

\begin{figure*}
\small
\centering
\renewcommand\arraystretch{1}
\renewcommand\tabcolsep{3pt}
\begin{tabular}{ccccc}
\rotatebox{90}{\qquad NYUD}&
\includegraphics[width=.22\textwidth,frame]{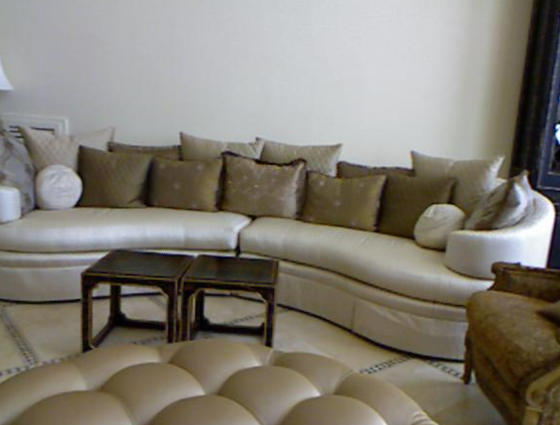} 
&\includegraphics[width=.22\textwidth,frame]{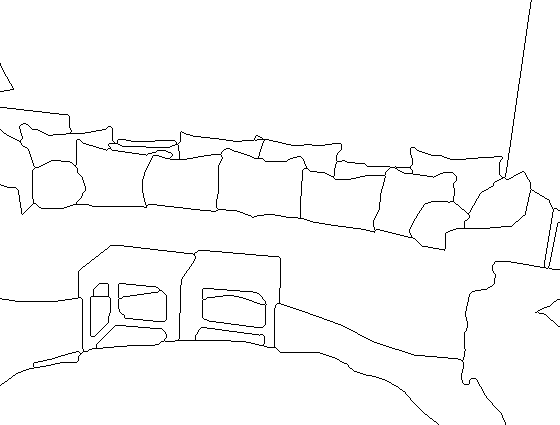}
&\includegraphics[width=.22\textwidth,frame]{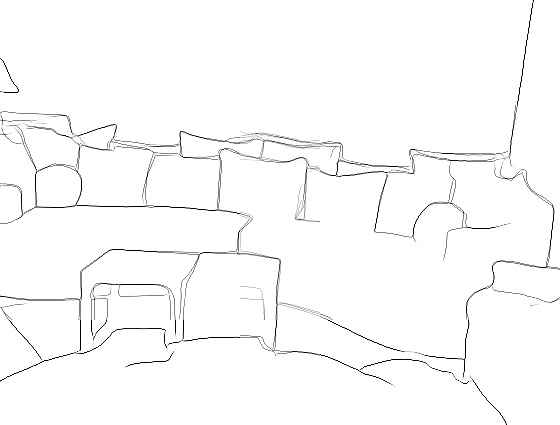}
&\includegraphics[width=.22\textwidth,frame]{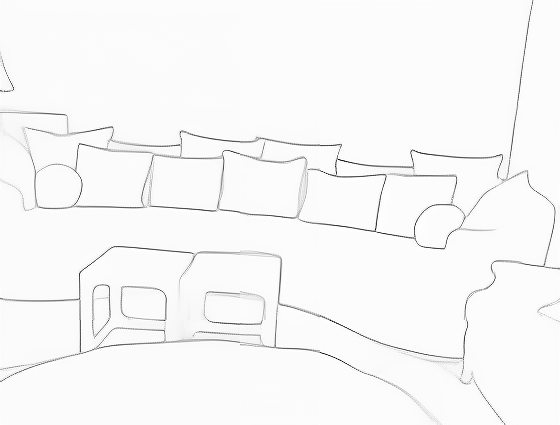}\\

\rotatebox{90}{ \quad  BIPED}&
\includegraphics[width=.22\textwidth,frame]{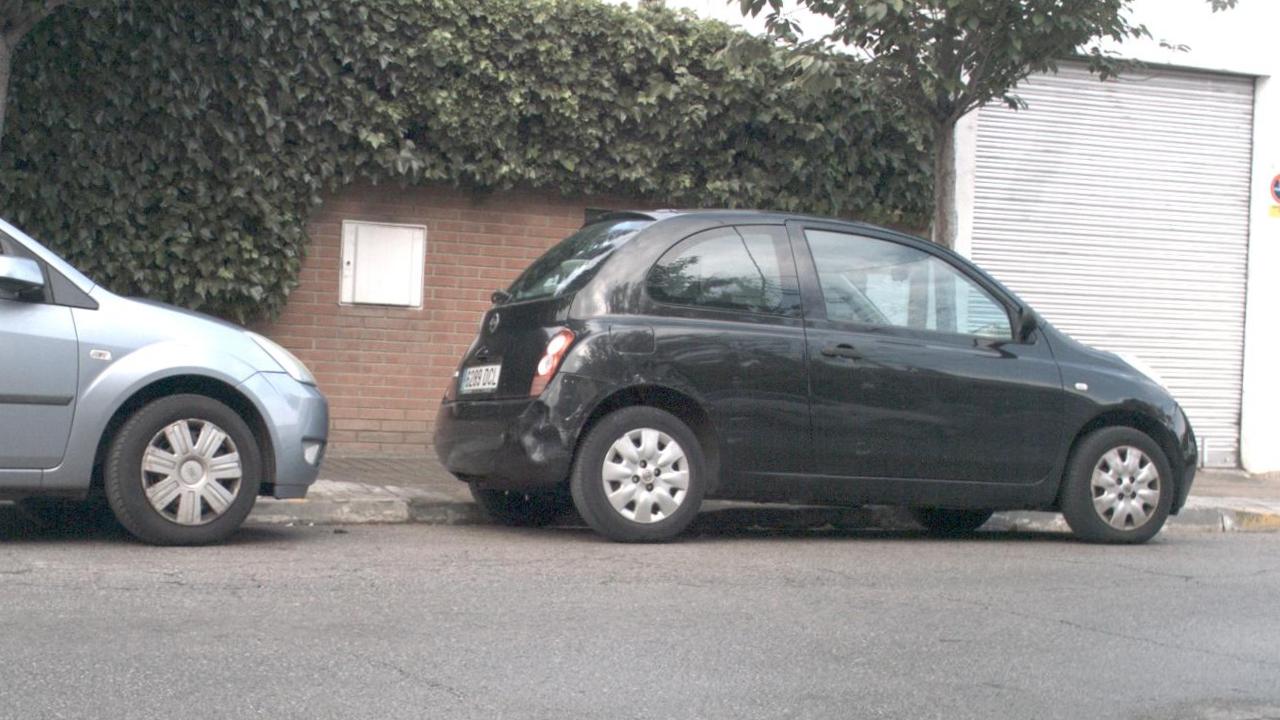} 
&\includegraphics[width=.22\textwidth,frame]{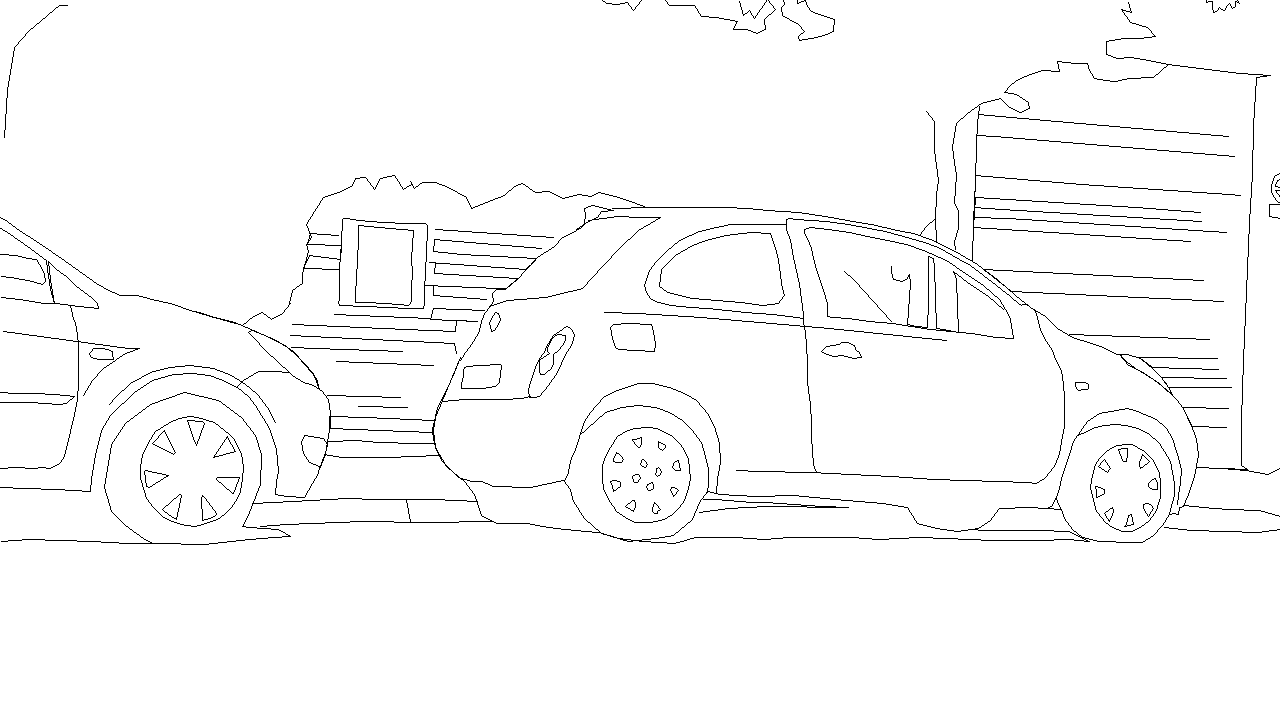}
&\includegraphics[width=.22\textwidth,frame]{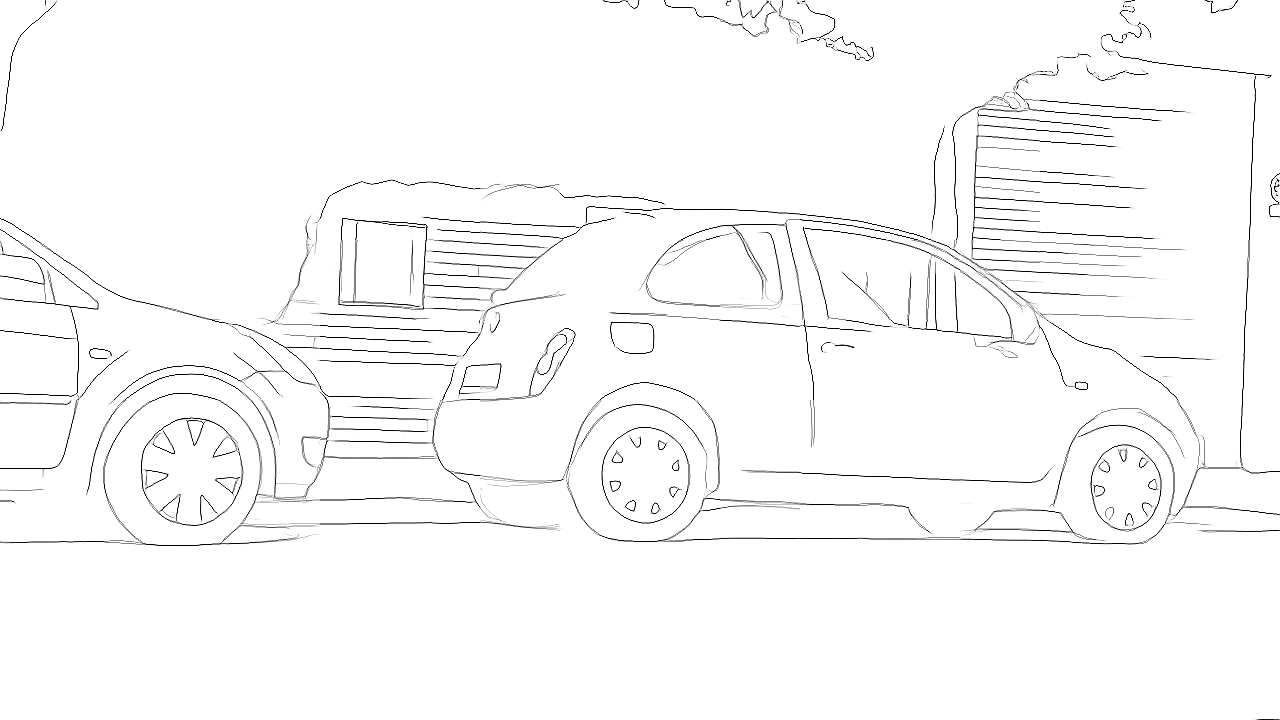}
&\includegraphics[width=.22\textwidth,frame]{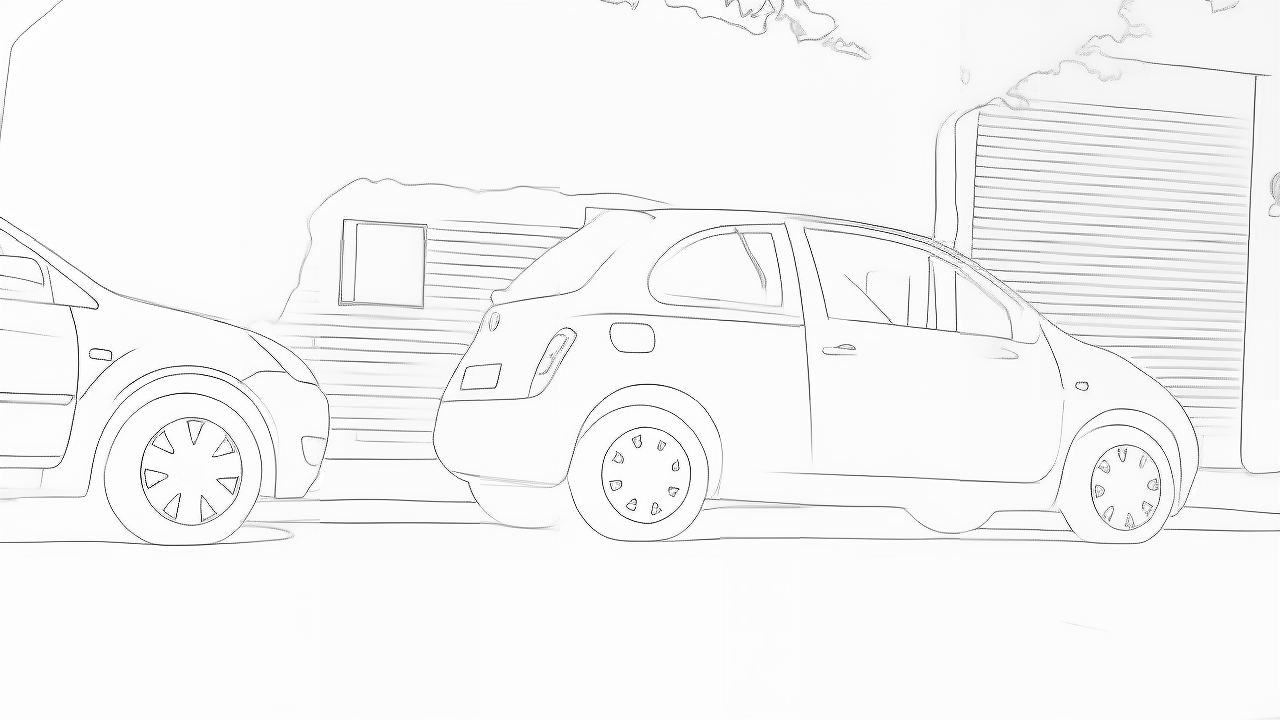}
\\
&Input& GT&DiffusionEdge~\cite{ye2024diffusionedge}& Ours \\
\end{tabular}
\caption{Qualitative comparisons on challenging samples in the NYUD and BIPED test set.}
\label{Fig5}
\vspace{-10pt}
\end{figure*}

\textbf{Results on the NYUD Dataset.} Since only one label is available on the NYUD Dataset, we do not use the granularity information. The results on the NYUD dataset are shown in Table~\ref{table2}. The previous best edge detector on this dataset is EdgeSAM-S, which obtains 0.783 ODS and 0.797 OIS. The proposal of our method once again improves performance by a large margin, improving ODS and OIS to 0.800 and 0.809 on the NYUD test dataset, which can also be observed from Fig.~\ref{Fig5}.

\textbf{Results on the BIPED Dataset.}
Table~\ref{table2} also shows the results on the BIPED Dataset. We can see that compared with DiffusionEdge, our proposed GED achieves the best OIS and much higher AP (from 0.642 to 0.934). The qualitative results in Fig.~\ref{Fig5} also demonstrate that our method can lead to more detailed and consistent predictions.

\subsection{Ablation Study}
Our proposed GED contains several important parts, including the network design, text prompt, the integration of the granularity, and granularity regularization. Here we conduct ablation studies to prove the effectiveness of each part. The results are presented in Table~\ref{tab:ablation} and Table~\ref{tab:step}.

\textbf{Network Design.} As shown in Fig.~\ref{Fig2}, different from other efforts, such as training a diffusion model from scratch (DiffusionEdge~\cite{ye2024diffusionedge}) or finetuning the denoising U-Net (Marigold~\cite{ke2023repurposing}), we propose to constrain the prediction aligned with ground truth in latent space without multi-step inference. Such design is the key foundation of our proposed GED. For verification, we follow Marigold~\cite{ke2023repurposing} to finetune the denoising U-Net as an edge detector on the BSDS dataset. As shown in Table~\ref{tab:step}, Marigold improves the performance of DiffusionEdge by 0.7\% and 0.8\% in ODS and OIS metrics, showing the effectiveness of the pre-trained diffusion model. However, the benefit of multi-step inference is marginal and even leads to performance drop. Moreover, the results of Marigold on the edge detection task are much lower than our design. Particularly, our proposed baseline achieves 0.859, 0.875 and 0.910 in ODS, OIS and AP (1st row in Table~\ref{tab:ablation}), which are superior to Marigold by 1.8\%, 1.9\% and 0.4\% on those metrics. We also compare with the Marigold~\cite{ke2023repurposing} in the depth estimation task to verify the effectiveness of our network design in Supplementary Material.

\textbf{Text Prompt.} In our method, we introduce the text prompt as a condition to predict the latent edge maps, where Minigpt-4 is used to generate captions for all images, and text embedding is fed to the denoising U-Net to enhance the performance. Table~\ref{tab:ablation} shows the introduction of text prompt improves ODS from 0.859 to 0.861, and OIS from 0.875 to 0.877. The text prompt also improves performance on the BIPED dataset, increasing the ODS from 0.892 to 0.896, and on the NYUD dataset, raising the ODS from 0.795 to 0.800.

\begin{table}[ht]
\setlength{\abovecaptionskip}{5pt}
\centering
\caption{The evaluation without NMS.}
\renewcommand\arraystretch{1.1}
\renewcommand\tabcolsep{5pt}
\label{tab:nonms}
\begin{tabular}{c|cc}
\toprule
Method&DiffusionEdge & GED (Ours) \\
\toprule
ODS&0.745&0.779\\
OIS&0.753&0.788\\
AP&0.585&0.759\\
\bottomrule
\end{tabular}
\vspace{-10pt}
\end{table}

\begin{table}[ht]
\setlength{\abovecaptionskip}{5pt}
\centering
\caption{Different granularity integration.}
\renewcommand\arraystretch{1.1}
\renewcommand\tabcolsep{5pt}
\label{tab:granularity}
\begin{tabular}{c|ccc}
\toprule
Method&Text Prompt & Time Step& Encoding \\
\toprule
ODS&0.864&0.867&0.870\\
OIS&0.880&0.882&0.880\\
AP&0.909&0.911&0.907\\
\bottomrule
\end{tabular}
\vspace{-10pt}
\end{table}

\begin{table}[ht]
\setlength{\abovecaptionskip}{5pt}
\caption{The exploration of minimal supervision signal.}
\centering
\renewcommand\arraystretch{1.1}
\renewcommand\tabcolsep{5pt}
\label{tab:min}
\begin{tabular}{c|cccccc}
\toprule
Method& 1\%&5\%&10\% &20\%&50\%&100\%\\
\toprule
ODS&0.788&0.831&0.836&0.848&0.857&0.870\\
OIS&0.805&0.849&0.854&0.864&0.873&0.880\\
AP&0.825&0.873&0.877&0.885&0.900&0.907\\
\bottomrule
\end{tabular}
\vspace{-10pt}
\end{table}

\textbf{Granularity Guidance.} The guidance of edge granularity is capable of generating diverse and controllable edge predictions for different downstream applications, where edge maps with granularity 0 are more consistent with object boundaries, and edge maps with granularity 1 contain rich details (Fig.~\ref{fig4}). As shown in Table~\ref{tab:ablation}, introducing granularity improves ODS from 0.861 to 0.865, and OIS from 0.877 to 0.880. Besides, the explicit granularity regularization constraints further improve the precise control of edge predictors, which increases ODS from 0.865 to 0.870.

\subsection{Further Analysis}
We conduct more experiments to explore the advantages of our method, including crispness, the strategy of integrating granularity, and the generalization ability with limited data.

\textbf{Crispness without NMS.} To evaluate the crispness, we report metrics without non-maximum
suppression (NMS) following DiffusionEdge~\cite{ye2024diffusionedge} and CATS~\cite{huan2021unmixing} on the NYUD dataset ( Table~\ref{tab:nonms}). Our proposed GED also achieves excellent results without NMS post-processing. Compared with DiffusionEdge, the scores increase by 3.4\%, 3.5\% and 17.4\% in ODS, OIS and AP, demonstrating that GED can generate both accurate and crisp edges without extra post-processing.

\textbf{Strategy of Integrating Granularity.}
We compare three strategies for introducing granularity into the stable diffusion model. The first is designing the text prompt for denoising U-Net, \ie, `Edge granularity denotes different levels of detail, please extract the edges with the granularity of $g$.' The second is directly using the time step, where we align the granularity $g$ with the time step $t$ by transforming $g$ from $[0,1]$ to the integer in $[0, 1000]$. The third is encoding the granularity into the dimension same as the time embeddings by two liner layers and adding them pixel-wise. As shown in Table~\ref{tab:granularity}, those three methods can all generate controllable results, and we choose the encoding strategy for better quantitative performance.

\textbf{Generalization with Minimum Supervision Data.} The pre-trained stable diffusion has rich prior, so we are interested in the exploration of minimal supervision signal need. We conduct experiments on the BSDS dataset. Table~\ref{tab:min} shows that only 1\% training data (\ie, 3 training images) can obtain 0.788 ODS and 0.805 OIS on the BSDS test set, which is comparable with the CNN-based HED (0.788 ODS and 0.808 OIS). Only using 5\% training data obtains 0.831 ODS and 0.849 OIS, which is close to the results of one-step DiffusionEdge (0.833 ODS and 0.844 OIS). Especially, only using 20\% training data, we can obtain 0.848 ODS and 0.864 OIS, which has surpassed the existing best results (EdgeSAM-S with 0.838 ODS and 852 OIS). The results suggest that our proposed GED exhibits excellent generalization ability by exploring the underlying rich scene understanding prior via the pre-trained diffusion model.

\section{Conclusion and Discussion}
\label{con}
Existing diffusion-based edge detectors rely on forward noise adding and reverse denoising processes, which is not the optimal choice for the edge detection task. In this paper, we utilize the rich prior in the pre-trained stable diffusion model, and propose a novel generative edge detector, named GED, waiving the multi-step denoising process. The proposed GED does not require designing any specific network structure, so it is easy to train and requests fewer training iterations. We also realize controllable and diverse edge predictions by introducing granularity as one of the conditions of denoising U-Net and keeping ordinal relationships with explicit granularity regularization.
The excellent performance on four common datasets proves the effectiveness of our exploration. 

While GED has achieved significant improvement for the multiple granularity edge detection task, one limitation is the quality of the granularity since the annotated edge maps lack explicit granularity guidance. We will explore constructing an edge detection dataset covering from the rough object contours to detailed edges in the future.

{
    \small
    \bibliographystyle{ieeenat_fullname}
    \bibliography{egbib}

\begin{thebibliography}{79}
\providecommand{\natexlab}[1]{#1}
\providecommand{\url}[1]{\texttt{#1}}
\expandafter\ifx\csname urlstyle\endcsname\relax
  \providecommand{\doi}[1]{doi: #1}\else
  \providecommand{\doi}{doi: \begingroup \urlstyle{rm}\Url}\fi

\bibitem[Arbelaez et~al.(2010)Arbelaez, Maire, Fowlkes, and Malik]{arbelaez2010contour}
Pablo Arbelaez, Michael Maire, Charless Fowlkes, and Jitendra Malik.
\newblock Contour detection and hierarchical image segmentation.
\newblock \emph{IEEE Trans. Pattern Anal. Mach. Intell.}, 33\penalty0 (5):\penalty0 898--916, 2010.

\bibitem[Baranchuk et~al.(2022)Baranchuk, Voynov, Rubachev, Khrulkov, and Babenko]{baranchuk2022labelefficient}
Dmitry Baranchuk, Andrey Voynov, Ivan Rubachev, Valentin Khrulkov, and Artem Babenko.
\newblock Label-efficient semantic segmentation with diffusion models.
\newblock In \emph{Int. Conf. Learn. Represent.}, 2022.

\bibitem[Bertasius et~al.(2015)Bertasius, Shi, and Torresani]{bertasius2015deepedge}
Gedas Bertasius, Jianbo Shi, and Lorenzo Torresani.
\newblock Deepedge: A multi-scale bifurcated deep network for top-down contour detection.
\newblock In \emph{IEEE Conf. Comput. Vis. Pattern Recog.}, pages 4380--4389, 2015.

\bibitem[Cabon et~al.(2020)Cabon, Murray, and Humenberger]{cabon2020virtual}
Yohann Cabon, Naila Murray, and Martin Humenberger.
\newblock Virtual kitti 2.
\newblock \emph{arXiv preprint arXiv:2001.10773}, 2020.

\bibitem[Canny(1986)]{canny1986computational}
John Canny.
\newblock A computational approach to edge detection.
\newblock \emph{IEEE Trans. Pattern Anal. Mach. Intell.}, 8\penalty0 (6):\penalty0 679--698, 1986.

\bibitem[Cetinkaya et~al.(2024)Cetinkaya, Kalkan, and Akbas]{cetinkaya2024ranked}
Bedrettin Cetinkaya, Sinan Kalkan, and Emre Akbas.
\newblock Ranked: Addressing imbalance and uncertainty in edge detection using ranking-based losses.
\newblock In \emph{IEEE Conf. Comput. Vis. Pattern Recog.}, 2024.

\bibitem[Chen et~al.(2023)Chen, Dong, Lu, Yu, and Han]{chen2023self}
Hao Chen, Yonghan Dong, Zheming Lu, Yunlong Yu, and Jungong Han.
\newblock Self-prompting perceptual edge learning for dense prediction.
\newblock \emph{IEEE Transactions on Circuits and Systems for Video Technology}, 2023.

\bibitem[Chen et~al.(2024)Chen, Sun, and Lin]{chen2024camodiffusion}
Zhongxi Chen, Ke Sun, and Xianming Lin.
\newblock Camodiffusion: Camouflaged object detection via conditional diffusion models.
\newblock In \emph{AAAI}, pages 1272--1280, 2024.

\bibitem[Deng and Liu(2020)]{deng2020deep}
Ruoxi Deng and Shengjun Liu.
\newblock Deep structural contour detection.
\newblock In \emph{ACM Int. Conf. Multimedia}, pages 304--312, 2020.

\bibitem[Deng et~al.(2018)Deng, Shen, Liu, Wang, and Liu]{deng2018learning}
Ruoxi Deng, Chunhua Shen, Shengjun Liu, Huibing Wang, and Xinru Liu.
\newblock Learning to predict crisp boundaries.
\newblock In \emph{Eur. Conf. Comput. Vis.}, pages 562--578, 2018.

\bibitem[Deng et~al.(2021)Deng, Liu, Wang, Wang, Zhao, and Zhang]{deng2021learning}
Ruoxi Deng, Shengjun Liu, Jinxin Wang, Huibing Wang, Hanli Zhao, and Xiaoqin Zhang.
\newblock Learning to decode contextual information for efficient contour detection.
\newblock In \emph{ACM Int. Conf. Multimedia}, pages 4435--4443, 2021.

\bibitem[Ding et~al.(2024)Ding, Zhao, Sun, Tan, Wang, Ma, and Fang]{ding2024edge}
Jin Ding, Jie-Chao Zhao, Yong-Zhi Sun, Ping Tan, Jia-Wei Wang, Ji-En Ma, and You-Tong Fang.
\newblock Edge detectors can make deep convolutional neural networks more robust.
\newblock \emph{arXiv preprint arXiv:2402.16479}, 2024.

\bibitem[Doll{\'a}r and Zitnick(2014)]{dollar2014fast}
Piotr Doll{\'a}r and C~Lawrence Zitnick.
\newblock Fast edge detection using structured forests.
\newblock \emph{IEEE Trans. Pattern Anal. Mach. Intell.}, 37\penalty0 (8):\penalty0 1558--1570, 2014.

\bibitem[Du et~al.(2023)Du, Kolkin, Shakhnarovich, and Bhattad]{du2023intrinsic}
Xiaodan Du, Nicholas Kolkin, Greg Shakhnarovich, and Anand Bhattad.
\newblock Intrinsic lora: A generalist approach for discovering knowledge in generative models.
\newblock \emph{arXiv preprint arXiv:2311.17137}, 2023.

\bibitem[Fu and Guo(2023)]{fu2023practical}
Yuanbin Fu and Xiaojie Guo.
\newblock Practical edge detection via robust collaborative learning.
\newblock In \emph{ACM Int. Conf. Multimedia}, 2023.

\bibitem[Gao et~al.(2021)Gao, Tang, Lang, and Lv]{gao2021end}
Yi Gao, Chenwei Tang, Jiulin Lang, and Jiancheng Lv.
\newblock End-to-end edge detection via improved transformer model.
\newblock In \emph{International Conference on Neural Information Processing}, pages 514--525. Springer, 2021.

\bibitem[He et~al.(2024)He, Li, Zhang, Zhang, You, Guo, Li, Danelljan, and Yu]{he2024strategic}
Chunming He, Kai Li, Yachao Zhang, Yulun Zhang, Chenyu You, Zhenhua Guo, Xiu Li, Martin Danelljan, and Fisher Yu.
\newblock Strategic preys make acute predators: Enhancing camouflaged object detectors by generating camouflaged objects.
\newblock In \emph{Int. Conf. Learn. Represent.}, 2024.

\bibitem[He et~al.(2019)He, Zhang, Yang, Shan, and Huang]{he2019bi}
Jianzhong He, Shiliang Zhang, Ming Yang, Yanhu Shan, and Tiejun Huang.
\newblock Bi-directional cascade network for perceptual edge detection.
\newblock In \emph{IEEE Conf. Comput. Vis. Pattern Recog.}, pages 3828--3837, 2019.

\bibitem[He et~al.(2023)He, Lin, Luo, Xie, Song, Liu, and Shen]{he2023shift}
Xilin He, Qinliang Lin, Cheng Luo, Weicheng Xie, Siyang Song, Feng Liu, and Linlin Shen.
\newblock Shift from texture-bias to shape-bias: Edge deformation-based augmentation for robust object recognition.
\newblock In \emph{Int. Conf. Comput. Vis.}, pages 1526--1535, 2023.

\bibitem[Ho et~al.(2020)Ho, Jain, and Abbeel]{ho2020denoising}
Jonathan Ho, Ajay Jain, and Pieter Abbeel.
\newblock Denoising diffusion probabilistic models.
\newblock \emph{Adv. Neural Inform. Process. Syst.}, 33:\penalty0 6840--6851, 2020.

\bibitem[Hu et~al.(2022)Hu, Shen, Wallis, Allen-Zhu, Li, Wang, Wang, and Chen]{hu2022lora}
Edward~J Hu, Yelong Shen, Phillip Wallis, Zeyuan Allen-Zhu, Yuanzhi Li, Shean Wang, Lu Wang, and Weizhu Chen.
\newblock Lo{RA}: Low-rank adaptation of large language models.
\newblock In \emph{Int. Conf. Learn. Represent.}, 2022.

\bibitem[Huan et~al.(2021)Huan, Xue, Zheng, He, Gong, and Xia]{huan2021unmixing}
Linxi Huan, Nan Xue, Xianwei Zheng, Wei He, Jianya Gong, and Gui-Song Xia.
\newblock Unmixing convolutional features for crisp edge detection.
\newblock \emph{PAMI}, 44\penalty0 (10):\penalty0 6602--6609, 2021.

\bibitem[Jaini et~al.(2024)Jaini, Clark, and Geirhos]{jaini2024intriguing}
Priyank Jaini, Kevin Clark, and Robert Geirhos.
\newblock Intriguing properties of generative classifiers.
\newblock In \emph{Int. Conf. Learn. Represent.}, 2024.

\bibitem[Ji et~al.(2023)Ji, Chen, Xie, Hong, Liu, Liu, Lu, Li, and Luo]{ji2023ddp}
Yuanfeng Ji, Zhe Chen, Enze Xie, Lanqing Hong, Xihui Liu, Zhaoqiang Liu, Tong Lu, Zhenguo Li, and Ping Luo.
\newblock Ddp: Diffusion model for dense visual prediction.
\newblock In \emph{Int. Conf. Comput. Vis.}, pages 21741--21752, 2023.

\bibitem[Ke et~al.(2024)Ke, Obukhov, Huang, Metzger, Daudt, and Schindler]{ke2023repurposing}
Bingxin Ke, Anton Obukhov, Shengyu Huang, Nando Metzger, Rodrigo~Caye Daudt, and Konrad Schindler.
\newblock Repurposing diffusion-based image generators for monocular depth estimation.
\newblock In \emph{IEEE Conf. Comput. Vis. Pattern Recog.}, 2024.

\bibitem[Khani et~al.(2024)Khani, Asgari, Sanghi, Amiri, and Hamarneh]{khani2024slime}
Aliasghar Khani, Saeid Asgari, Aditya Sanghi, Ali~Mahdavi Amiri, and Ghassan Hamarneh.
\newblock {SL}ime: Segment like me.
\newblock In \emph{Int. Conf. Learn. Represent.}, 2024.

\bibitem[Kim et~al.(2022)Kim, Hwang, and Kwak]{kim2022learning}
Namyup Kim, Sehyun Hwang, and Suha Kwak.
\newblock Learning to detect semantic boundaries with image-level class labels.
\newblock \emph{Int. J. Comput. Vis.}, pages 1--18, 2022.

\bibitem[Kingma and Welling(2013)]{kingma2013auto}
Diederik~P Kingma and Max Welling.
\newblock Auto-encoding variational bayes.
\newblock \emph{arXiv preprint arXiv:1312.6114}, 2013.

\bibitem[Kirillov et~al.(2023)Kirillov, Mintun, Ravi, Mao, Rolland, Gustafson, Xiao, Whitehead, Berg, Lo, et~al.]{kirillov2023segment}
Alexander Kirillov, Eric Mintun, Nikhila Ravi, Hanzi Mao, Chloe Rolland, Laura Gustafson, Tete Xiao, Spencer Whitehead, Alexander~C Berg, Wan-Yen Lo, et~al.
\newblock Segment anything.
\newblock In \emph{Int. Conf. Comput. Vis.}, pages 4015--4026, 2023.

\bibitem[Kittler(1983)]{kittler1983accuracy}
Josef Kittler.
\newblock On the accuracy of the sobel edge detector.
\newblock \emph{Image and Vision Computing}, 1\penalty0 (1):\penalty0 37--42, 1983.

\bibitem[Kokkinos(2016)]{kokkinos2015pushing}
Iasonas Kokkinos.
\newblock Pushing the boundaries of boundary detection using deep learning.
\newblock \emph{Int. Conf. Learn. Represent.}, 2016.

\bibitem[Kondapaneni et~al.(2023)Kondapaneni, Marks, Knott, Guimaraes, and Perona]{kondapaneni2023text}
Neehar Kondapaneni, Markus Marks, Manuel Knott, Rog{\'e}rio Guimaraes, and Pietro Perona.
\newblock Text-image alignment for diffusion-based perception.
\newblock \emph{arXiv preprint arXiv:2310.00031}, 2023.

\bibitem[Lee et~al.(2024)Lee, Tseng, Lee, and Yang]{lee2024dmp}
Hsin-Ying Lee, Hung-Yu Tseng, Hsin-Ying Lee, and Ming-Hsuan Yang.
\newblock Exploiting diffusion prior for generalizable dense prediction.
\newblock In \emph{IEEE Conf. Comput. Vis. Pattern Recog.}, 2024.

\bibitem[Li et~al.(2018)Li, Yang, Cheng, Liu, and Shen]{li2018contour}
Xin Li, Fan Yang, Hong Cheng, Wei Liu, and Dinggang Shen.
\newblock Contour knowledge transfer for salient object detection.
\newblock In \emph{Eur. Conf. Comput. Vis.}, pages 355--370, 2018.

\bibitem[Li et~al.(2023)Li, Zhou, Zhang, Zhang, Wang, and Xie]{li2023open}
Ziyi Li, Qinye Zhou, Xiaoyun Zhang, Ya Zhang, Yanfeng Wang, and Weidi Xie.
\newblock Open-vocabulary object segmentation with diffusion models.
\newblock In \emph{Int. Conf. Comput. Vis.}, pages 7667--7676, 2023.

\bibitem[Lim et~al.(2013)Lim, Zitnick, and Doll{\'{a}}r]{lim2013sketch}
Joseph~J. Lim, C.~Lawrence Zitnick, and Piotr Doll{\'{a}}r.
\newblock Sketch tokens: {A} learned mid-level representation for contour and object detection.
\newblock In \emph{IEEE Conf. Comput. Vis. Pattern Recog.}, pages 3158--3165, 2013.

\bibitem[Liu and Lew(2016)]{liu2016learning}
Yu Liu and Michael~S Lew.
\newblock Learning relaxed deep supervision for better edge detection.
\newblock In \emph{IEEE Conf. Comput. Vis. Pattern Recog.}, pages 231--240, 2016.

\bibitem[Liu et~al.(2017)Liu, Cheng, Hu, Wang, and Bai]{liu2017richer}
Yun Liu, Ming-Ming Cheng, Xiaowei Hu, Kai Wang, and Xiang Bai.
\newblock Richer convolutional features for edge detection.
\newblock In \emph{IEEE Conf. Comput. Vis. Pattern Recog.}, pages 3000--3009, 2017.

\bibitem[Loshchilov and Hutter(2019)]{loshchilov2018decoupled}
Ilya Loshchilov and Frank Hutter.
\newblock Decoupled weight decay regularization.
\newblock In \emph{Int. Conf. Learn. Represent.}, 2019.

\bibitem[Maninis et~al.(2016)Maninis, Pont-Tuset, Arbel{\'a}ez, and Van~Gool]{maninis2016cob}
Kevis-Kokitsi Maninis, Jordi Pont-Tuset, Pablo Arbel{\'a}ez, and Luc Van~Gool.
\newblock Convolutional oriented boundaries.
\newblock In \emph{Eur. Conf. Comput. Vis.}, pages 580--596. Springer, 2016.

\bibitem[Marr and Hildreth(1980)]{marr1980theory}
David Marr and Ellen Hildreth.
\newblock Theory of edge detection.
\newblock \emph{Proceedings of the Royal Society of London. Series B. Biological Sciences}, 207\penalty0 (1167):\penalty0 187--217, 1980.

\bibitem[Martin et~al.(2004)Martin, Fowlkes, and Malik]{martin2004learning}
David~R Martin, Charless~C Fowlkes, and Jitendra Malik.
\newblock Learning to detect natural image boundaries using local brightness, color, and texture cues.
\newblock \emph{IEEE Trans. Pattern Anal. Mach. Intell.}, 26\penalty0 (5):\penalty0 530--549, 2004.

\bibitem[M{\'e}ly et~al.(2016)M{\'e}ly, Kim, McGill, Guo, and Serre]{mely2016systematic}
David~A M{\'e}ly, Junkyung Kim, Mason McGill, Yuliang Guo, and Thomas Serre.
\newblock A systematic comparison between visual cues for boundary detection.
\newblock \emph{Vision research}, 120:\penalty0 93--107, 2016.

\bibitem[Paszke et~al.(2019)Paszke, Gross, Massa, Lerer, Bradbury, Chanan, Killeen, Lin, Gimelshein, Antiga, et~al.]{paszke2019pytorch}
Adam Paszke, Sam Gross, Francisco Massa, Adam Lerer, James Bradbury, Gregory Chanan, Trevor Killeen, Zeming Lin, Natalia Gimelshein, Luca Antiga, et~al.
\newblock Pytorch: An imperative style, high-performance deep learning library.
\newblock \emph{Adv. Neural Inform. Process. Syst.}, 32, 2019.

\bibitem[Patni et~al.(2024)Patni, Agarwal, and Arora]{patni2024ecodepth}
Suraj Patni, Aradhye Agarwal, and Chetan Arora.
\newblock Ecodepth: Effective conditioning of diffusion models for monocular depth estimation.
\newblock In \emph{IEEE Conf. Comput. Vis. Pattern Recog.}, 2024.

\bibitem[Poma et~al.(2020)Poma, Riba, and Sappa]{poma2020dense}
Xavier~Soria Poma, Edgar Riba, and Angel Sappa.
\newblock Dense extreme inception network: Towards a robust cnn model for edge detection.
\newblock In \emph{IEEE Winter Conf. Appl. Comput. Vis.}, pages 1923--1932, 2020.

\bibitem[Pu et~al.(2022)Pu, Huang, Liu, Guan, and Ling]{pu2022edter}
Mengyang Pu, Yaping Huang, Yuming Liu, Qingji Guan, and Haibin Ling.
\newblock Edter: Edge detection with transformer.
\newblock In \emph{IEEE Conf. Comput. Vis. Pattern Recog.}, pages 1402--1412, 2022.

\bibitem[Ramesh et~al.(2022)Ramesh, Dhariwal, Nichol, Chu, and Chen]{ramesh2022hierarchical}
Aditya Ramesh, Prafulla Dhariwal, Alex Nichol, Casey Chu, and Mark Chen.
\newblock Hierarchical text-conditional image generation with clip latents.
\newblock \emph{arXiv preprint arXiv:2204.06125}, 1\penalty0 (2):\penalty0 3, 2022.

\bibitem[Rombach et~al.(2022)Rombach, Blattmann, Lorenz, Esser, and Ommer]{rombach2022high}
Robin Rombach, Andreas Blattmann, Dominik Lorenz, Patrick Esser, and Bj{\"o}rn Ommer.
\newblock High-resolution image synthesis with latent diffusion models.
\newblock In \emph{IEEE Conf. Comput. Vis. Pattern Recog.}, pages 10684--10695, 2022.

\bibitem[Saharia et~al.(2022)Saharia, Chan, Saxena, Li, Whang, Denton, Ghasemipour, Gontijo~Lopes, Karagol~Ayan, Salimans, et~al.]{saharia2022photorealistic}
Chitwan Saharia, William Chan, Saurabh Saxena, Lala Li, Jay Whang, Emily~L Denton, Kamyar Ghasemipour, Raphael Gontijo~Lopes, Burcu Karagol~Ayan, Tim Salimans, et~al.
\newblock Photorealistic text-to-image diffusion models with deep language understanding.
\newblock \emph{Adv. Neural Inform. Process. Syst.}, 35:\penalty0 36479--36494, 2022.

\bibitem[Schuhmann et~al.(2022)Schuhmann, Beaumont, Vencu, Gordon, Wightman, Cherti, Coombes, Katta, Mullis, Wortsman, et~al.]{schuhmann2022laion}
Christoph Schuhmann, Romain Beaumont, Richard Vencu, Cade Gordon, Ross Wightman, Mehdi Cherti, Theo Coombes, Aarush Katta, Clayton Mullis, Mitchell Wortsman, et~al.
\newblock Laion-5b: An open large-scale dataset for training next generation image-text models.
\newblock \emph{Adv. Neural Inform. Process. Syst.}, 35:\penalty0 25278--25294, 2022.

\bibitem[Shen et~al.(2015)Shen, Wang, Wang, Bai, and Zhang]{shen2015deepcontour}
Wei Shen, Xinggang Wang, Yan Wang, Xiang Bai, and Zhijiang Zhang.
\newblock Deepcontour: A deep convolutional feature learned by positive-sharing loss for contour detection.
\newblock In \emph{IEEE Conf. Comput. Vis. Pattern Recog.}, pages 3982--3991, 2015.

\bibitem[Silberman et~al.(2012)Silberman, Hoiem, Kohli, and Fergus]{silberman2012indoor}
Nathan Silberman, Derek Hoiem, Pushmeet Kohli, and Rob Fergus.
\newblock Indoor segmentation and support inference from rgbd images.
\newblock In \emph{Eur. Conf. Comput. Vis.}, pages 746--760. Springer, 2012.

\bibitem[Song et~al.(2021)Song, Meng, and Ermon]{song2021denoising}
Jiaming Song, Chenlin Meng, and Stefano Ermon.
\newblock Denoising diffusion implicit models.
\newblock In \emph{Int. Conf. Learn. Represent.}, 2021.

\bibitem[Su et~al.(2021)Su, Liu, Yu, Hu, Liao, Tian, Pietik{\"a}inen, and Liu]{su2021pixel}
Zhuo Su, Wenzhe Liu, Zitong Yu, Dewen Hu, Qing Liao, Qi Tian, Matti Pietik{\"a}inen, and Li Liu.
\newblock Pixel difference networks for efficient edge detection.
\newblock In \emph{Int. Conf. Comput. Vis.}, pages 5117--5127, 2021.

\bibitem[Sun et~al.(2021)Sun, Li, Xiao, Qiu, Kailkhura, Liu, and Li]{sun2021can}
Mingjie Sun, Zichao Li, Chaowei Xiao, Haonan Qiu, Bhavya Kailkhura, Mingyan Liu, and Bo Li.
\newblock Can shape structure features improve model robustness under diverse adversarial settings?
\newblock In \emph{Int. Conf. Comput. Vis.}, pages 7526--7535, 2021.

\bibitem[Talker et~al.(2024)Talker, Cohen, Yosef, Dana, and Dinerstein]{talker2022mind}
Lior Talker, Aviad Cohen, Erez Yosef, Alexandra Dana, and Michael Dinerstein.
\newblock Mind the edge: Refining depth edges in sparsely-supervised monocular depth estimation.
\newblock In \emph{IEEE Conf. Comput. Vis. Pattern Recog.}, 2024.

\bibitem[Tripathi et~al.(2023)Tripathi, Singh, Chakraborty, and Shenoy]{tripathi2023edges}
Aditay Tripathi, Rishubh Singh, Anirban Chakraborty, and Pradeep Shenoy.
\newblock Edges to shapes to concepts: Adversarial augmentation for robust vision.
\newblock In \emph{IEEE Conf. Comput. Vis. Pattern Recog.}, pages 24470--24479, 2023.

\bibitem[Van~Gansbeke and De~Brabandere(2024)]{van2024simple}
Wouter Van~Gansbeke and Bert De~Brabandere.
\newblock A simple latent diffusion approach for panoptic segmentation and mask inpainting.
\newblock \emph{arXiv preprint arXiv:2401.10227}, 2024.

\bibitem[Voynov et~al.(2023)Voynov, Aberman, and Cohen-Or]{voynov2023sketch}
Andrey Voynov, Kfir Aberman, and Daniel Cohen-Or.
\newblock Sketch-guided text-to-image diffusion models.
\newblock In \emph{ACM SIGGRAPH 2023 Conference Proceedings}, pages 1--11, 2023.

\bibitem[Wang et~al.(2022)Wang, Zhang, Cui, Ren, Yang, Xie, Hua, Bao, and Xu]{wang2022active}
Chi Wang, Yunke Zhang, Miaomiao Cui, Peiran Ren, Yin Yang, Xuansong Xie, Xian-Sheng Hua, Hujun Bao, and Weiwei Xu.
\newblock Active boundary loss for semantic segmentation.
\newblock In \emph{AAAI}, pages 2397--2405, 2022.

\bibitem[Wang et~al.(2024)Wang, Cao, Xie, Yang, and Pang]{wang2024implicit}
Hefeng Wang, Jiale Cao, Jin Xie, Aiping Yang, and Yanwei Pang.
\newblock Implicit and explicit language guidance for diffusion-based visual perception.
\newblock \emph{arXiv preprint arXiv:2404.07600}, 2024.

\bibitem[Wang et~al.(2023)Wang, Li, Zhang, Xu, Zhou, Yu, Sheng, and Xu]{wang2023diffusion}
Jinglong Wang, Xiawei Li, Jing Zhang, Qingyuan Xu, Qin Zhou, Qian Yu, Lu Sheng, and Dong Xu.
\newblock Diffusion model is secretly a training-free open vocabulary semantic segmenter.
\newblock \emph{arXiv preprint arXiv:2309.02773}, 2023.

\bibitem[Wang et~al.(2017)Wang, Zhao, and Huang]{wang2017deep}
Yupei Wang, Xin Zhao, and Kaiqi Huang.
\newblock Deep crisp boundaries.
\newblock In \emph{IEEE Conf. Comput. Vis. Pattern Recog.}, pages 3892--3900, 2017.

\bibitem[Xie and Tu(2015)]{xie2015holistically}
Saining Xie and Zhuowen Tu.
\newblock Holistically-nested edge detection.
\newblock In \emph{Int. Conf. Comput. Vis.}, pages 1395--1403, 2015.

\bibitem[Xu et~al.(2017)Xu, Ouyang, Alameda-Pineda, Ricci, Wang, and Sebe]{xu2017AMHNet}
Dan Xu, Wanli Ouyang, Xavier Alameda-Pineda, Elisa Ricci, Xiaogang Wang, and Nicu Sebe.
\newblock Learning deep structured multi-scale features using attention-gated crfs for contour prediction.
\newblock In \emph{Adv. Neural Inform. Process. Syst.}, pages 3961--3970, 2017.

\bibitem[Xuan et~al.(2022)Xuan, Huang, Liu, and Du]{xuan2022fcl}
Wenjie Xuan, Shaoli Huang, Juhua Liu, and Bo Du.
\newblock Fcl-net: Towards accurate edge detection via fine-scale corrective learning.
\newblock \emph{Neural Networks}, 145:\penalty0 248--259, 2022.

\bibitem[Yamagiwa et~al.(2024)Yamagiwa, Takase, Kambe, and Nakamoto]{yamagiwa2024zero}
Hiroaki Yamagiwa, Yusuke Takase, Hiroyuki Kambe, and Ryosuke Nakamoto.
\newblock Zero-shot edge detection with scesame: Spectral clustering-based ensemble for segment anything model estimation.
\newblock In \emph{IEEE Winter Conf. Appl. Comput. Vis.}, pages 541--551, 2024.

\bibitem[Yang et~al.(2024)Yang, Chen, Wu, Qin, Yan, Mao, and Song]{yang2024boosting}
Wenya Yang, Xiao-Diao Chen, Wen Wu, Hongshuai Qin, Kangming Yan, Xiaoyang Mao, and Haichuan Song.
\newblock Boosting deep unsupervised edge detection via segment anything model.
\newblock \emph{IEEE Transactions on Industrial Informatics}, 2024.

\bibitem[Ye et~al.(2023)Ye, Yi, Gao, Cai, and Xu]{ye2023delving}
Yunfan Ye, Renjiao Yi, Zhirui Gao, Zhiping Cai, and Kai Xu.
\newblock Delving into crispness: Guided label refinement for crisp edge detection.
\newblock \emph{IEEE Trans. Image Process.}, 2023.

\bibitem[Ye et~al.(2024)Ye, Xu, Huang, Yi, and Cai]{ye2024diffusionedge}
Yunfan Ye, Kai Xu, Yuhang Huang, Renjiao Yi, and Zhiping Cai.
\newblock Diffusionedge: Diffusion probabilistic model for crisp edge detection.
\newblock In \emph{AAAI}, 2024.

\bibitem[Zbinden et~al.(2023)Zbinden, Doorenbos, Pissas, Huber, Sznitman, and M{\'a}rquez-Neila]{zbinden2023stochastic}
Lukas Zbinden, Lars Doorenbos, Theodoros Pissas, Adrian~Thomas Huber, Raphael Sznitman, and Pablo M{\'a}rquez-Neila.
\newblock Stochastic segmentation with conditional categorical diffusion models.
\newblock In \emph{Int. Conf. Comput. Vis.}, pages 1119--1129, 2023.

\bibitem[Zhang et~al.(2023)Zhang, Tian, Liao, Hua, Zou, and Xu]{zhang2023learning}
Yuhang Zhang, Shishun Tian, Muxin Liao, Guoguang Hua, Wenbin Zou, and Chen Xu.
\newblock Learning shape-invariant representation for generalizable semantic segmentation.
\newblock \emph{IEEE Trans. Image Process.}, 2023.

\bibitem[Zhao et~al.(2023)Zhao, Rao, Liu, Liu, Zhou, and Lu]{zhao2023unleashing}
Wenliang Zhao, Yongming Rao, Zuyan Liu, Benlin Liu, Jie Zhou, and Jiwen Lu.
\newblock Unleashing text-to-image diffusion models for visual perception.
\newblock In \emph{Int. Conf. Comput. Vis.}, pages 5729--5739, 2023.

\bibitem[Zheng et~al.(2021)Zheng, Lu, Zhao, Zhu, Luo, Wang, Fu, Feng, Xiang, Torr, et~al.]{zheng2021rethinking}
Sixiao Zheng, Jiachen Lu, Hengshuang Zhao, Xiatian Zhu, Zekun Luo, Yabiao Wang, Yanwei Fu, Jianfeng Feng, Tao Xiang, Philip~HS Torr, et~al.
\newblock Rethinking semantic segmentation from a sequence-to-sequence perspective with transformers.
\newblock In \emph{IEEE Conf. Comput. Vis. Pattern Recog.}, pages 6881--6890, 2021.

\bibitem[Zhou et~al.(2023)Zhou, Huang, Pu, Guan, Huang, and Ling]{zhou2023treasure}
Caixia Zhou, Yaping Huang, Mengyang Pu, Qingji Guan, Li Huang, and Haibin Ling.
\newblock The treasure beneath multiple annotations: An uncertainty-aware edge detector.
\newblock In \emph{IEEE Conf. Comput. Vis. Pattern Recog.}, pages 15507--15517, 2023.

\bibitem[Zhou et~al.(2024)Zhou, Huang, Pu, Guan, Deng, and Ling]{zhou2024muge}
Caixia Zhou, Yaping Huang, Mengyang Pu, Qingji Guan, Ruoxi Deng, and Haibin Ling.
\newblock Muge: Multiple granularity edge detection.
\newblock In \emph{IEEE Conf. Comput. Vis. Pattern Recog.}, 2024.

\bibitem[Zhu et~al.(2023)Zhu, Chen, Shen, xiang Li, and Elhoseiny]{zhu2022minigpt4}
Deyao Zhu, Jun Chen, Xiaoqian Shen, xiang Li, and Mohamed Elhoseiny.
\newblock Minigpt-4: Enhancing vision-language understanding with advanced large language models, 2023.

\bibitem[Zou et~al.(2024)Zou, Yang, Zhang, Li, Li, Wang, Wang, Gao, and Lee]{zou2024segment}
Xueyan Zou, Jianwei Yang, Hao Zhang, Feng Li, Linjie Li, Jianfeng Wang, Lijuan Wang, Jianfeng Gao, and Yong~Jae Lee.
\newblock Segment everything everywhere all at once.
\newblock \emph{Adv. Neural Inform. Process. Syst.}, 36, 2024.

\end{thebibliography}
}


\clearpage
\setcounter{page}{1}
\makeatletter
\newenvironment{tablehere}
  {\def\@captype{table}}
 {}

\newenvironment{figurehere}
 {\def\@captype{figure}}
 {}
\makeatother
\onecolumn

\begin{center}
        \Large
        \textbf{\thetitle}\\
        \vspace{0.5em}Supplementary Material \\
        \vspace{0.5em}    
\end{center}
\appendix
\begin{multicols}{2}

In this supplementary material, we display our proposed GED compared with the Marigold~\cite{ke2023repurposing} in the depth estimation task to verify the effectiveness of our network design. Then we provide a more lightweight training strategy with LoRA~\cite{hu2022lora}, which has some performance drop. 
We also show the Pytorch-like pseudocode and more additional visualization results.

\section{Experiments on the Depth Estimation Task}
\label{marigold}
Edge is an important cue to identify objects at different distances, so depth estimation is likely related with the task of edge detection. Moreover, as a representative work of using the pre-trained diffusion model, Marigold~\cite{ke2023repurposing} provides affine-ambiguous depth prediction results with high accuracy and sharp boundaries. So in this section, we perform the performance comparison in the depth estimation to verify the effectiveness and potential of our proposed edge detector.

Specifically, Marigold~\cite{ke2023repurposing} encodes the RGB image and the dense ground truth depth map into the latent space to obtain the latent image feature maps $\mathbf{z}_i$ and the depth maps $\mathbf{z}_d$ by the encoder of the variational autoencoder (VAE)~\cite{kingma2013auto}. Then it adds noise to the latent depth maps, which can be denoted as $\mathbf{z}_d^t$. The latent image feature maps $\mathbf{z}_i$ and noisy latent depth maps $\mathbf{z}_d^t$ are concatenated by the channel and fed into the denoising U-Net to predict the current noise level. During inference, denoising diffusion implicit model (DDIM)~\cite{song2021denoising} is used to generate clean depth latent maps, which are then decoded by the VAE decoder into a predicted depth map.

In contrast to Marigold, our proposed GED takes only clean latent image feature maps $\mathbf{z}_i$ as inputs and predicts the latent depth maps $\mathbf{z}_d$, which does not need multi-step inference. The result on KITTI~\cite{cabon2020virtual} dataset is shown in Table~\ref{tab:supp_depth_training}. We can see that our proposed method achieves better performance in both AbsRel and $\delta_1$, demonstrating that our proposed GED is also effective in the depth estimation task. 
\end{multicols}

\begin{table*}[ht] 
\caption{The results on the KITTI~\cite{cabon2020virtual} dataset for the monocular depth estimation task.}
\centering
\renewcommand\arraystretch{1.1}
\renewcommand\tabcolsep{8pt}
\begin{tabular}{l|ccc|cc}
\Xhline{1px}
Methods&Input&Prediction&Time Step&AbsRel$\downarrow$ & $\delta_1$$\uparrow$\\
\hline
Marigold~\cite{ke2023repurposing}&$\textrm{concat}(\textbf{z}_i; \textbf{z}_d^t)$  & noise & $t\sim[1,1000]$ & 9.90 & 91.60\\
GED (Ours)&$\textbf{z}_i$ & $\textbf{z}_d$ & $t=1$ & 9.64 & 96.21\\
\Xhline{1px}
\end{tabular}
\label{tab:supp_depth_training}
\end{table*}

\begin{table*}[ht] 
\caption{The results on the BSDS dataset~\cite{arbelaez2010contour} for comparing LoRA and our proposed method.}
\centering
\renewcommand\arraystretch{1.1}
\renewcommand\tabcolsep{6pt}
\begin{tabular}{l|cc|ccc}
\Xhline{1px}
Methods&Total Parameters&Trainable Parameters&ODS&OIS&AP\\
\hline
LoRA~\cite{hu2022lora}&866.8M&0.83M&0.812 & 0.827&0.857\\
GED (Ours)&866M&116M&0.870 & 0.880&0.907\\
\Xhline{1px}
\end{tabular}
\label{tab:lora}
\end{table*}

\begin{multicols}{2}

\section{Finetuning with LoRA}
\label{lora}
In our proposed GED, we design to finetune a part of the denoising U-Net to decrease the training cost. Another possible solution is to use a low-rank adaption strategy, LoRA~\cite{hu2022lora}, which freezes the pre-trained model weights and injects trainable rank decomposition matrices. Though LoRA increases the total parameters, it reduces the trainable parameters to 0.83M with rank 4 in our experiments. To verify the effectiveness of our finetuning strategy, we compare the LoRA based strategy with ours. As shown in Table~\ref{tab:lora}, stable diffusion adapted with LoRA achieves 0.812, 0.827 and 0.857 in terms of ODS, OID and AP on the BSDS dataset~\cite{li2018contour}, which is much lower than our proposed GED.

\section{Pytorch-like Pseudocode}
\label{pse}
In this section, we provide the main Pytorch-like pseudocode in Algorithm~\ref{alg:loss}.

\section{More Visualization Results}
\label{vis}
\begin{figure*}
\small
\centering
\renewcommand\arraystretch{1}
\renewcommand\tabcolsep{3pt}
\begin{tabular}{cccccc}

\includegraphics[width=.15\textwidth,frame]{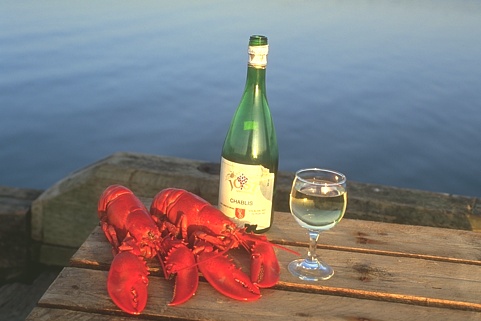} 
&\includegraphics[width=.15\textwidth,frame]{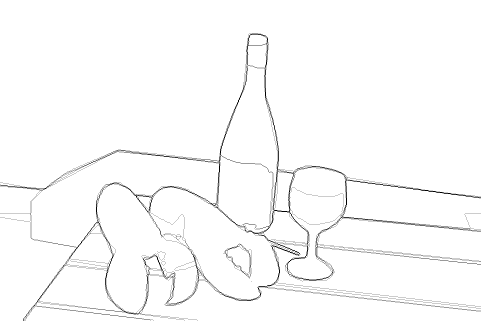}
&\includegraphics[width=.15\textwidth,frame]{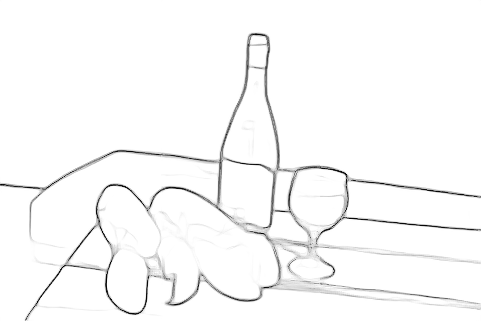}
&\includegraphics[width=.15\textwidth,frame]{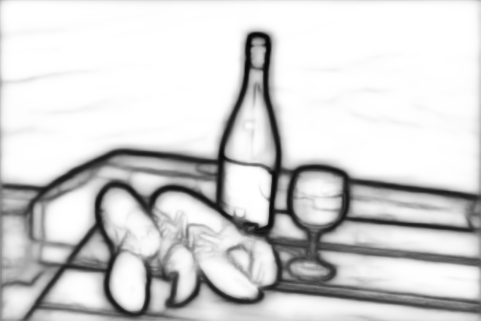}
&\includegraphics[width=.15\textwidth,frame]{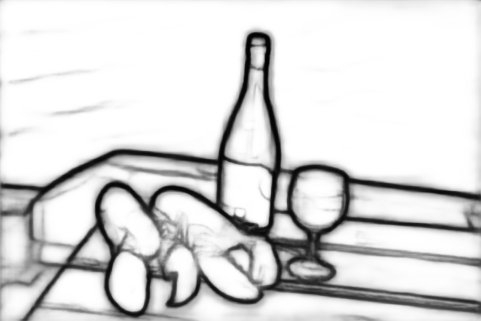}
&\includegraphics[width=.15\textwidth,frame]{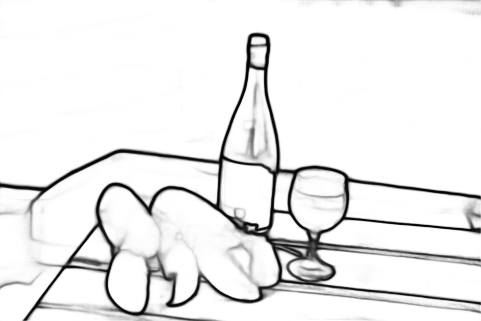}
\\
Input&  GT& DiffEdge~\cite{ye2024diffusionedge}&  RCF~\cite{liu2017richer}&  BDCN~\cite{he2019bi}& EDTER~\cite{pu2022edter}\\
\includegraphics[width=.15\textwidth,frame]{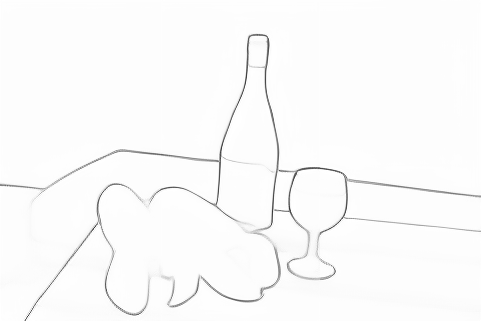}
&\includegraphics[width=.15\textwidth,frame]{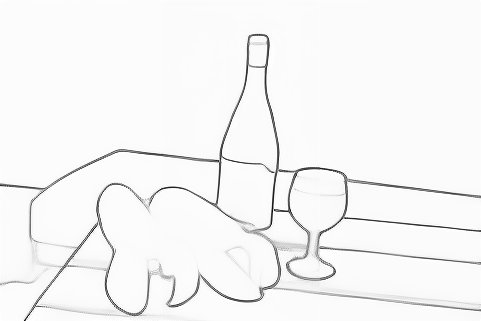}
&\includegraphics[width=.15\textwidth,frame]{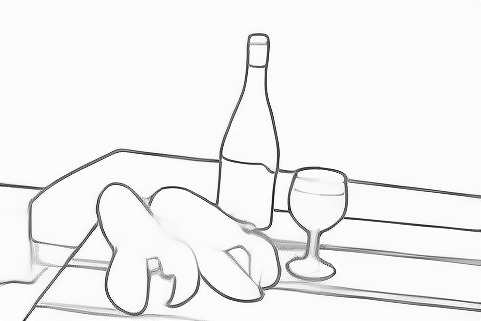}
&\includegraphics[width=.15\textwidth,frame]{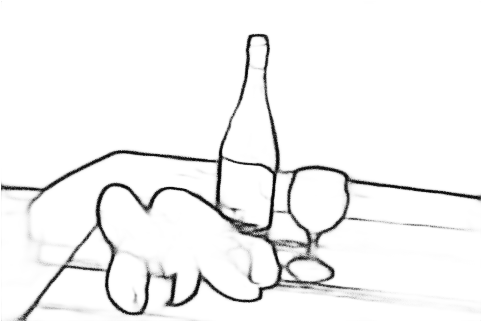} 
&\includegraphics[width=.15\textwidth,frame]{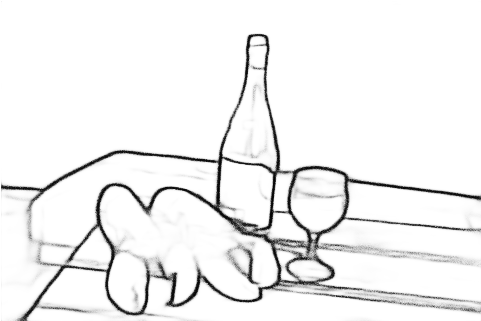}
&\includegraphics[width=.15\textwidth,frame]{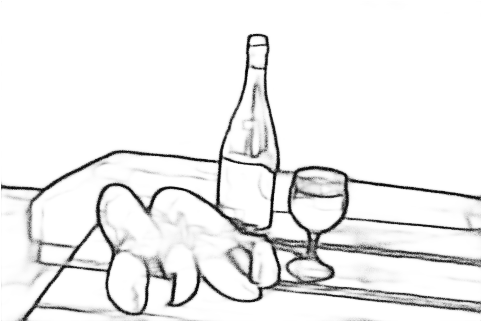}
\\  
Ours (0) & Ours (0.5) & Ours (1)&MuGE (0)~\cite{zhou2024muge} &  MuGE (0.5)~\cite{zhou2024muge} &  MuGE (1)~\cite{zhou2024muge} \\

\includegraphics[width=.15\textwidth,frame]{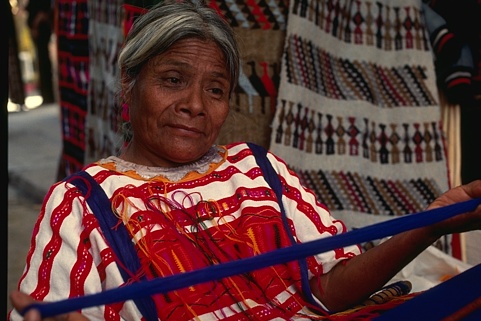} 
&\includegraphics[width=.15\textwidth,frame]{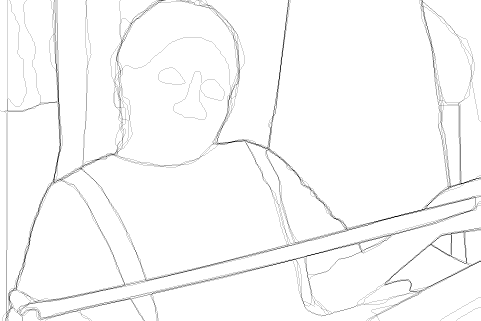}
&\includegraphics[width=.15\textwidth,frame]{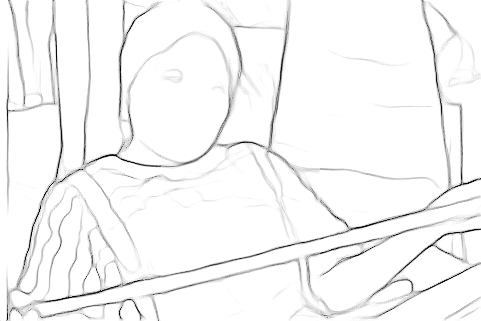}
&\includegraphics[width=.15\textwidth,frame]{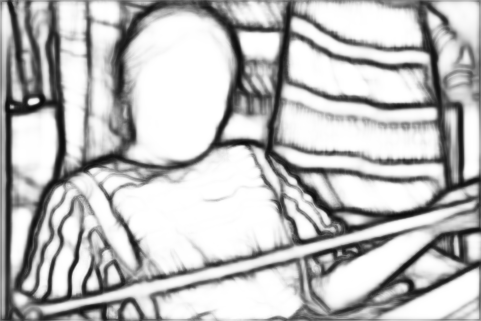}
&\includegraphics[width=.15\textwidth,frame]{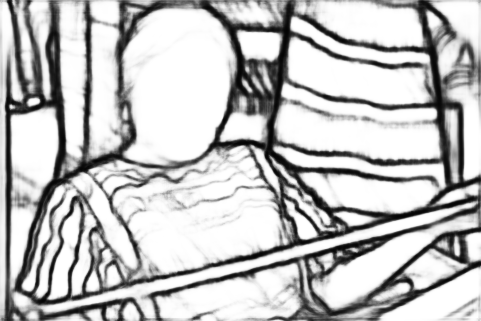}
&\includegraphics[width=.15\textwidth,frame]{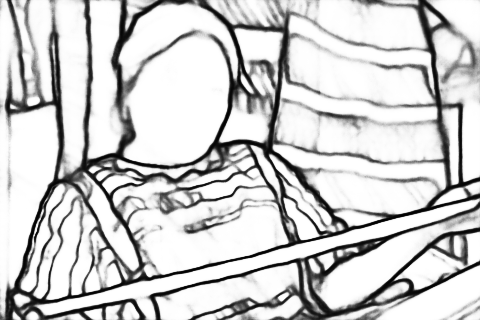}
\\
Input&  GT&  DiffEdge~\cite{ye2024diffusionedge}&RCF~\cite{liu2017richer}& BDCN~\cite{he2019bi}& EDTER~\cite{pu2022edter} \\
\includegraphics[width=.15\textwidth,frame]{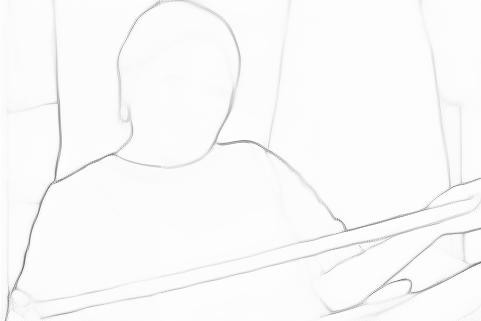} 
&\includegraphics[width=.15\textwidth,frame]{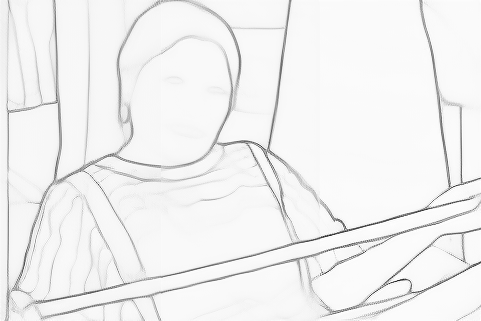}
&\includegraphics[width=.15\textwidth,frame]{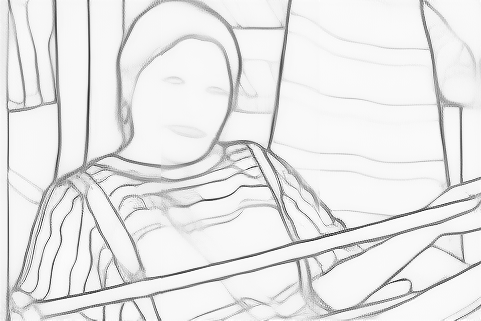}
&\includegraphics[width=.15\textwidth,frame]{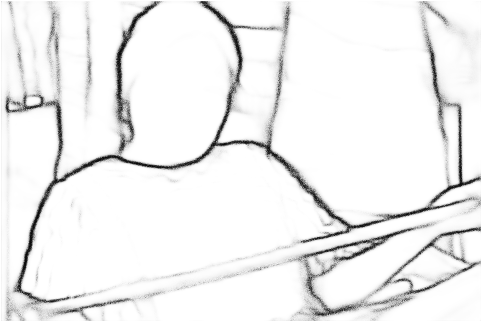} 
&\includegraphics[width=.15\textwidth,frame]{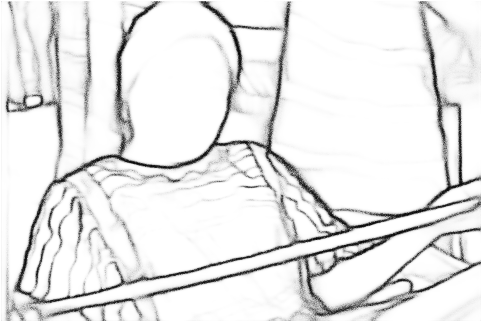}
&\includegraphics[width=.15\textwidth,frame]{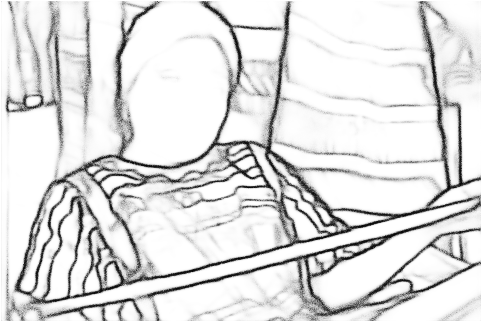}
\\
Ours (0) & Ours (0.5) & Ours (1)&MuGE (0)~\cite{zhou2024muge} &  MuGE (0.5)~\cite{zhou2024muge} &  MuGE (1)~\cite{zhou2024muge}  \\

\includegraphics[width=.15\textwidth,frame]{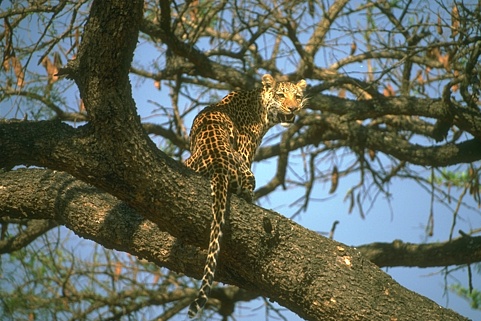} 
&\includegraphics[width=.15\textwidth,frame]{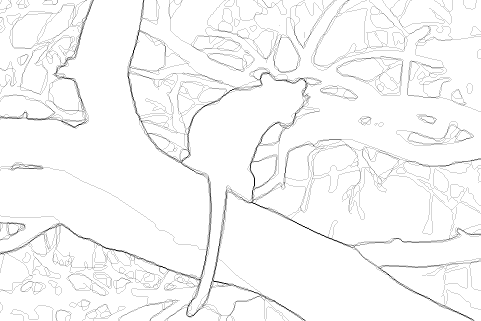}
&\includegraphics[width=.15\textwidth,frame]{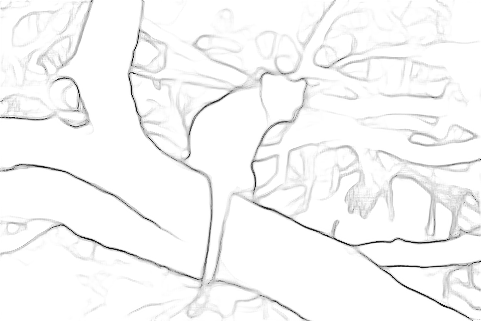}
&\includegraphics[width=.15\textwidth,frame]{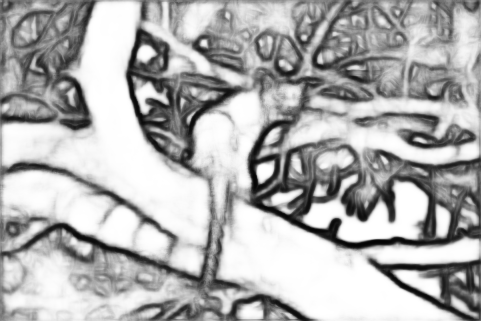}
&\includegraphics[width=.15\textwidth,frame]{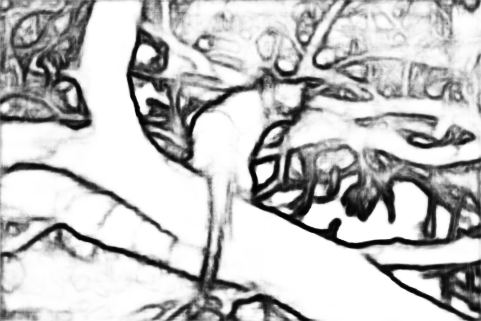}
&\includegraphics[width=.15\textwidth,frame]{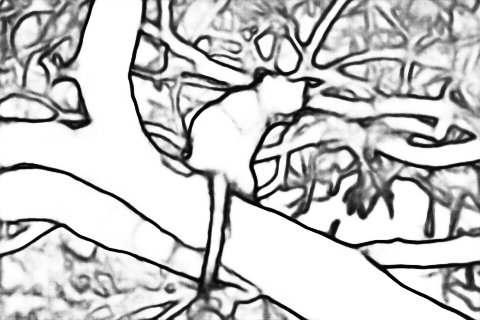}
\\
Input&  GT&  DiffEdge~\cite{ye2024diffusionedge}&RCF~\cite{liu2017richer}& BDCN~\cite{he2019bi}& EDTER~\cite{pu2022edter} \\
\includegraphics[width=.15\textwidth,frame]{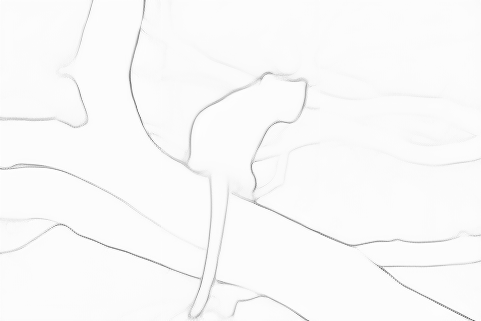} 
&\includegraphics[width=.15\textwidth,frame]{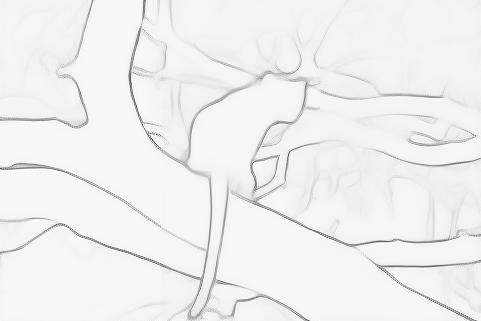}
&\includegraphics[width=.15\textwidth,frame]{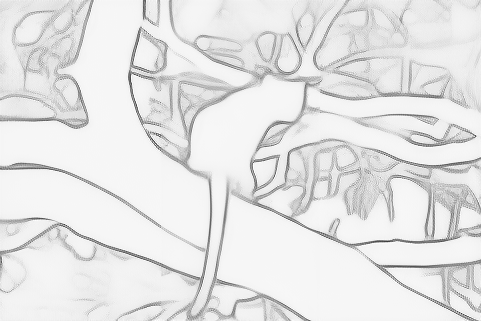}
&\includegraphics[width=.15\textwidth,frame]{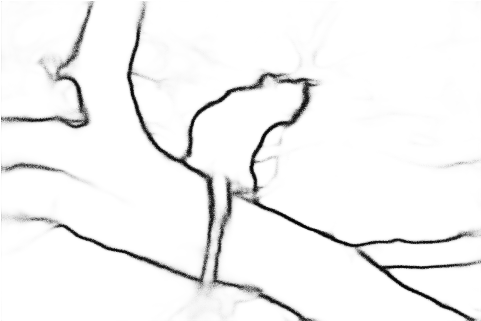} 
&\includegraphics[width=.15\textwidth,frame]{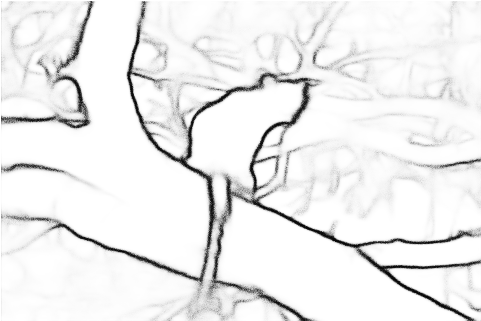}
&\includegraphics[width=.15\textwidth,frame]{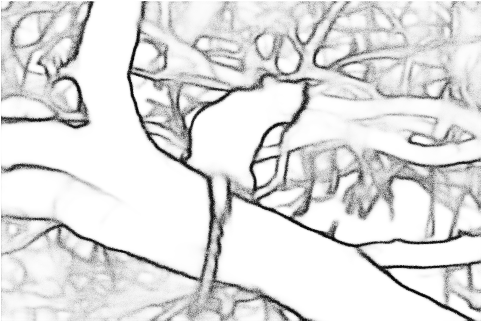}
\\  
Ours (0) & Ours (0.5) & Ours (1) & MuGE (0)~\cite{zhou2024muge} &  MuGE (0.5)~\cite{zhou2024muge} &  MuGE (1)~\cite{zhou2024muge}  \\

\includegraphics[width=.15\textwidth,frame]{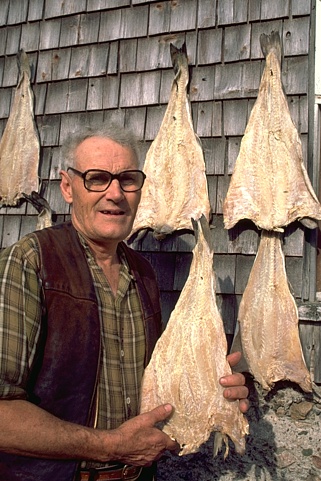} 
&\includegraphics[width=.15\textwidth,frame]{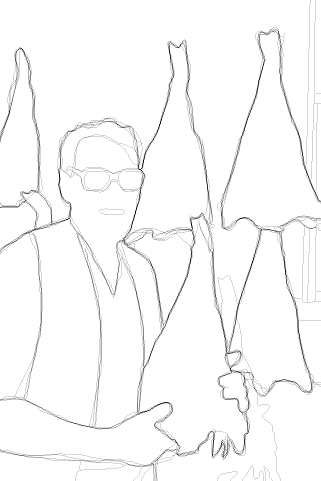}
&\includegraphics[width=.15\textwidth,frame]{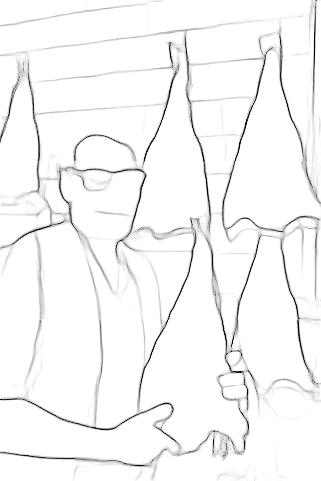}
&\includegraphics[width=.15\textwidth,frame]{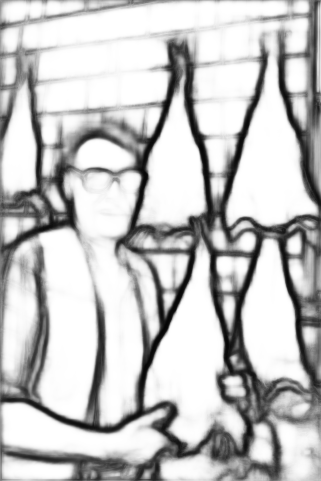}
&\includegraphics[width=.15\textwidth,frame]{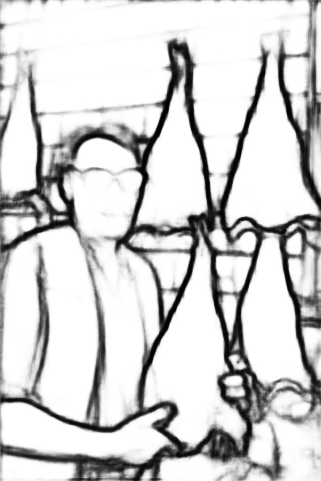}
&\includegraphics[width=.15\textwidth,frame]{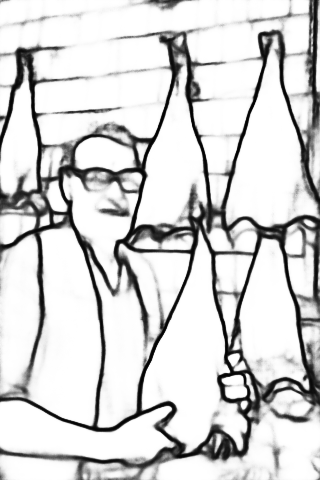}
\\
Input&  GT&  DiffEdge~\cite{ye2024diffusionedge}&RCF~\cite{liu2017richer}& BDCN~\cite{he2019bi}& EDTER~\cite{pu2022edter} \\

\includegraphics[width=.15\textwidth,frame]{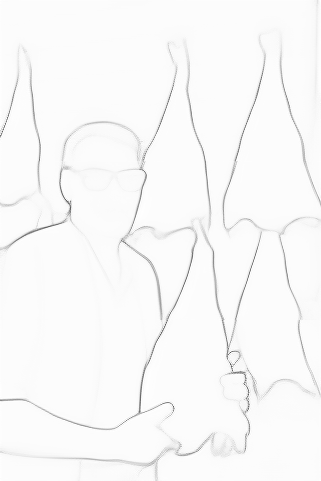} 
&\includegraphics[width=.15\textwidth,frame]{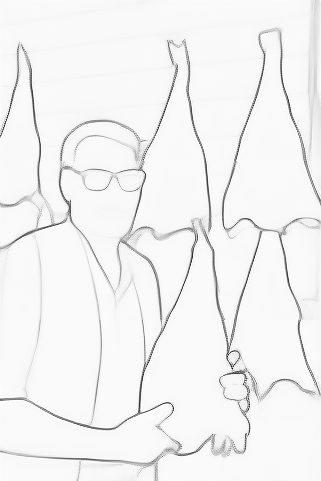}
&\includegraphics[width=.15\textwidth,frame]{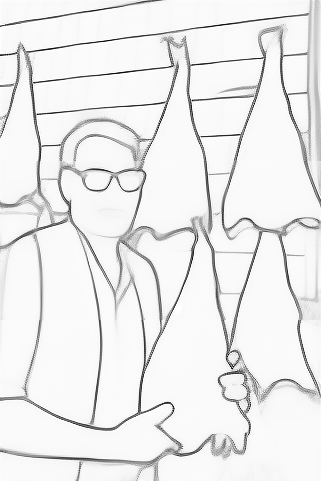}
&\includegraphics[width=.15\textwidth,frame]{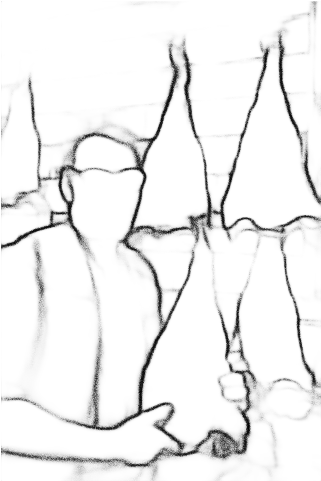} 
&\includegraphics[width=.15\textwidth,frame]{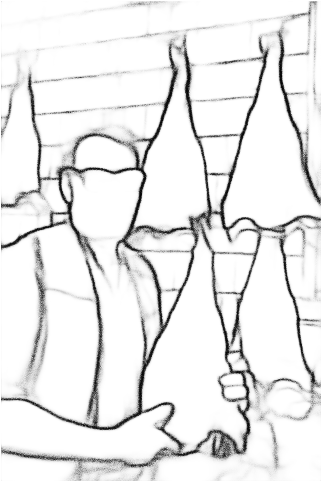}
&\includegraphics[width=.15\textwidth,frame]{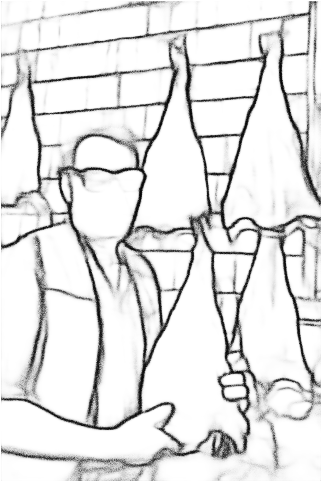}
\\ Ours (0) & Ours (0.5) & Ours (1)&MuGE (0)~\cite{zhou2024muge} &  MuGE (0.5)~\cite{zhou2024muge} &  MuGE (1)~\cite{zhou2024muge}  \\

\end{tabular}
\caption{Qualitative comparisons on challenging samples in the BSDS500 test set. Note that MuGE and our proposed GED produce diverse results with edge granularity of 0, 0.5, and 1, respectively.}
\label{fig_compa_bsds}
\vspace{-10pt}
\end{figure*}

\begin{figure*}[!t]
\small
\setlength{\abovecaptionskip}{3pt}
\setlength{\belowcaptionskip}{0pt}
\centering
\begin{tabular}{cccc}
\hspace{-.23cm}
\includegraphics[width=.21\textwidth]{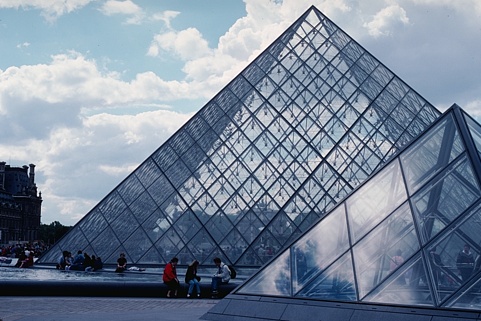}
&\hspace{-.23cm}\includegraphics[width=.21\textwidth,frame]{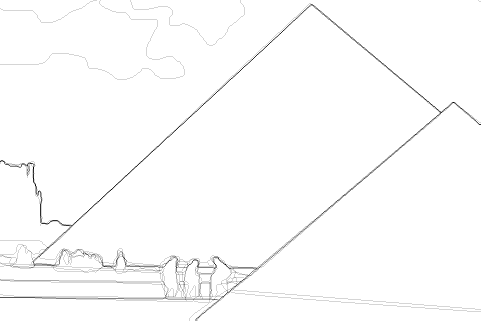}
&\hspace{-.23cm}\includegraphics[width=.21\textwidth,frame]{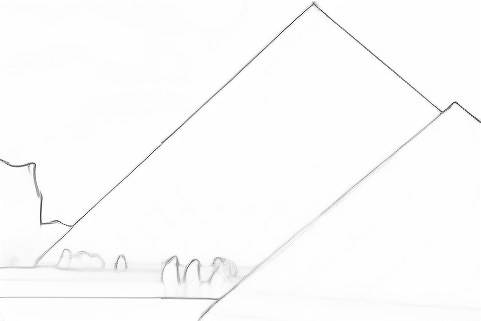}
&\hspace{-.23cm}\includegraphics[width=.21\textwidth,frame]{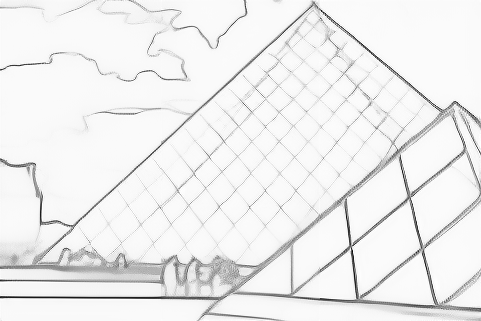}
\vspace{-.06cm}\\

\hspace{-.23cm}
\includegraphics[width=.21\textwidth]{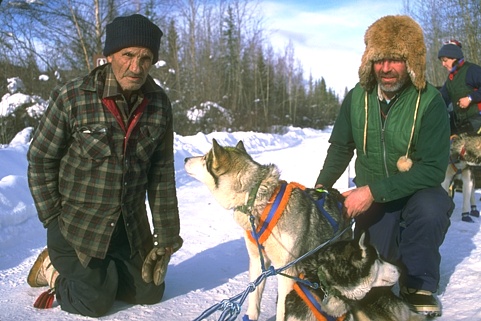}
&\hspace{-.23cm}\includegraphics[width=.21\textwidth,frame]{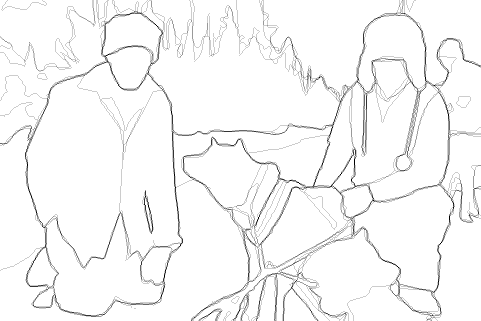}
&\hspace{-.23cm}\includegraphics[width=.21\textwidth,frame]{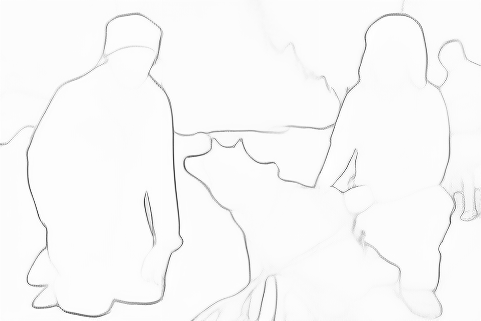}
&\hspace{-.23cm}\includegraphics[width=.21\textwidth,frame]{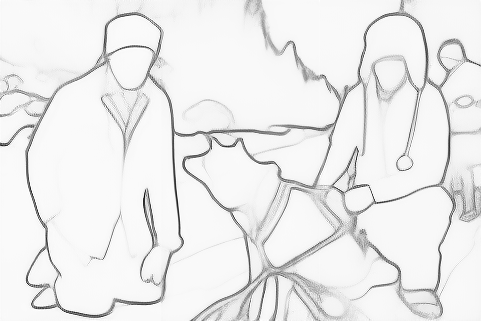}
\vspace{-.06cm}\\

\hspace{-.23cm} Input & \hspace{-.23cm} GT &\hspace{-.23cm}  Ours (0) &\hspace{-.23cm}  Ours (1) \\
\end{tabular}
\caption{Qualitative results with different edge granularity on the BSDS~\cite{arbelaez2010contour} test dataset.}
\label{fig_bsds}
\end{figure*}

\begin{figure*}[!t]
\small
\setlength{\abovecaptionskip}{3pt}
\setlength{\belowcaptionskip}{0pt}
\centering
\begin{tabular}{cccc}
\hspace{-.23cm}
\includegraphics[width=.21\textwidth]{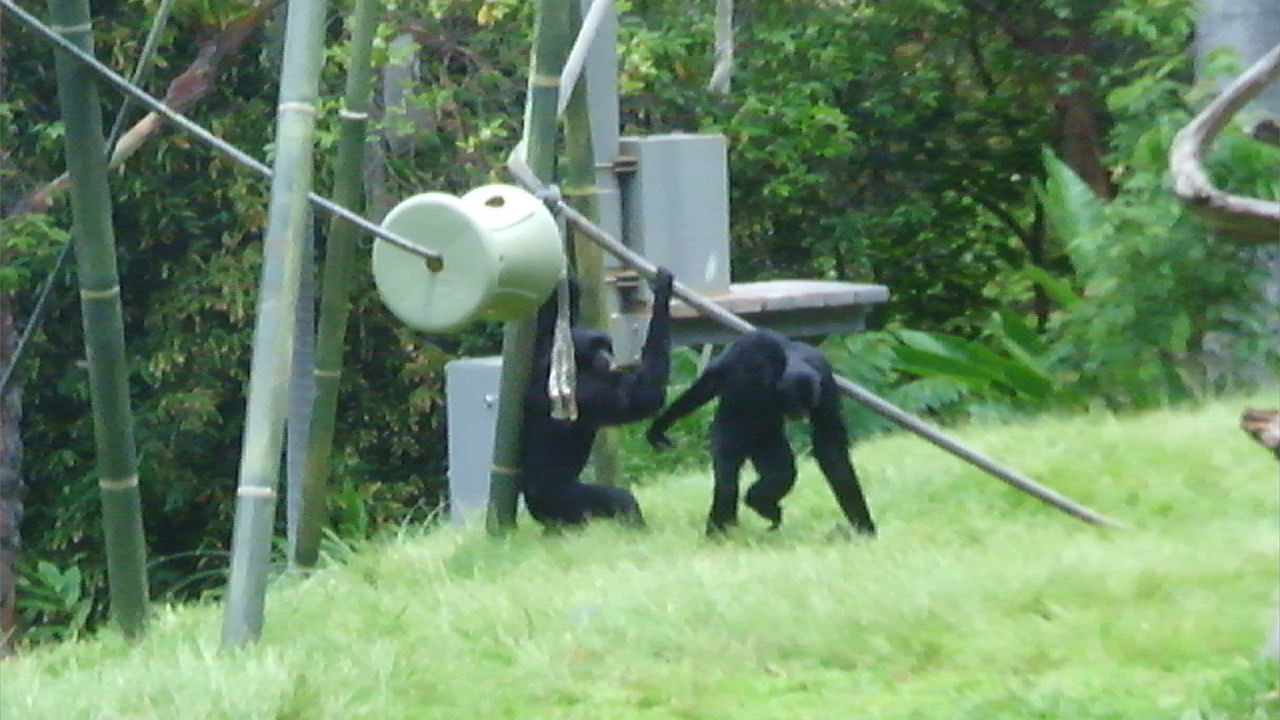}
&\hspace{-.23cm}\includegraphics[width=.21\textwidth,frame]{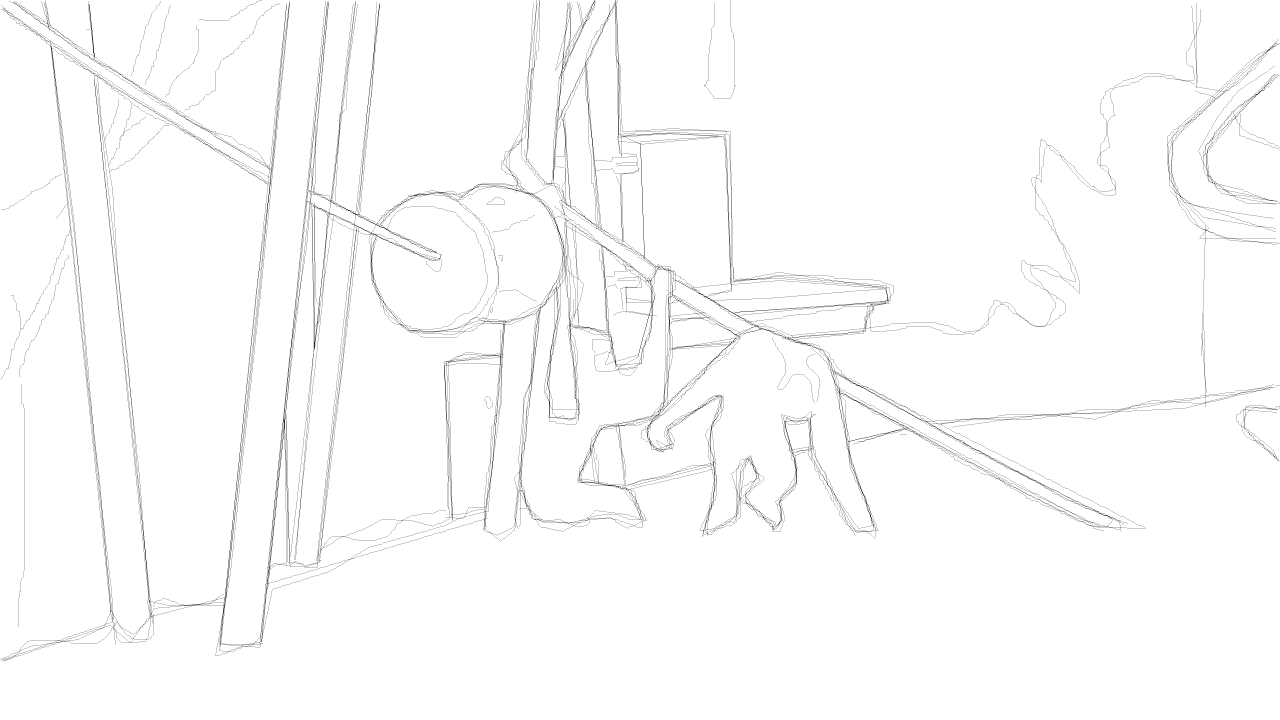}
&\hspace{-.23cm}\includegraphics[width=.21\textwidth,frame]{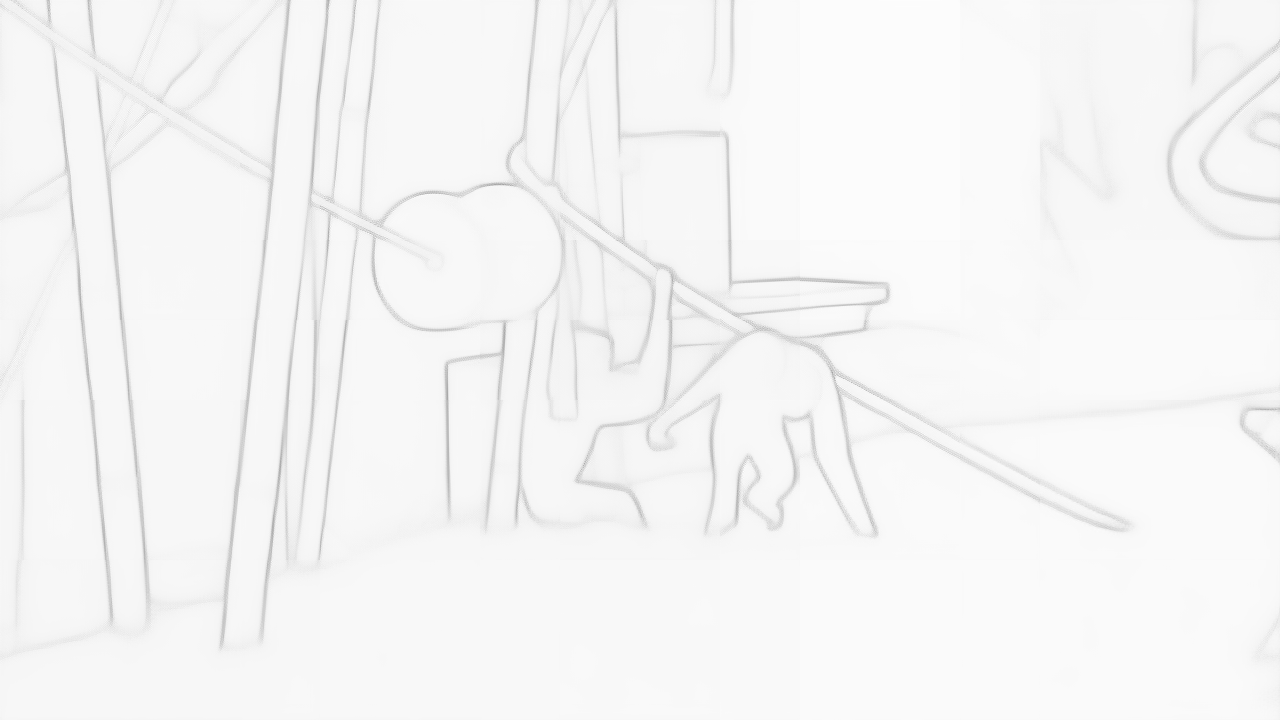}
&\hspace{-.23cm}\includegraphics[width=.21\textwidth,frame]{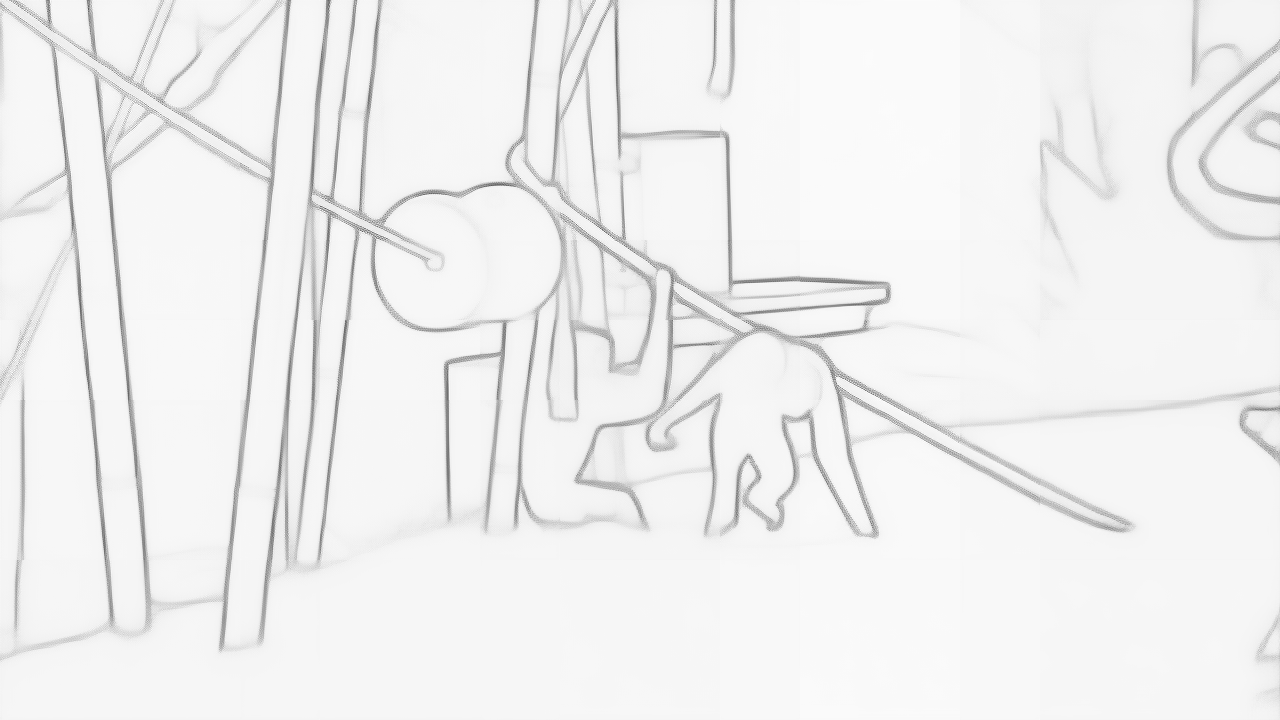}
\vspace{-.06cm}\\

\includegraphics[width=.21\textwidth]{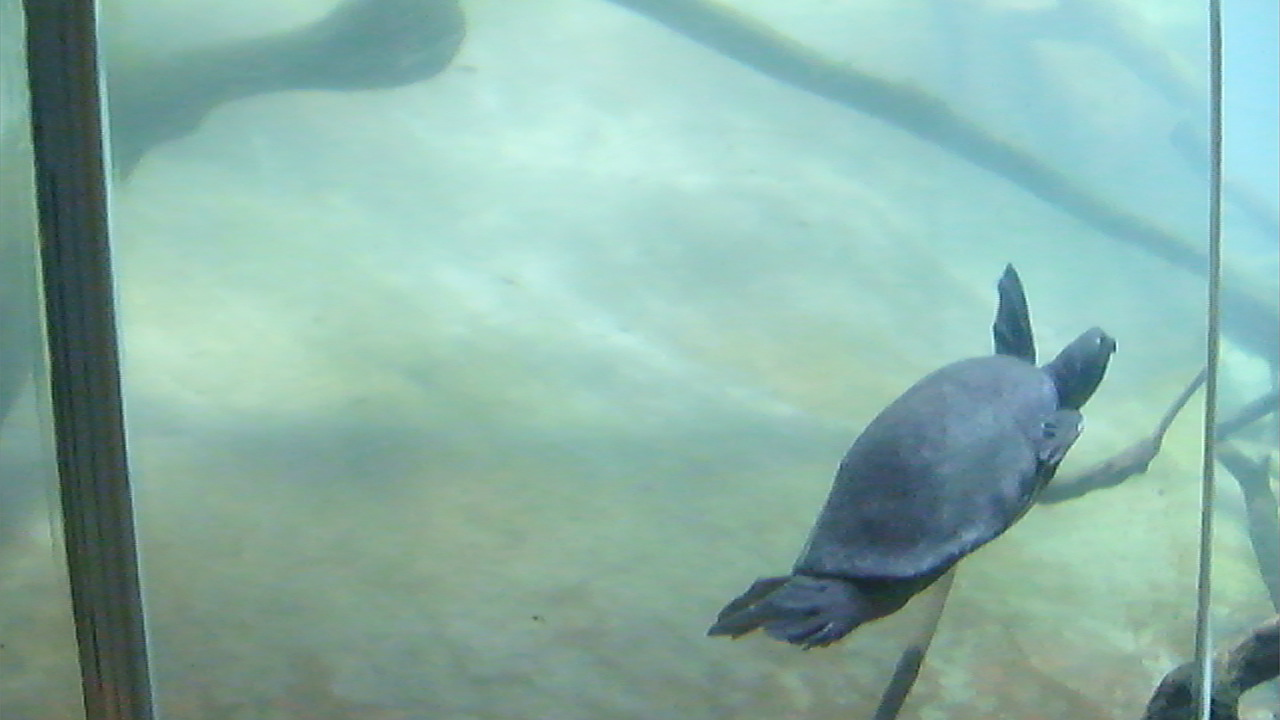}
&\hspace{-.23cm}\includegraphics[width=.21\textwidth,frame]{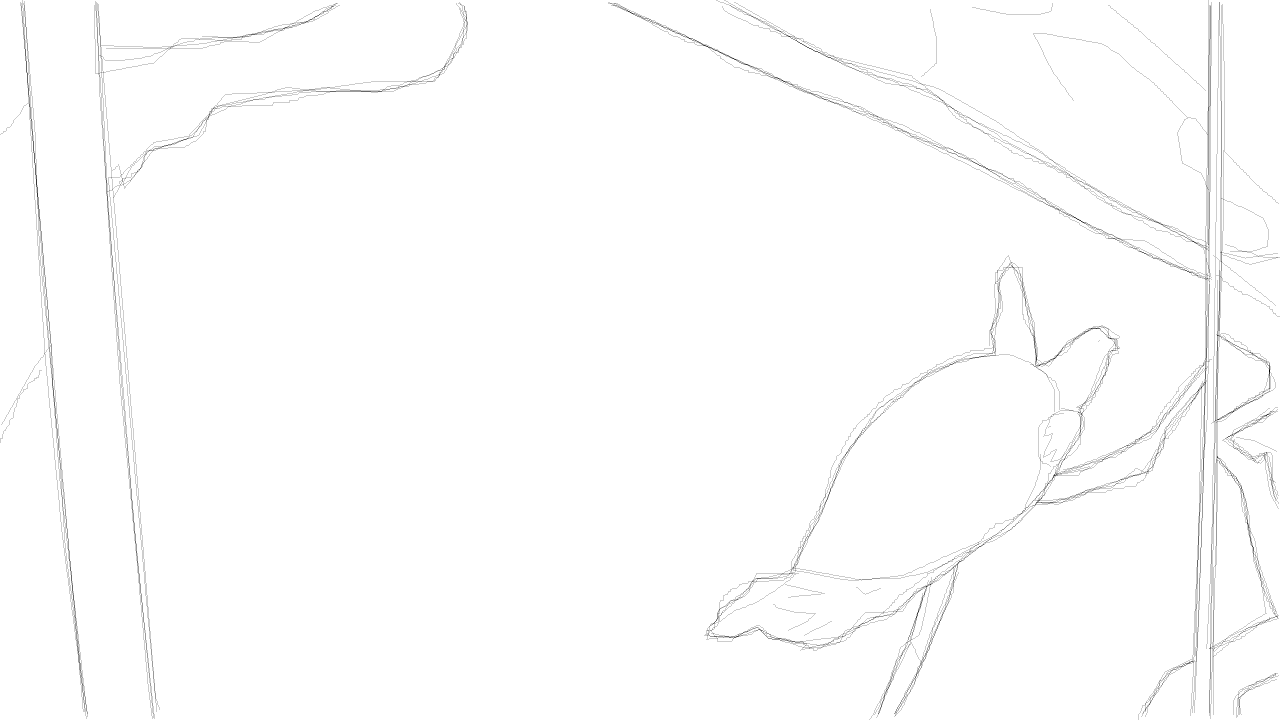}
&\hspace{-.23cm}\includegraphics[width=.21\textwidth,frame]{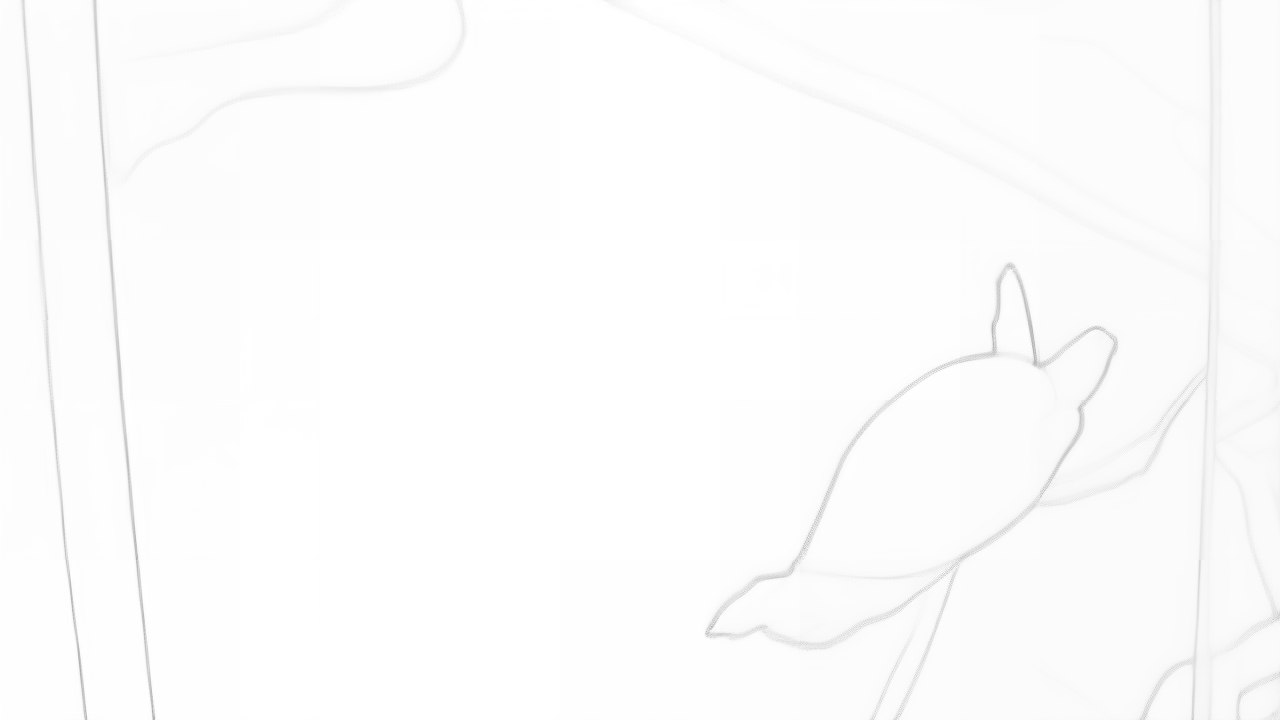}
&\hspace{-.23cm}\includegraphics[width=.21\textwidth,frame]{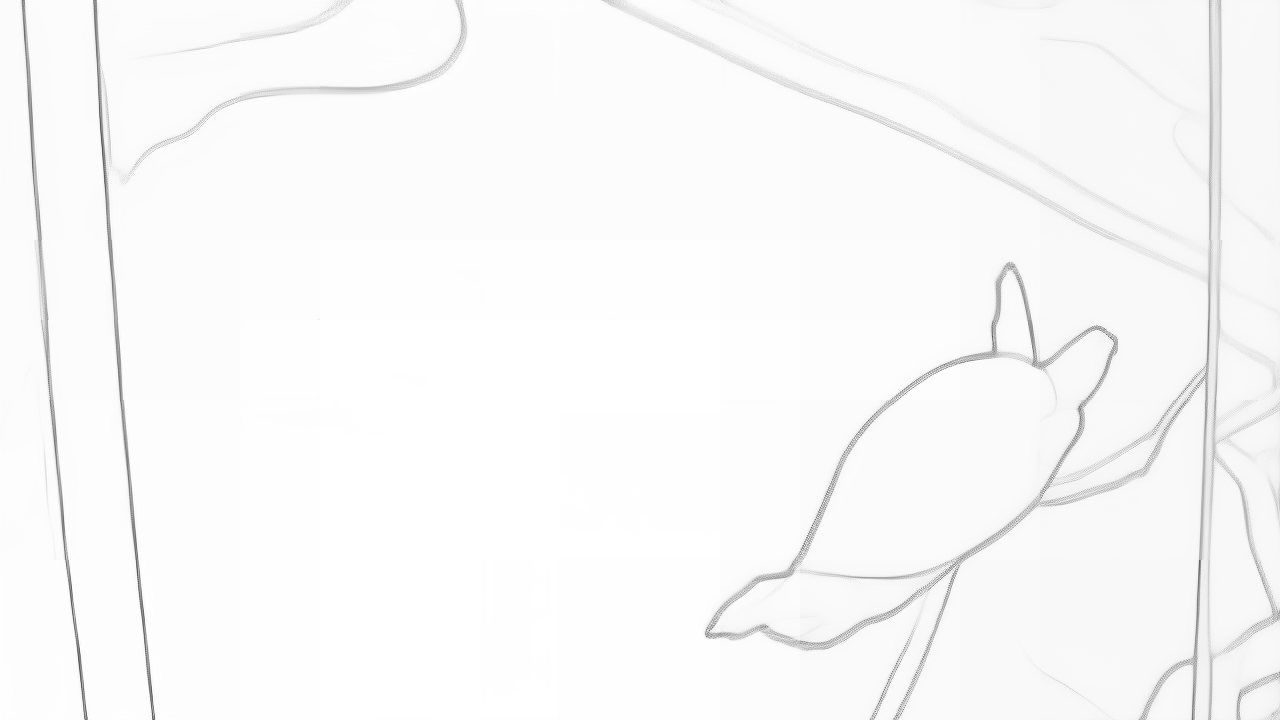}
\vspace{-.06cm}\\
\includegraphics[width=.21\textwidth]{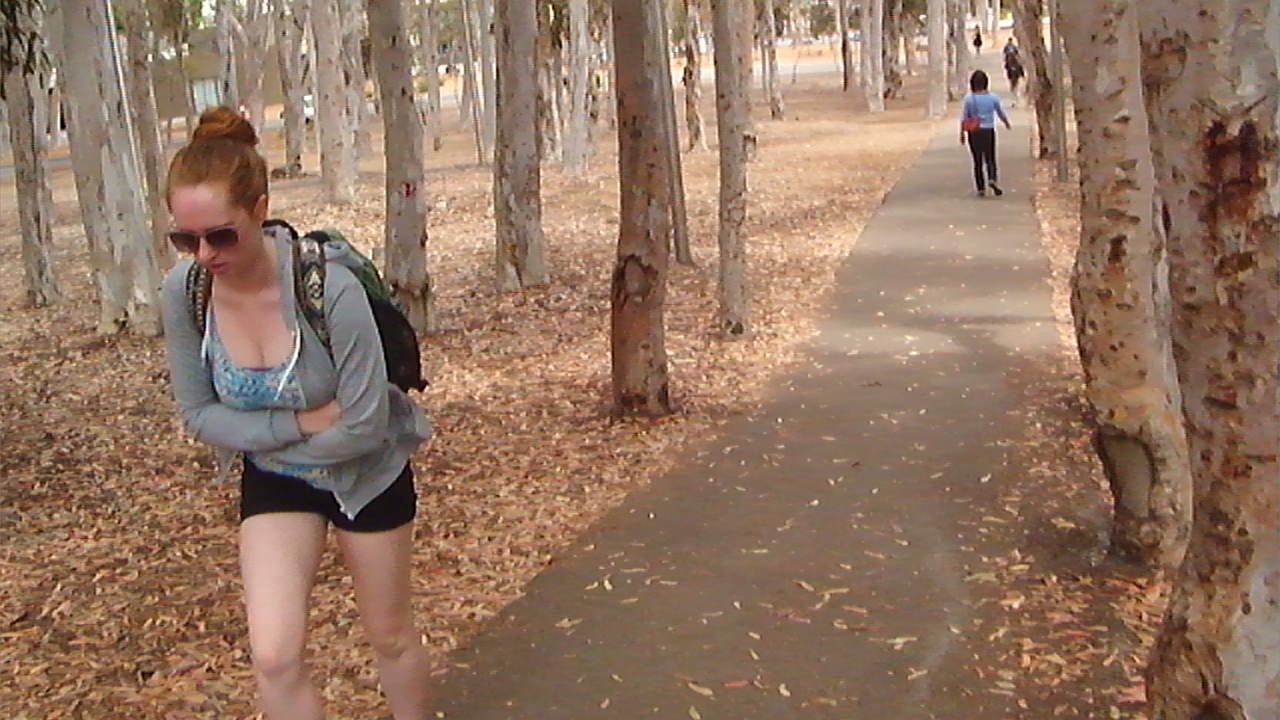}
&\hspace{-.23cm}\includegraphics[width=.21\textwidth,frame]{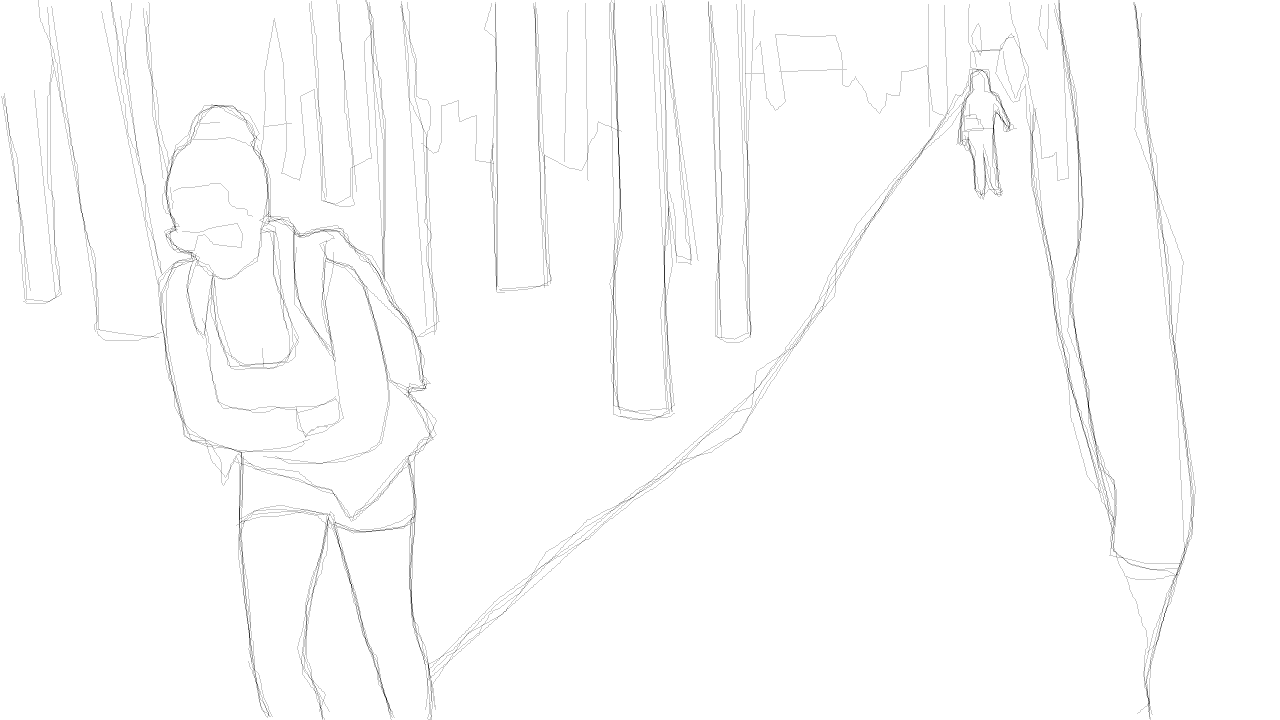}
&\hspace{-.23cm}\includegraphics[width=.21\textwidth,frame]{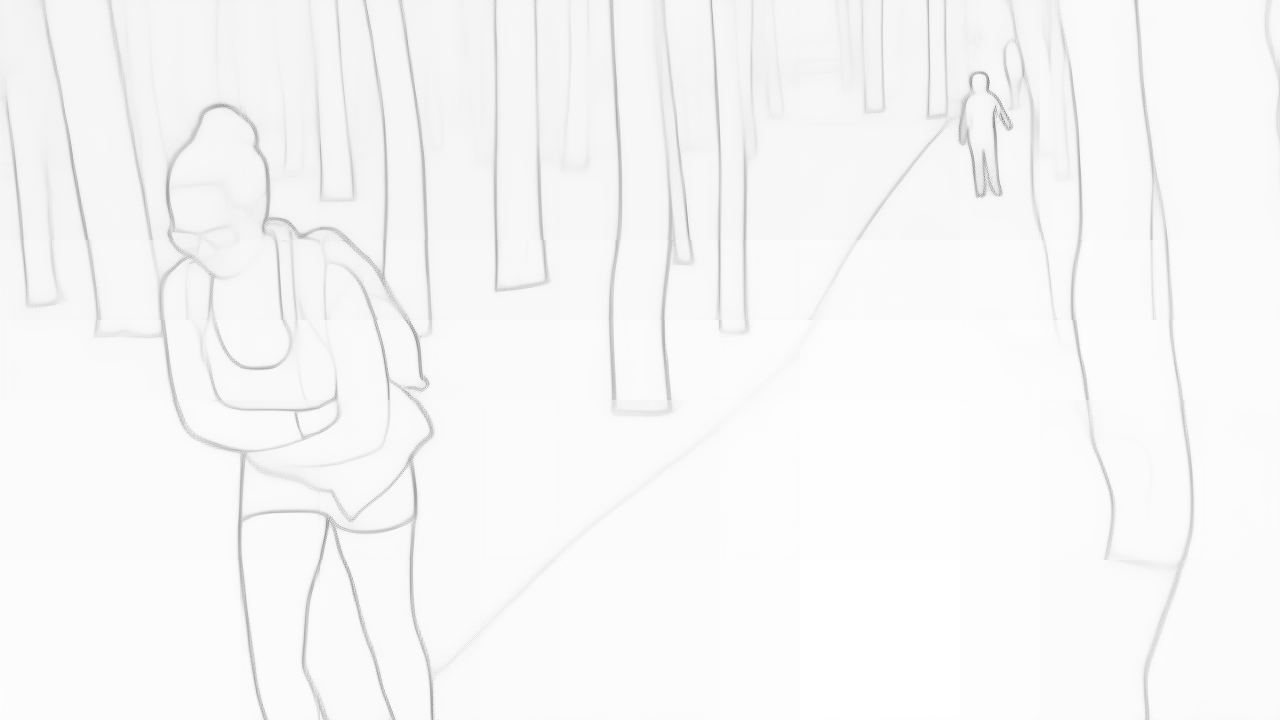}
&\hspace{-.23cm}\includegraphics[width=.21\textwidth,frame]{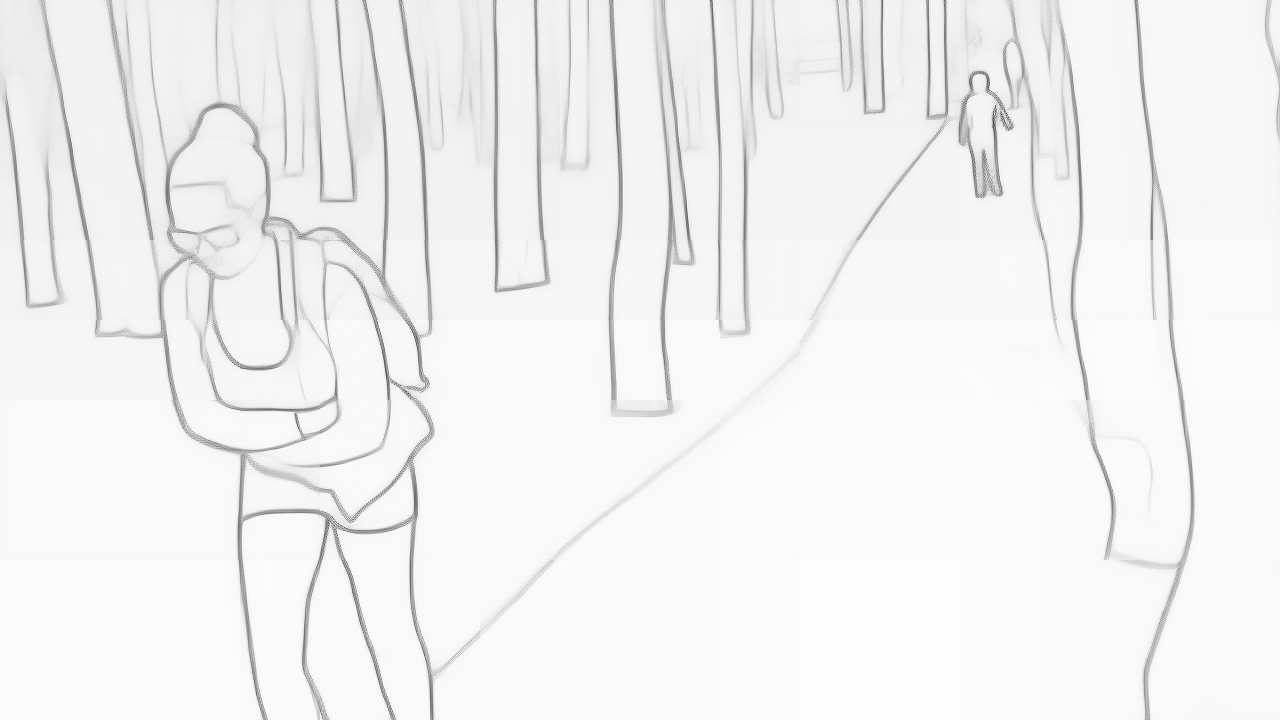}
\vspace{-.06cm}\\

\hspace{-.23cm} Input & \hspace{-.23cm} GT-Boundary &\hspace{-.23cm}  Ours-Boundary (0) &\hspace{-.23cm}  Ours-Boundary (1) \\
\end{tabular}
\caption{Qualitative results with different edge granularity on the Multicue~\cite{mely2016systematic} Boundary.}
\label{fig_multicue_boundary}
\end{figure*}

\begin{figure*}[!t]
\small
\setlength{\abovecaptionskip}{3pt}
\setlength{\belowcaptionskip}{0pt}
\centering
\begin{tabular}{cccc}
\hspace{-.23cm}
\includegraphics[width=.21\textwidth]{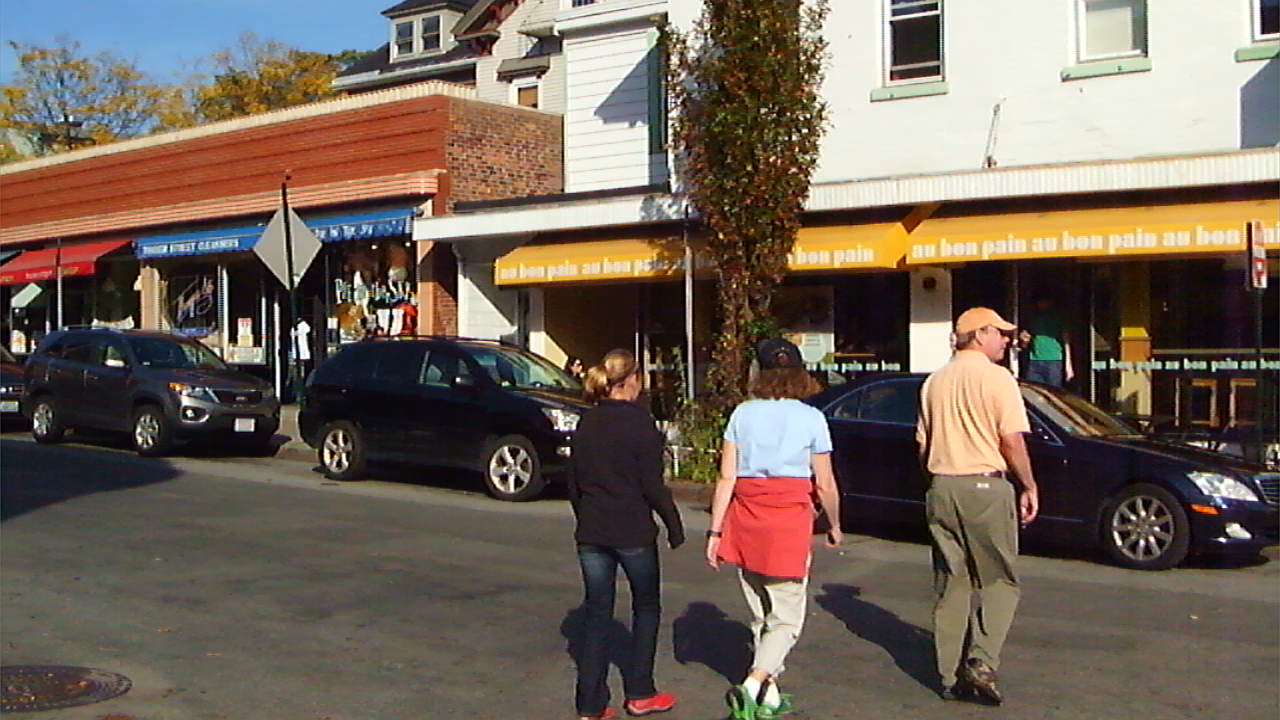}
&\hspace{-.23cm}\includegraphics[width=.21\textwidth,frame]{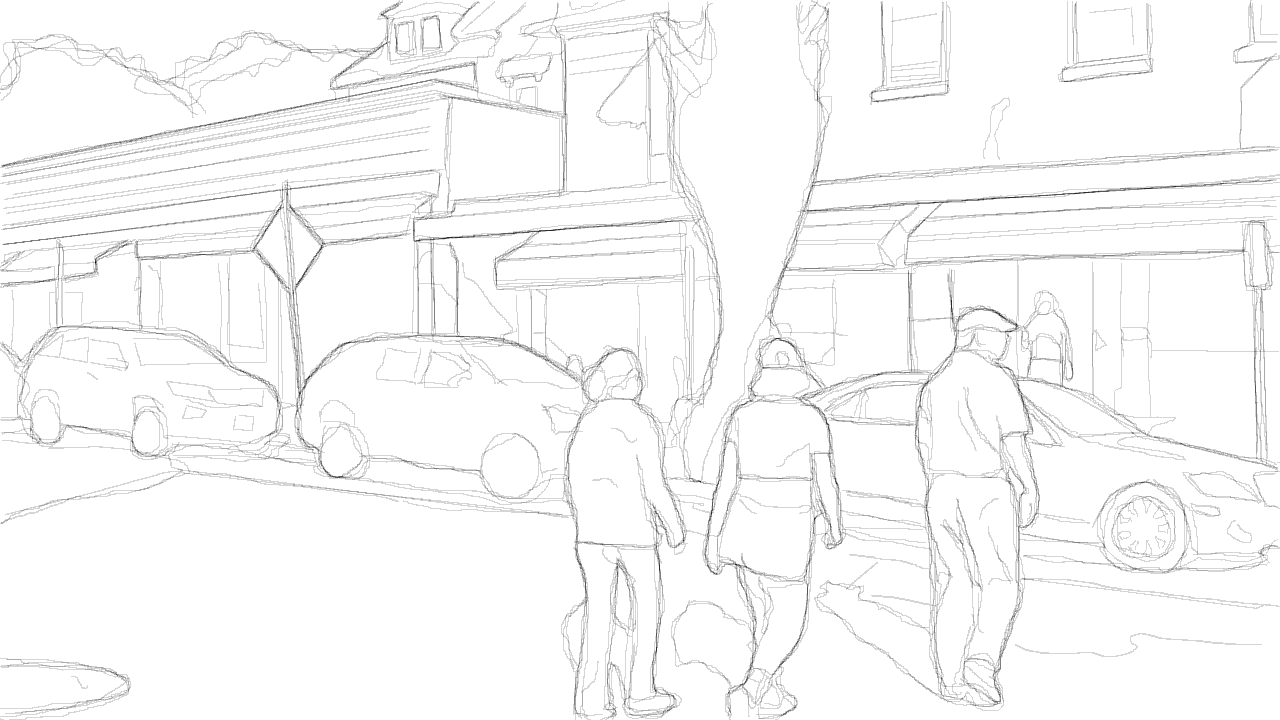}
&\hspace{-.23cm}\includegraphics[width=.21\textwidth,frame]{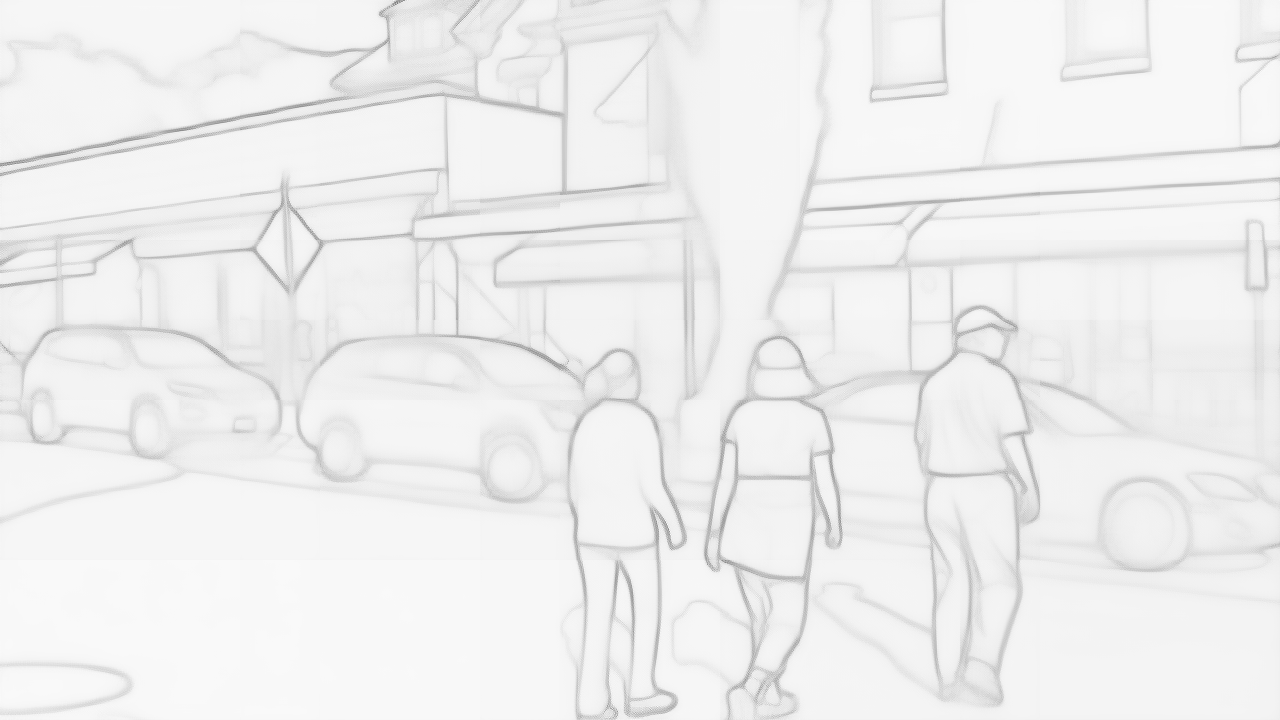}
&\hspace{-.23cm}\includegraphics[width=.21\textwidth,frame]{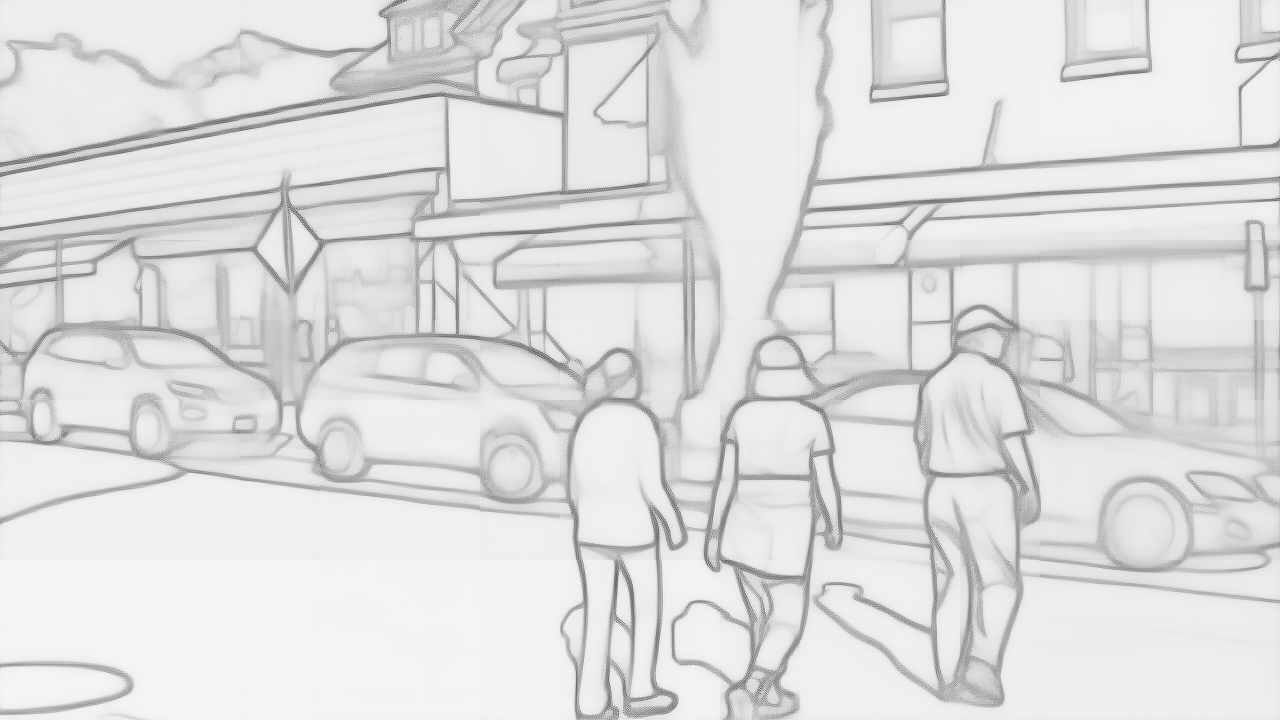}
\vspace{-.06cm}\\

\includegraphics[width=.21\textwidth]{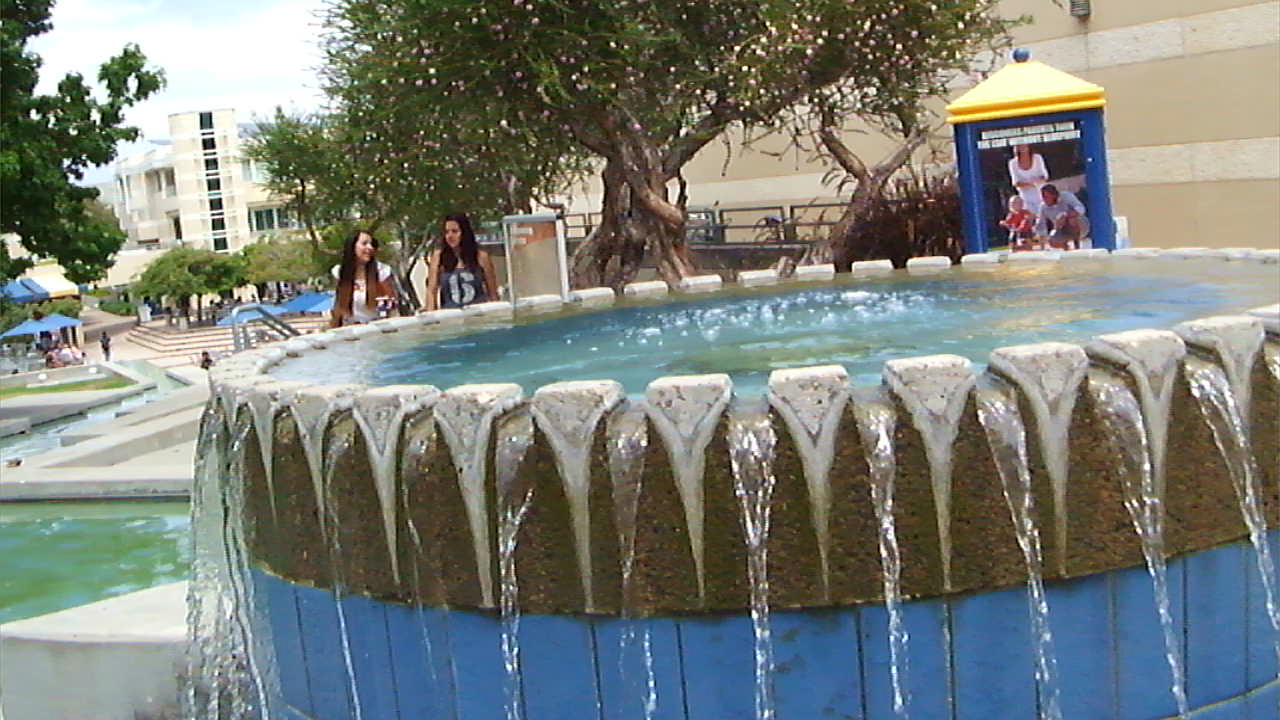}
&\hspace{-.23cm}\includegraphics[width=.21\textwidth,frame]{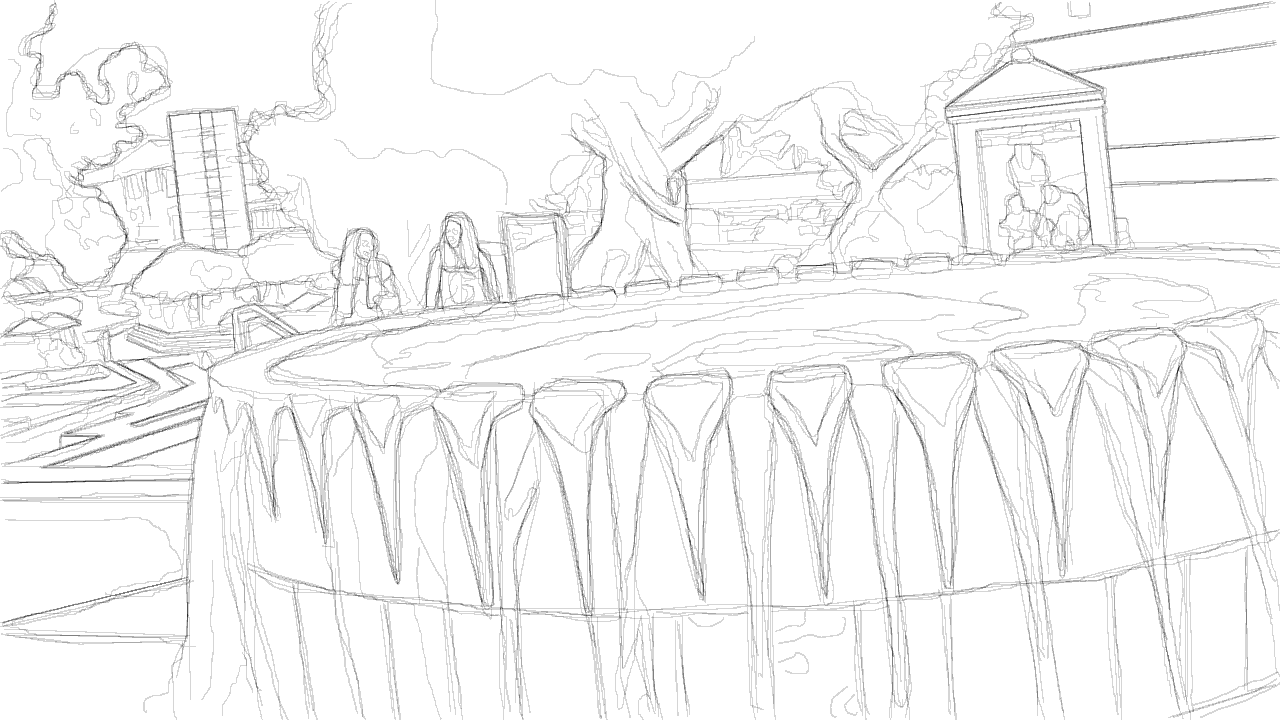}
&\hspace{-.23cm}\includegraphics[width=.21\textwidth,frame]{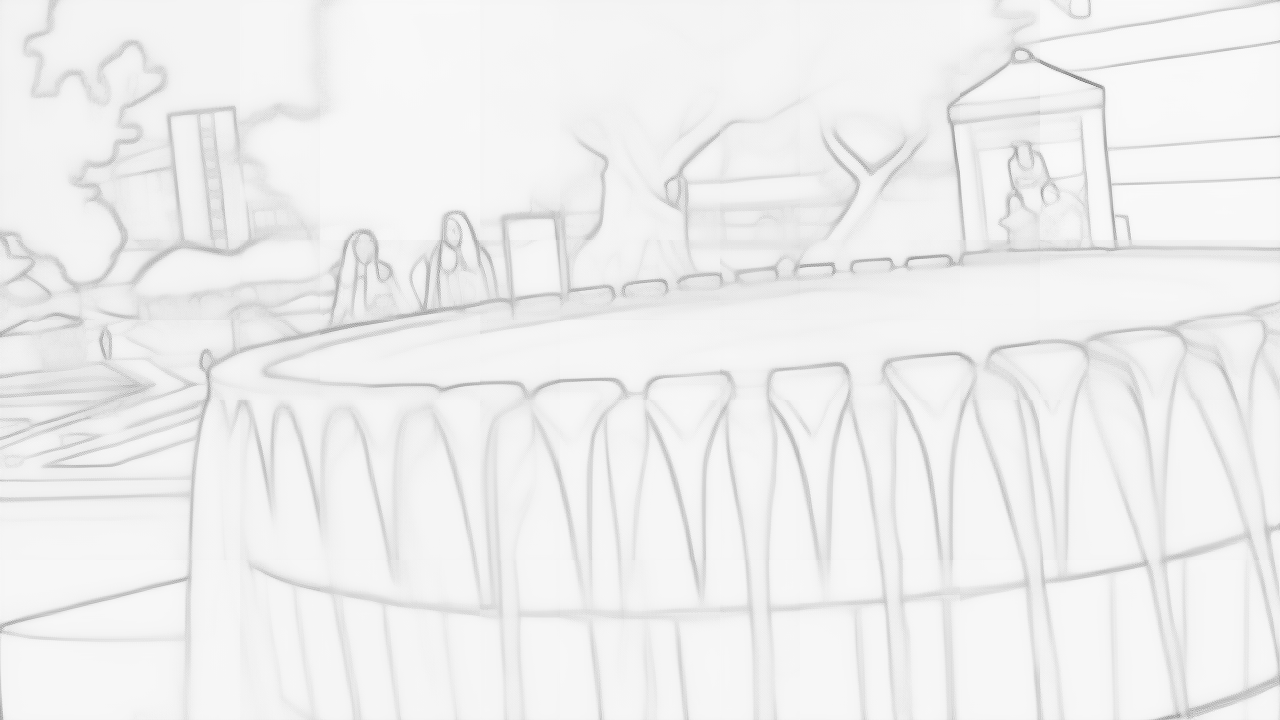}
&\hspace{-.23cm}\includegraphics[width=.21\textwidth,frame]{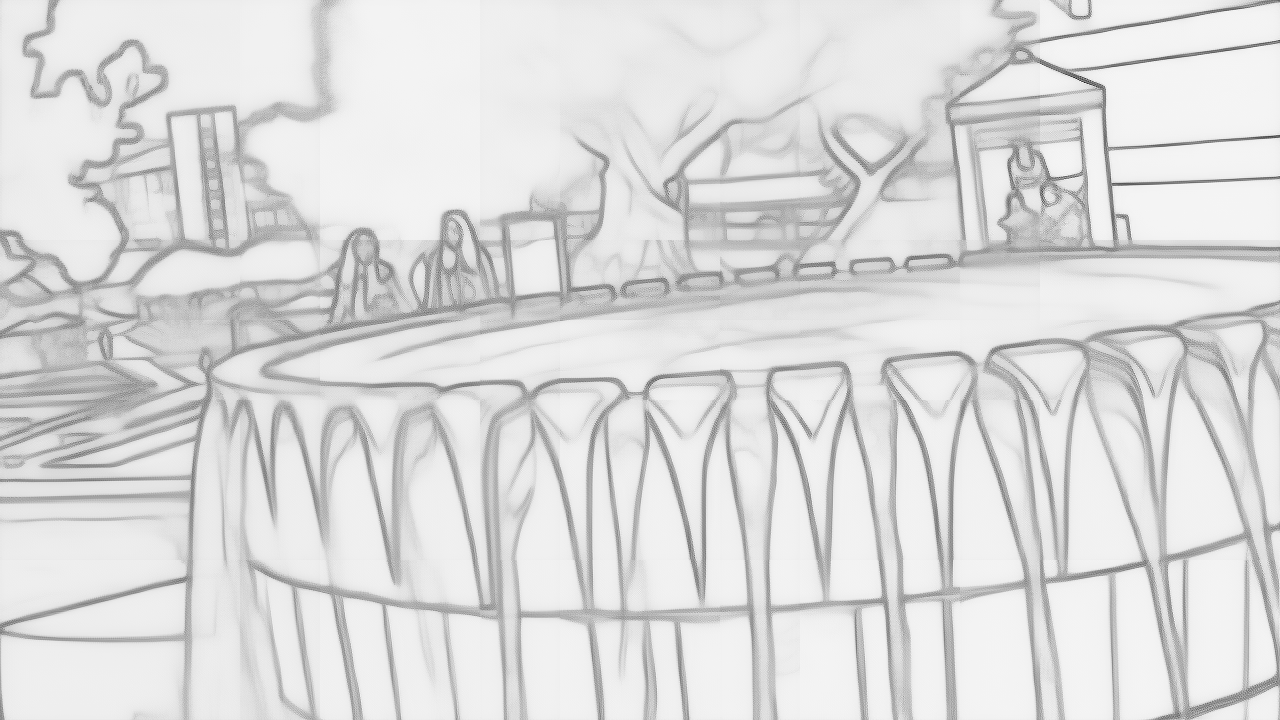}
\vspace{-.06cm}\\
\includegraphics[width=.21\textwidth]{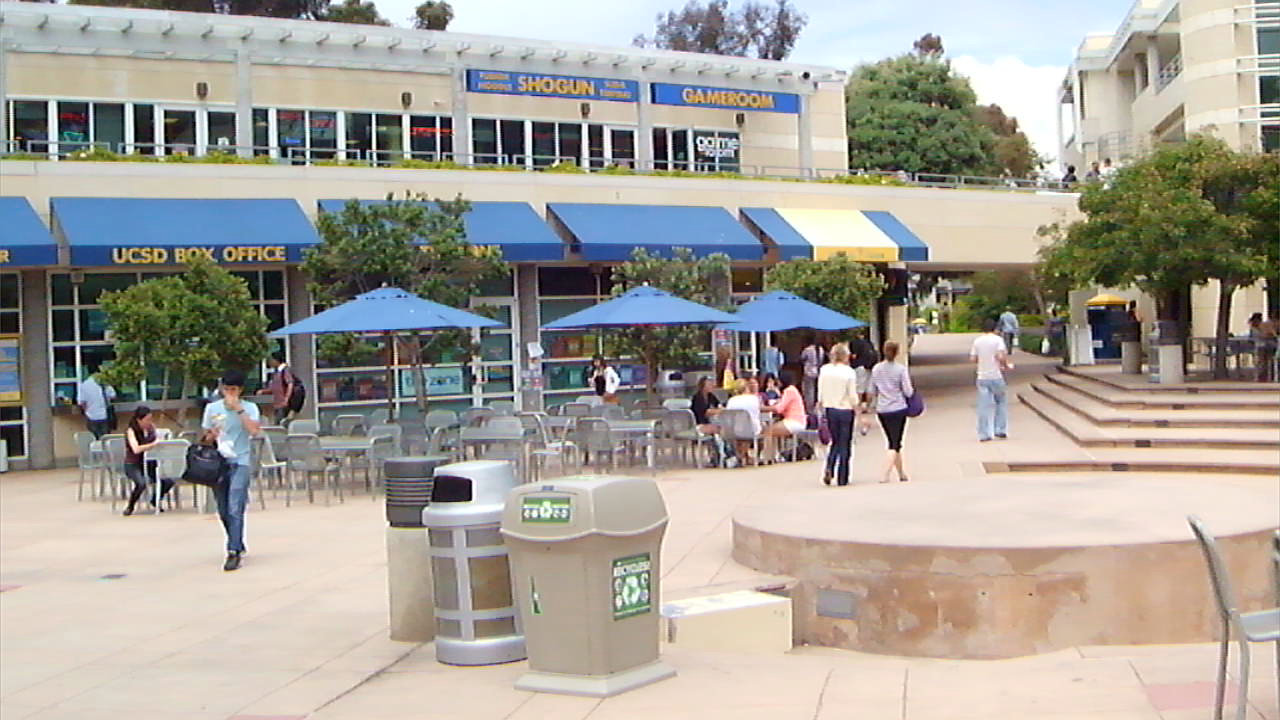}
&\hspace{-.23cm}\includegraphics[width=.21\textwidth,frame]{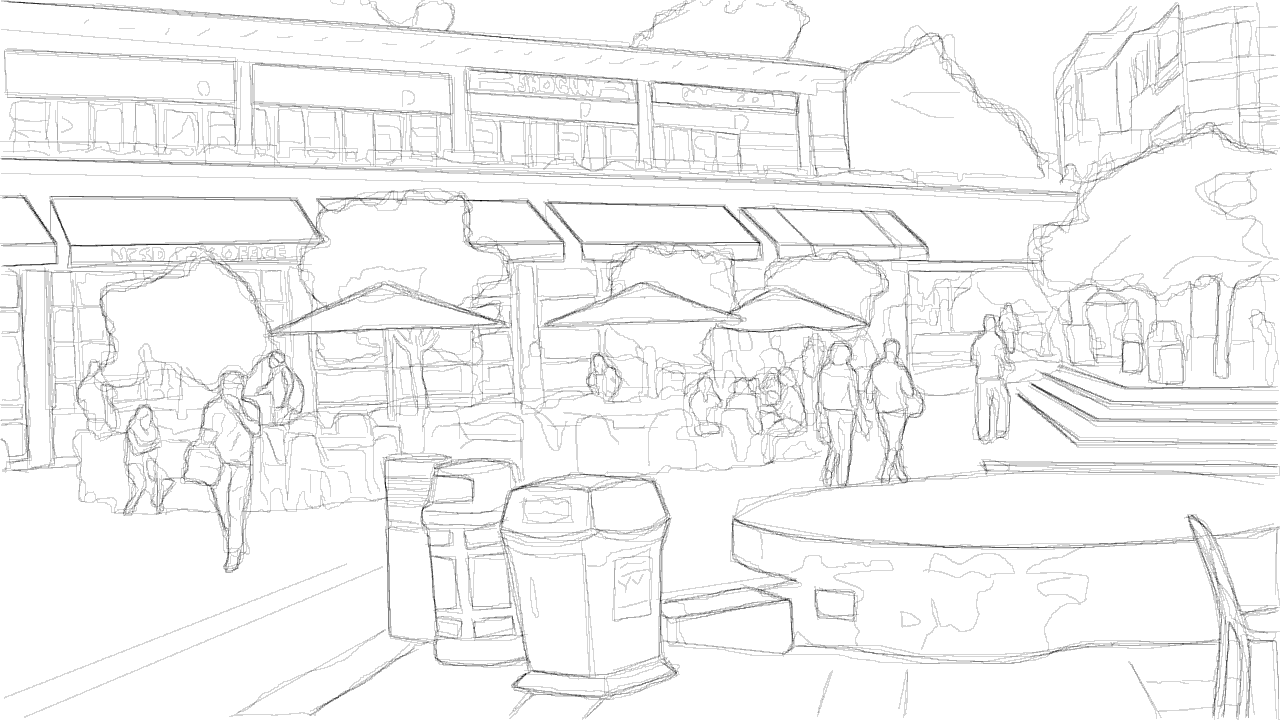}
&\hspace{-.23cm}\includegraphics[width=.21\textwidth,frame]{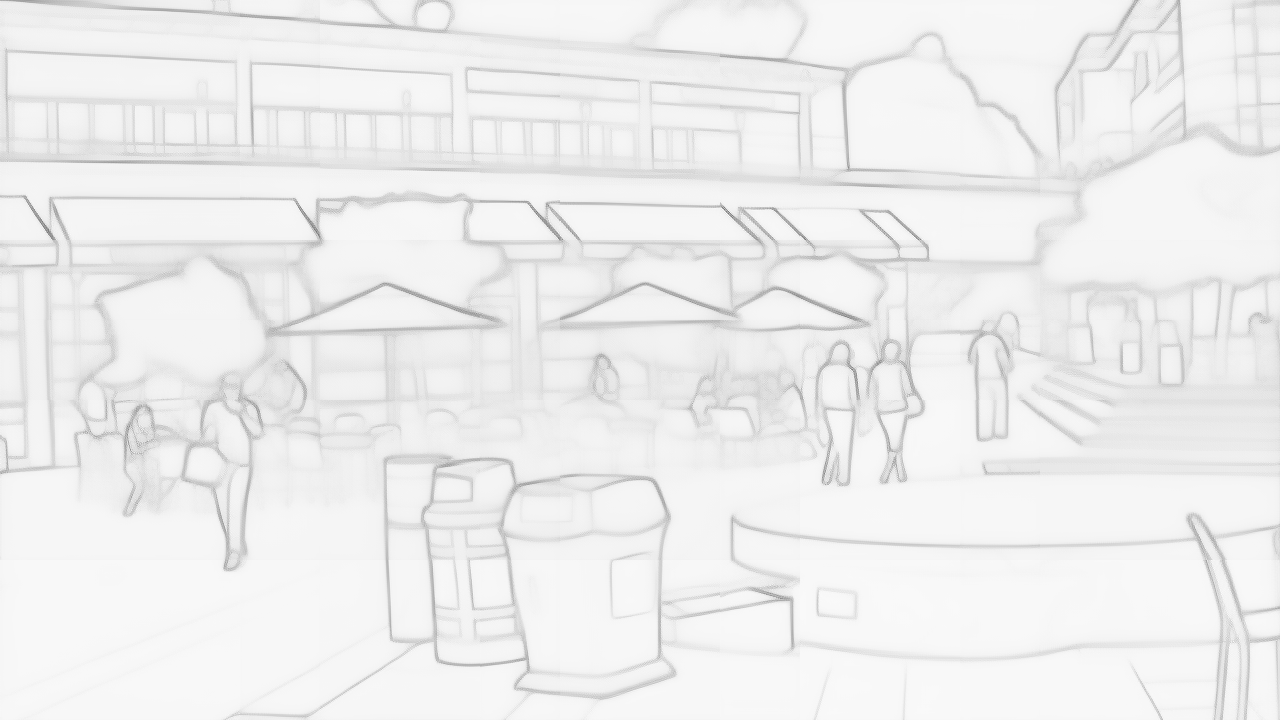}
&\hspace{-.23cm}\includegraphics[width=.21\textwidth,frame]{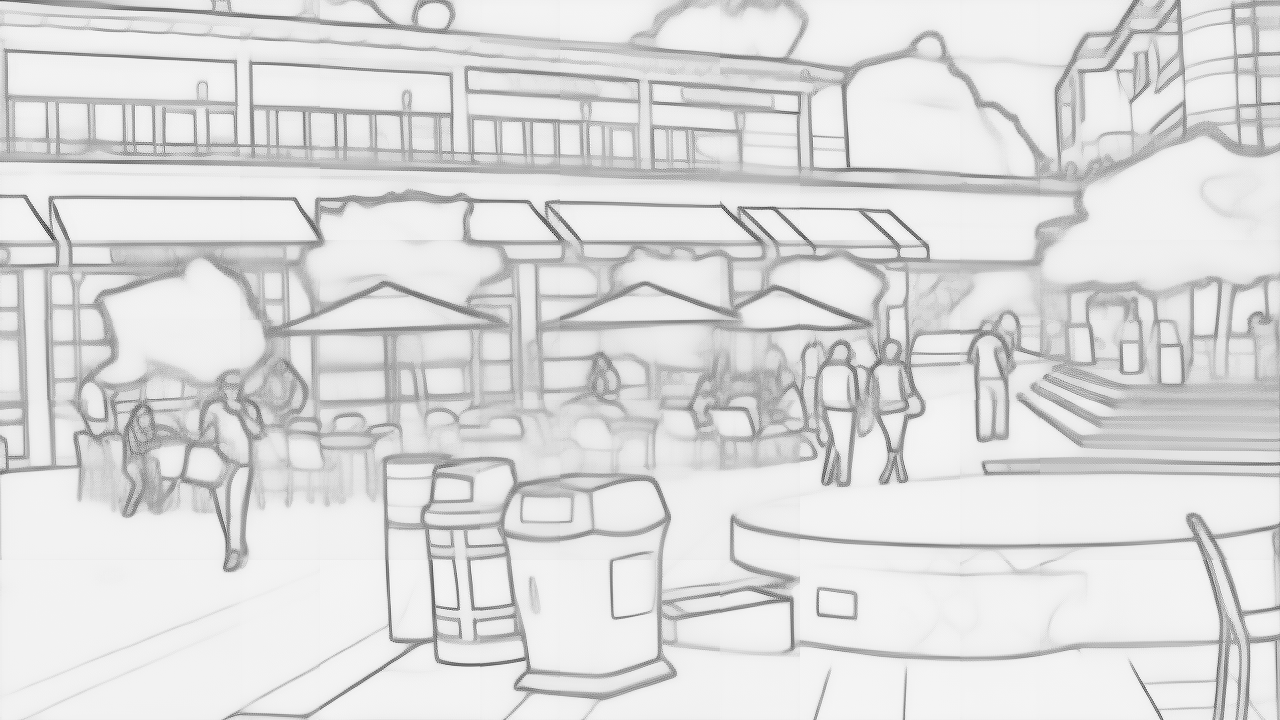}
\vspace{-.06cm}\\
\hspace{-.23cm} Input & \hspace{-.23cm} GT-Edge &\hspace{-.23cm}  Ours-Edge (0) &\hspace{-.5cm}  Ours-Edge (1) \\
\end{tabular}
\caption{Qualitative results with different edge granularity on the Multicue~\cite{mely2016systematic} Edge.}
\label{fig_multicue_edge}
\end{figure*}

\begin{figure*}[!t]
\small
\setlength{\abovecaptionskip}{3pt}
\setlength{\belowcaptionskip}{0pt}
\centering
\begin{tabular}{ccc}
\hspace{-.23cm}
\includegraphics[width=.28\textwidth]{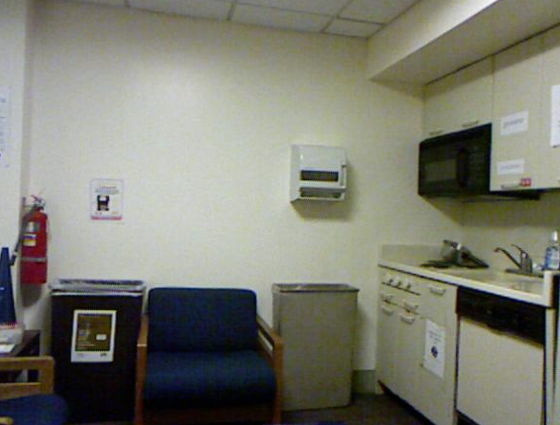}
&\hspace{-.23cm}\includegraphics[width=.28\textwidth,frame]{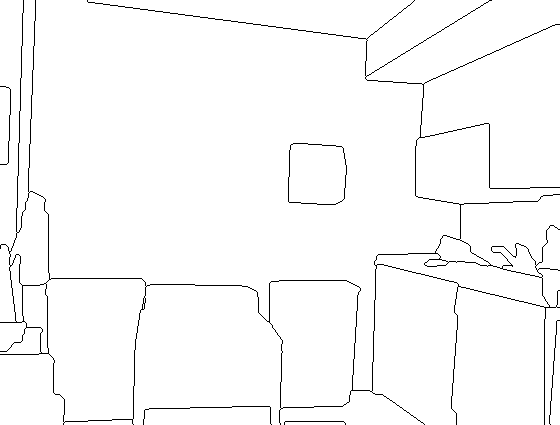}
&\hspace{-.23cm}\includegraphics[width=.28\textwidth,frame]{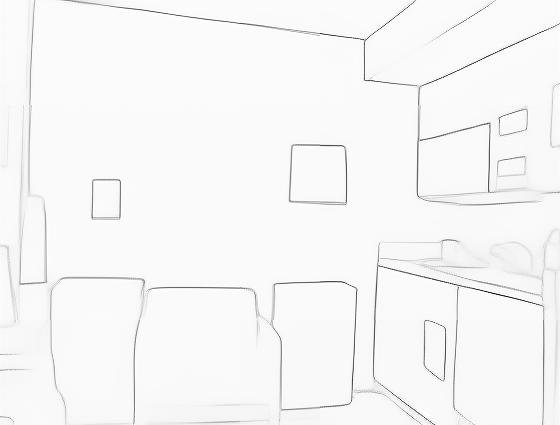}
\vspace{-.06cm}\\
\hspace{-.23cm}
\includegraphics[width=.28\textwidth]{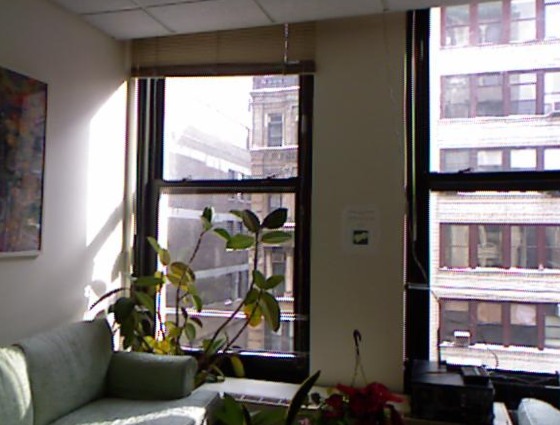}
&\hspace{-.23cm}\includegraphics[width=.28\textwidth,frame]{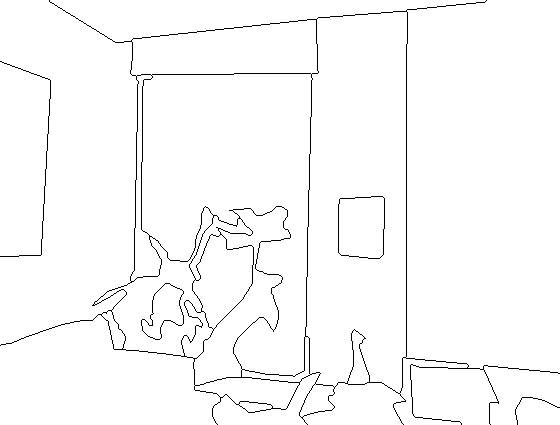}
&\hspace{-.23cm}\includegraphics[width=.28\textwidth,frame]{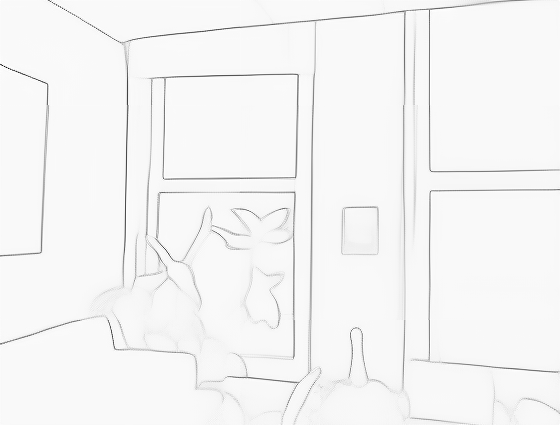}
\vspace{-.06cm}\\
\hspace{-.23cm}
\includegraphics[width=.28\textwidth]{Fig/supp/nyud/image/img_5021.png}
&\hspace{-0.2cm}\includegraphics[width=.28\textwidth,frame]{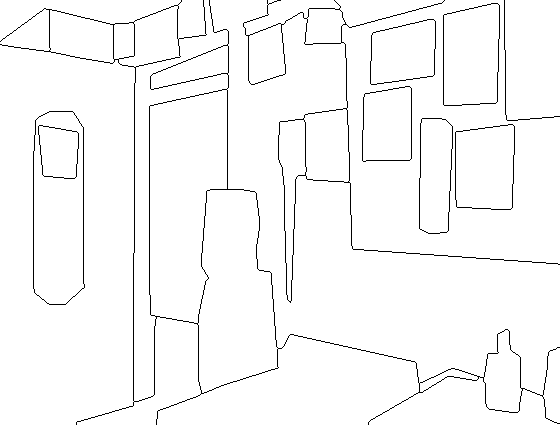}
&\hspace{-0.2cm}\includegraphics[width=.28\textwidth,frame]{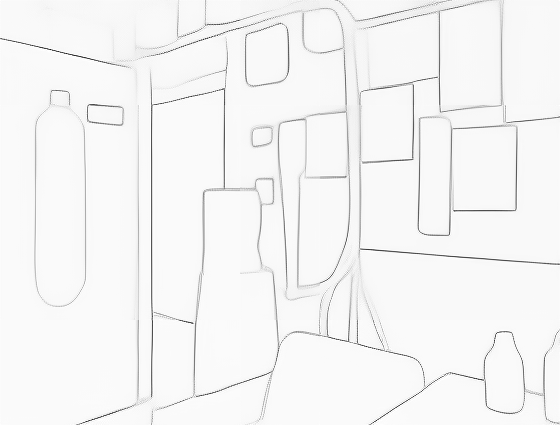}
\vspace{-.06cm}\\
\hspace{-.23cm} Input & \hspace{-.23cm} GT &\hspace{-.23cm}  Ours  \\
\end{tabular}
\caption{Qualitative results on the NYUD~\cite{silberman2012indoor} dataset.}
\label{fig_nyud}
\end{figure*}

\begin{figure*}[!t]
\setlength{\abovecaptionskip}{3pt}
\setlength{\belowcaptionskip}{0pt}
\centering
\begin{tabular}{ccc}
\hspace{-.23cm}
\includegraphics[width=.28\textwidth]{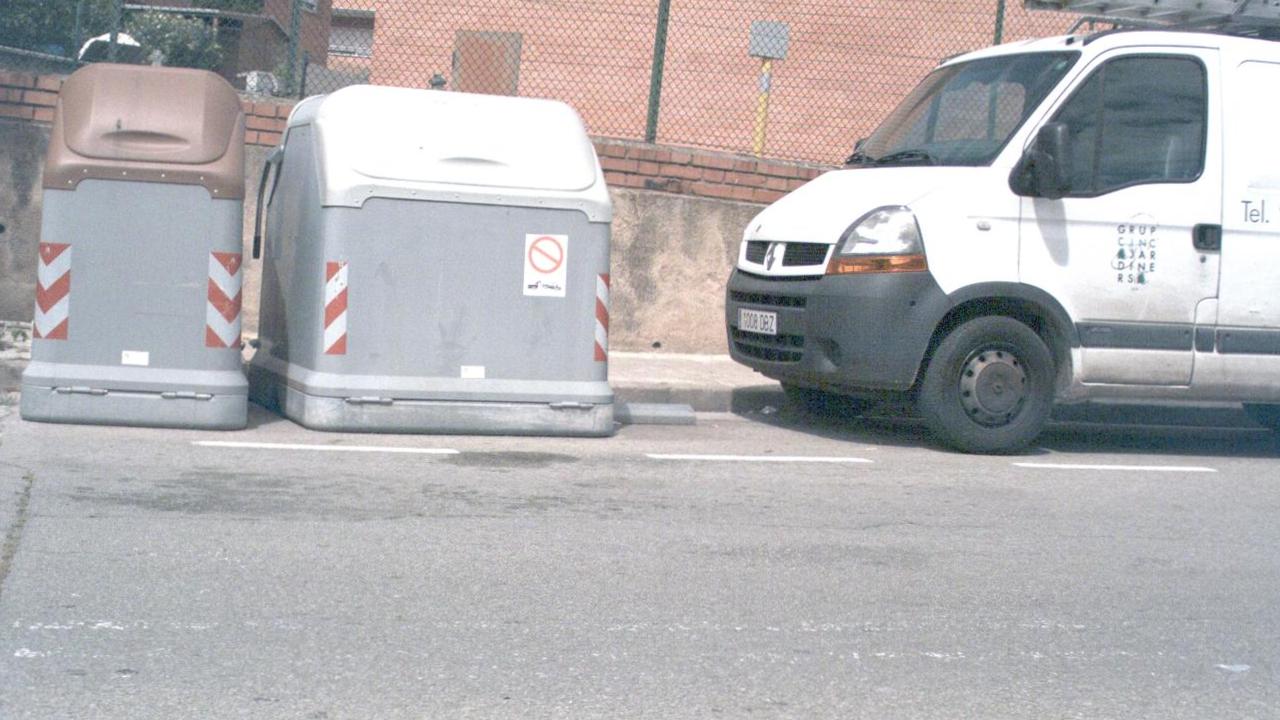}
&\hspace{-.23cm}\includegraphics[width=.28\textwidth,frame]{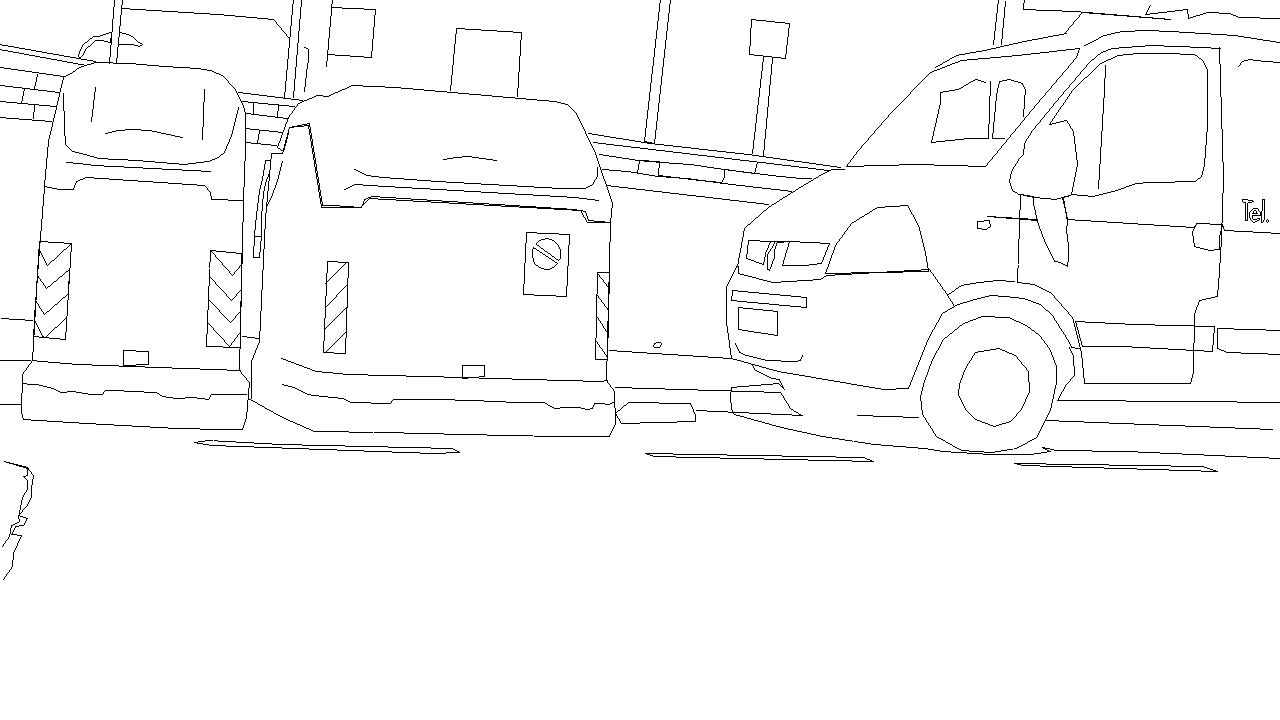}
&\hspace{-.23cm}\includegraphics[width=.28\textwidth,frame]{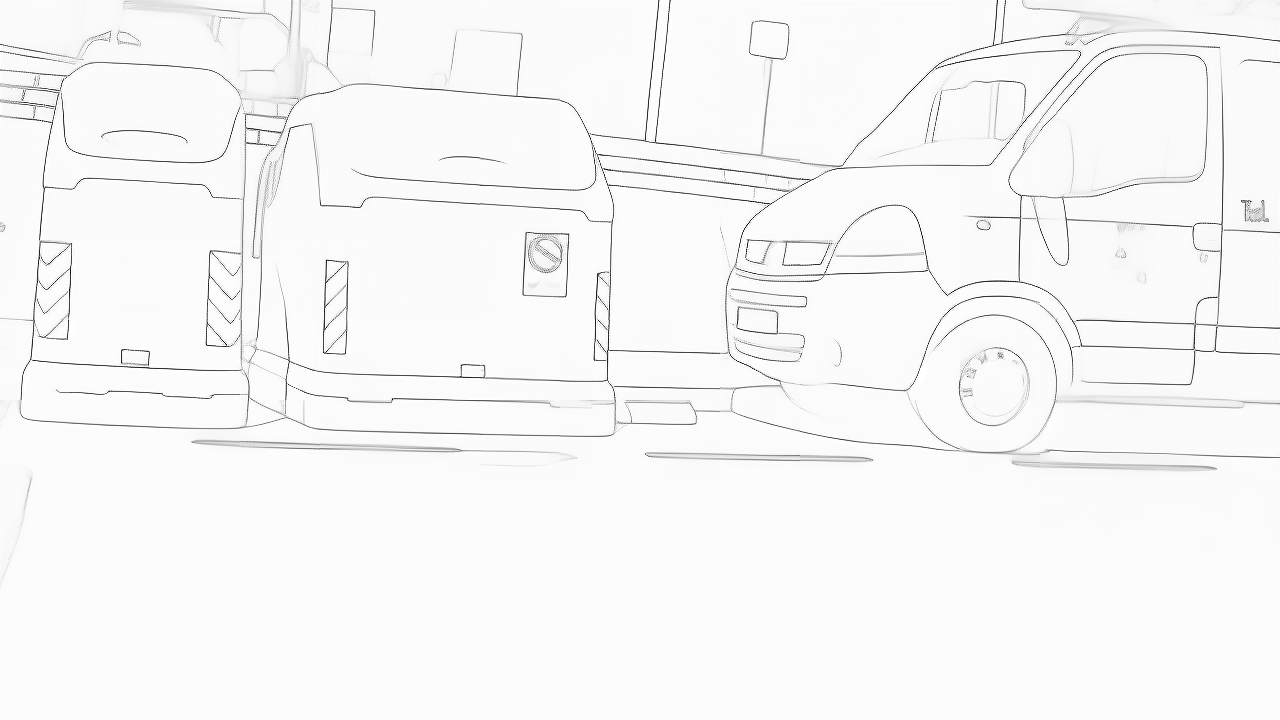}
\vspace{-.06cm}\\

\includegraphics[width=.28\textwidth]{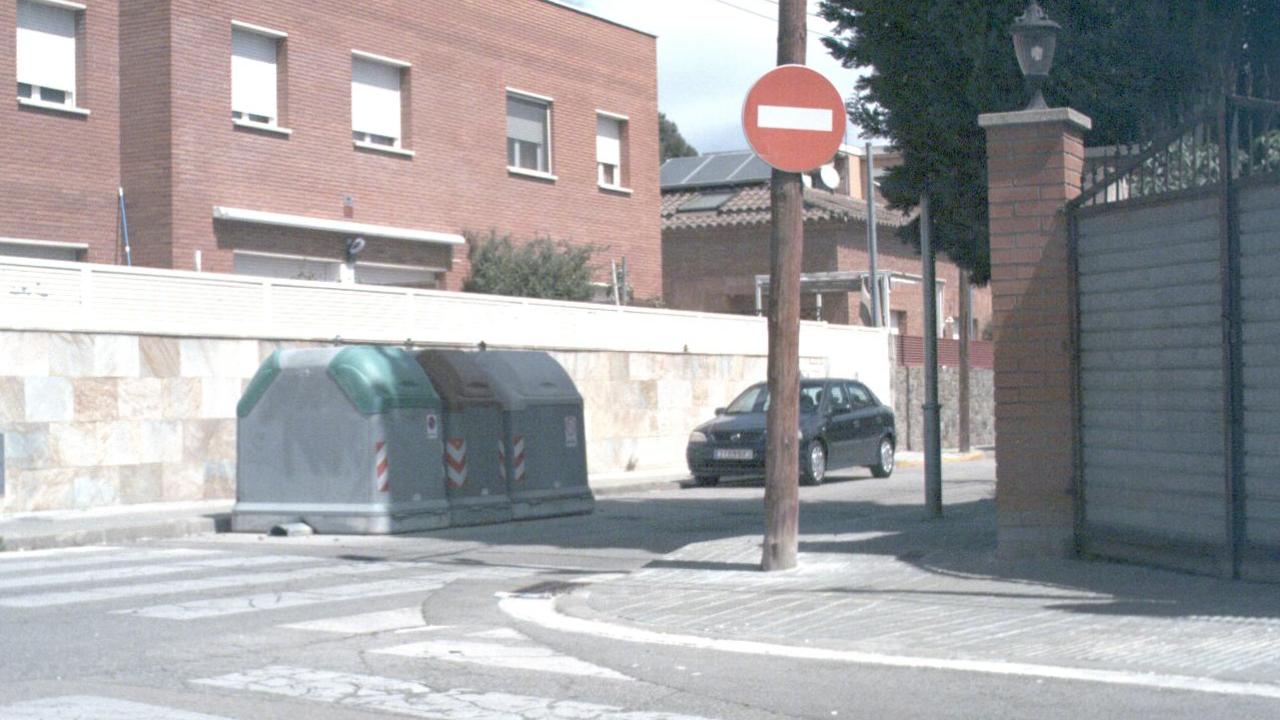}
&\hspace{-.23cm}\includegraphics[width=.28\textwidth,frame]{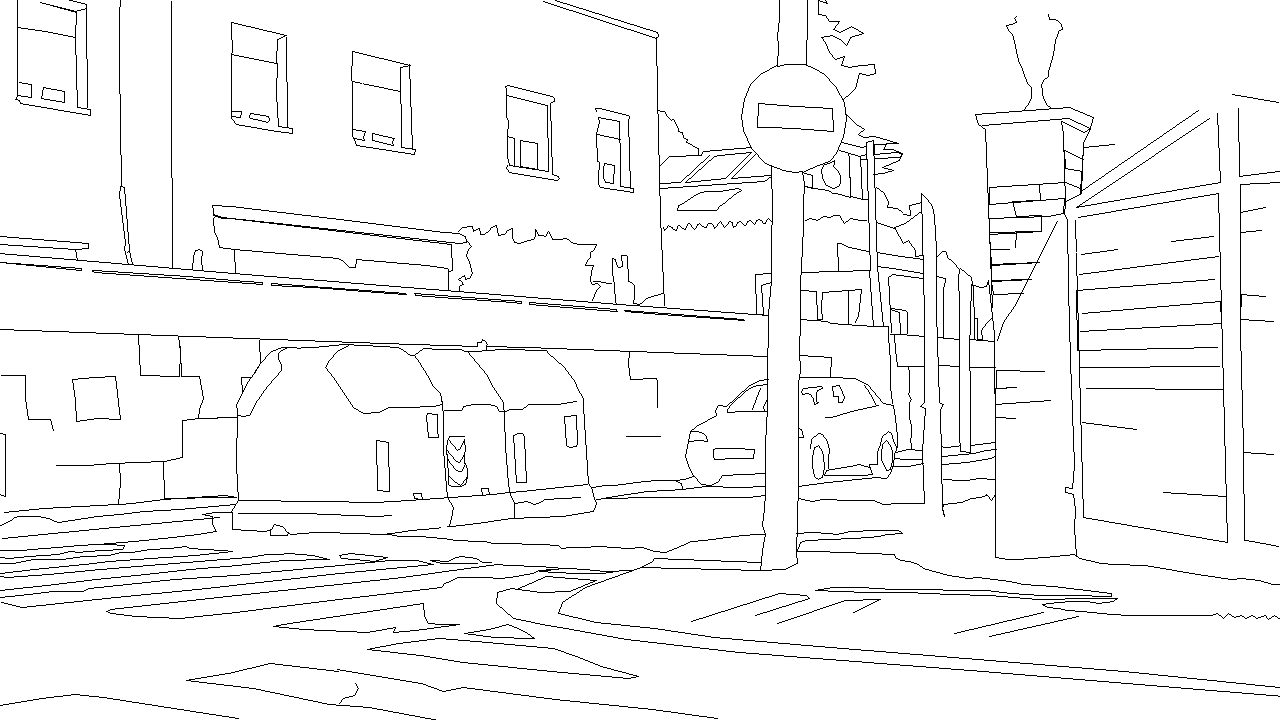}
&\hspace{-.23cm}\includegraphics[width=.28\textwidth,frame]{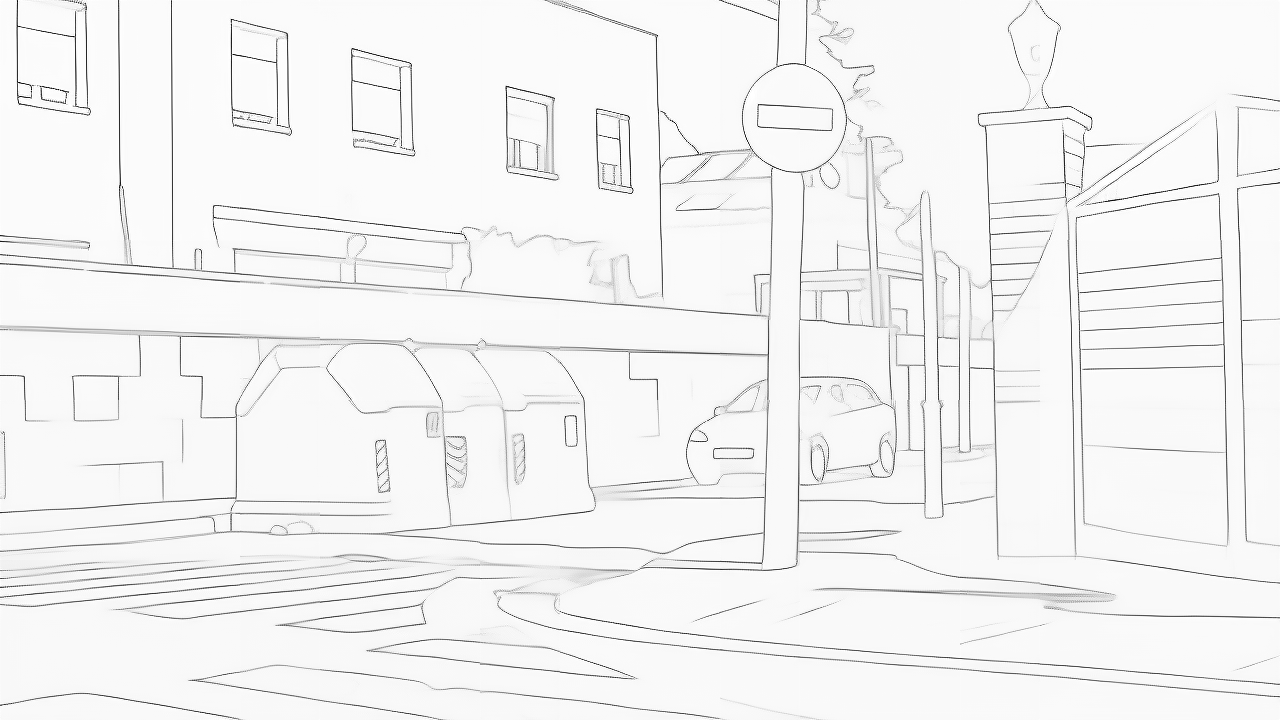}
\vspace{-.06cm}\\

\includegraphics[width=.28\textwidth]{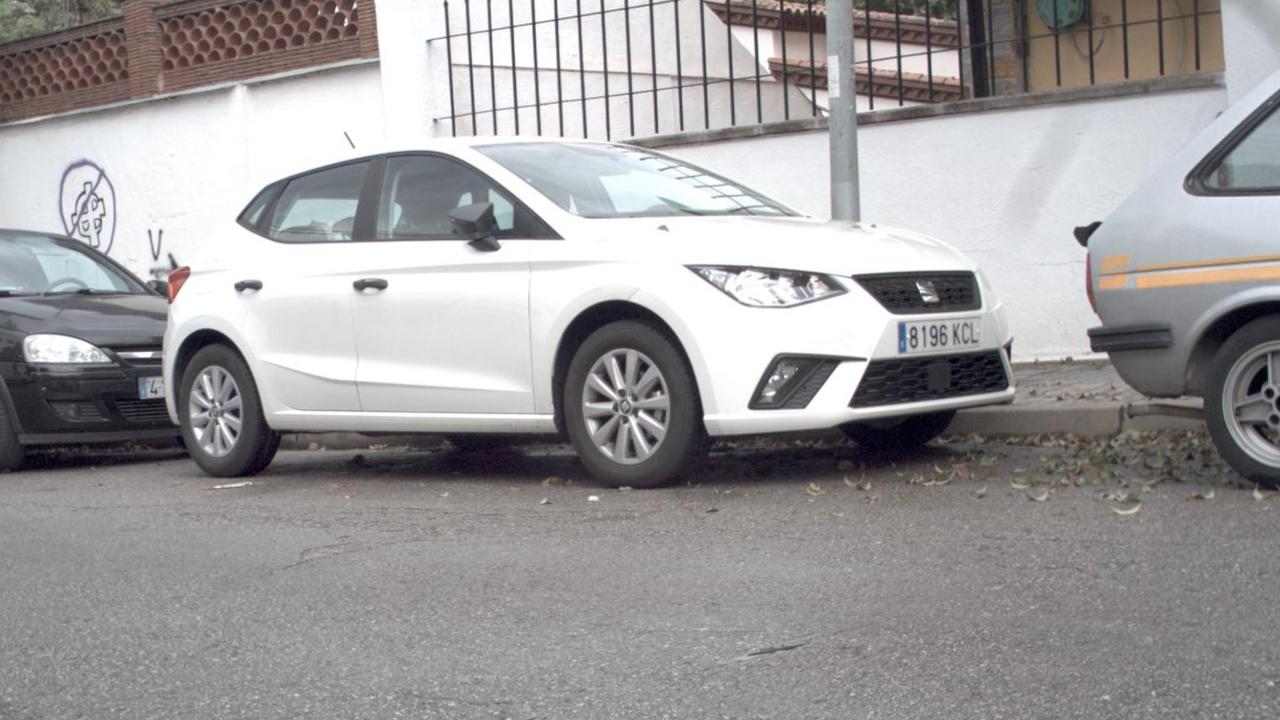}
&\hspace{-.23cm}\includegraphics[width=.28\textwidth,frame]{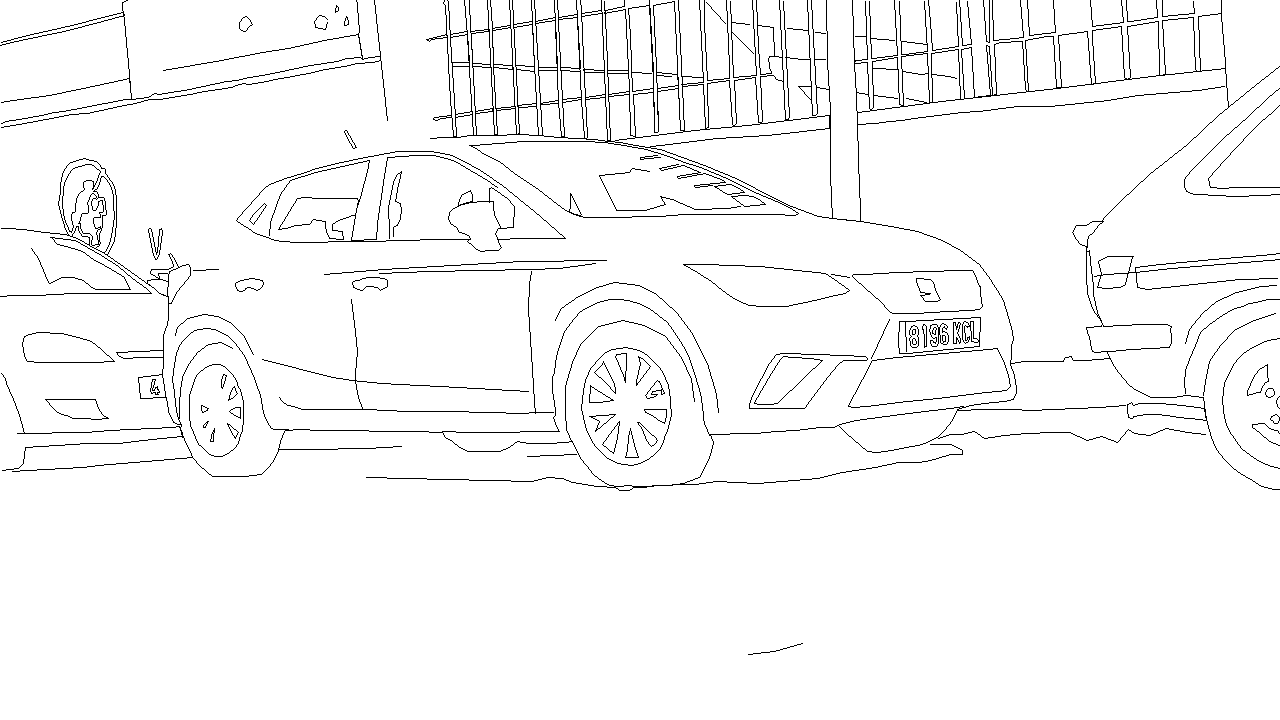}
&\hspace{-.23cm}\includegraphics[width=.28\textwidth,frame]{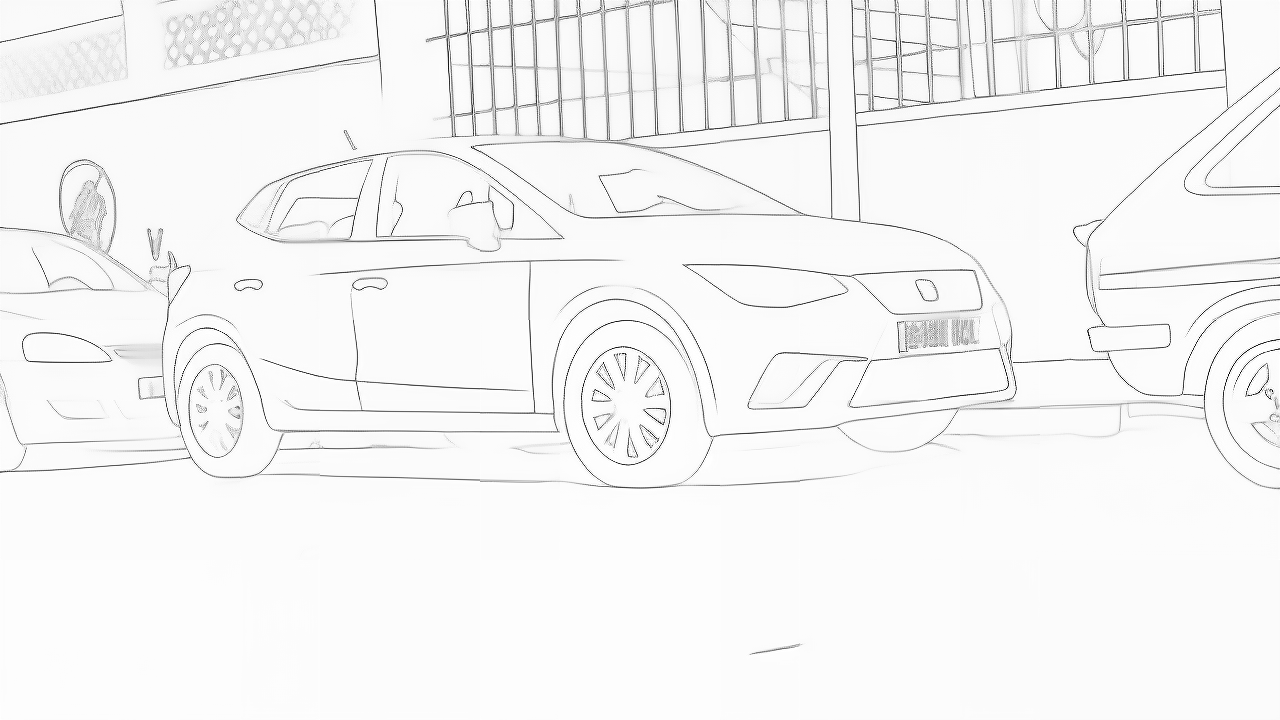}
\vspace{-.06cm}\\

\hspace{-.23cm} Input & \hspace{-.23cm} GT  &\hspace{-.23cm}  Ours\\
\end{tabular}
\caption{Qualitative results on the BIPED~\cite{poma2020dense} dataset.}
\label{fig_biped}
\end{figure*}

In this section, we report more qualitative results on BSDS500~\cite{arbelaez2010contour} , Multicue~\cite{mely2016systematic}, NYUD~\cite{silberman2012indoor} and BIPED~\cite{poma2020dense} datasets. More specifically, Fig.~\ref{fig_compa_bsds} and Fig.~\ref{fig_bsds} shows the visual results on the BSDS500~\cite{arbelaez2010contour} dataset. Fig.~\ref{fig_multicue_boundary} and Fig.~\ref{fig_multicue_edge} 
depicts qualitative results on the Multicue edge and boundary~\cite{mely2016systematic}. Fig.~\ref{fig_nyud} and Fig.~\ref{fig_biped} show the results on the NYUD~\cite{silberman2012indoor} and BIPED~\cite{poma2020dense} datasets, respectively.
\end{multicols}

\begin{algorithm}[htbp]
\caption{Pseudocode of GED in a PyTorch-like style.}
\definecolor{codeblue}{rgb}{0.25,0.5,0.5}
\lstset{
  backgroundcolor=\color{white},
  basicstyle=\fontsize{7.2pt}{7.2pt}\ttfamily\selectfont,
  columns=fullflexible,
  breaklines=true,
  captionpos=b,
  commentstyle=\fontsize{7.2pt}{7.2pt}\color{codeblue},
  keywordstyle=\fontsize{7.2pt}{7.2pt},
}
\begin{lstlisting}[language=python,mathescape=true]
# img: the RGB image condition
# edge: the corresponding edge maps
# gran: compute by normalization according to the edge pixels in each edge maps
# p: image caption generated by Minigpt-4
# pre-trained stable diffusion (SD): including VAE encoder $\mathcal{E}$, VAE decoder $\mathcal{D}$, denoising U-Net $\mathcal{U}_\theta$ and text encoder $\mathcal{E}_l$

$\text{\textbf{def}}$ computer_loss(img, edge, gran, p, SD):
    # return loss function for training
    # img shape: (1, 3, H, W), edge shape: (4, 1, H, W), gran shape: (4, 1), one image corresponding 4 different edge maps

    # For each edge map, the image is the same
    image = image.repeat(4, 1, 1, 1)
    # The input channel for VAE encoder should be 3
    edge = edge.repeat(1, 3, 1, 1)

    # Extract latent image and edge maps, the shape is (4, 4, H/8, W/8)
    with torch.no_grad():
        latent_img = $\mathcal{E}$(img) 
        latent_edge = $\mathcal{E}$(edge)

    # The corresponding timestep is fixed as 1
    timestep = torch.ones((img.shape[0], )).long()

    # Extract text features, the shape is (4, 77, 1024)
    text_embeds = $\mathcal{E}_l$(p)

    #Make the output of denoising U-Net to align with latent edge maps, gran is encoder by two FCs
    predict_latent_edge = $\mathcal{U}_\theta$(latent_img, timestep, encoder_hidden_states = text_embeds, timestep_cond = gran)

    #Decode the edge prediction into the pixel space, the output channel is 3, so we compute the mean as the final results
    with torch.no_grad():
        predict_edge = $\mathcal{D}$(predict_latent_edge)
        predict_edge = torch.mean(predict_edge, dim = 1, keepdim = True)
        predict_edge = (predict_edge + 1.0) / 2.0

    #Normalize the predicted edge maps to predicted granularity, max and min are obtained from the dataset
    predict_gran = (torch.sum(predict_edge) - min) / (max - min) 

    #Calculate the loss between predicted latent edge maps and ground truth latent edge maps 
    loss_mse = F.mse_loss(predict_latent_edge, latent_edge, reduction = "mean")
    
    #Calculate the loss to keep the relative ordinal relationships among output granularities, d means Euclidean distance
    loss_ord = F.mse_loss(d(predict_latent_edge), d(latent_edge), reduction = "mean") + F.mse_loss (predict_gran, gran, reduction = "mean")

    loss = loss_mse + loss_ord
    
    $\text{\textbf{return}}$ loss
\end{lstlisting}
\label{alg:loss}
\end{algorithm}

\end{document}